\documentclass{article}

\usepackage{PRIMEarxiv}

\usepackage[utf8]{inputenc} 
\usepackage[T1]{fontenc}    
\usepackage{hyperref}       
\usepackage{url}            
\usepackage{booktabs}       
\usepackage{amsfonts}       
\usepackage{nicefrac}       
\usepackage{microtype}      
\usepackage{lipsum}
\usepackage{fancyhdr}       
\usepackage{graphicx}       
\graphicspath{{media/}}     

\usepackage{multirow}
\usepackage{subfigure}
\usepackage{amssymb}
\usepackage{amsmath}
\usepackage{siunitx}

\newcommand{\by}{{\bf y}}
\newcommand{\bz}{{\bf z}}
\newcommand{\bW}{{\bf W}}
\newcommand{\bu}{{\bf u}}
\newcommand{\bb}{{\bf b}}
\newcommand{\bV}{{\bf V}}

\newcommand{\R}{{\mathbb R}}
\newcommand{\Dt}{{\Delta t}}

\title{Neural oscillators for magnetic hysteresis modeling
}

\author{
  Abhishek Chandra\thanks{Equal contribution} \\
   Department of Electrical Engineering\\
  TU Eindhoven, The Netherlands \\
  \texttt{a.chandra@tue.nl} \\
   \And
  Taniya Kapoor\textsuperscript{*} \\
  Faculty of Civil Engineering and Geosciences \\
  TU Delft, The Netherlands \\
  \And
  Bram Daniels, Mitrofan Curti \\
  Department of Electrical Engineering \\
  TU Eindhoven, The Netherlands \\
  \And
  Koen Tiels \\
  Department of Mechanical Engineering \\
  TU Eindhoven, The Netherlands \\
  \And
  Daniel M. Tartakovsky \\
  Department of Energy Science and Engineering \\
  Stanford University, USA \\
  \And
  Elena A. Lomonova \\
  Department of Electrical Engineering \\
  TU Eindhoven, The Netherlands \\
}

\begin{document}
\maketitle

\begin{abstract}
\emph{Hysteresis} is a ubiquitous phenomenon in science and engineering; its modeling and identification are crucial for understanding and optimizing the behavior of various systems. We develop an ordinary differential equation-based recurrent neural network (RNN) approach to model and quantify the hysteresis, which manifests itself in sequentiality and history-dependence. Our \emph{neural oscillator, HystRNN}, draws inspiration from coupled-oscillatory RNN and phenomenological hysteresis models to update the hidden states. The performance of HystRNN is evaluated to \emph{predict generalized scenarios}, involving first-order reversal curves and minor loops. The findings show the ability of HystRNN to generalize its behavior to previously untrained regions, an essential feature that hysteresis models must have. This research highlights the \emph{advantage} of neural oscillators over the traditional RNN-based methods in capturing complex hysteresis patterns in magnetic materials, where traditional rate-dependent methods are inadequate to capture intrinsic nonlinearity. 
\end{abstract}


\section{Introduction}
\begin{figure}[t]
    \centering
    \subfigure[Major loop]{\label{fig:1a}{\includegraphics[height=4.5cm, width=4.5cm]{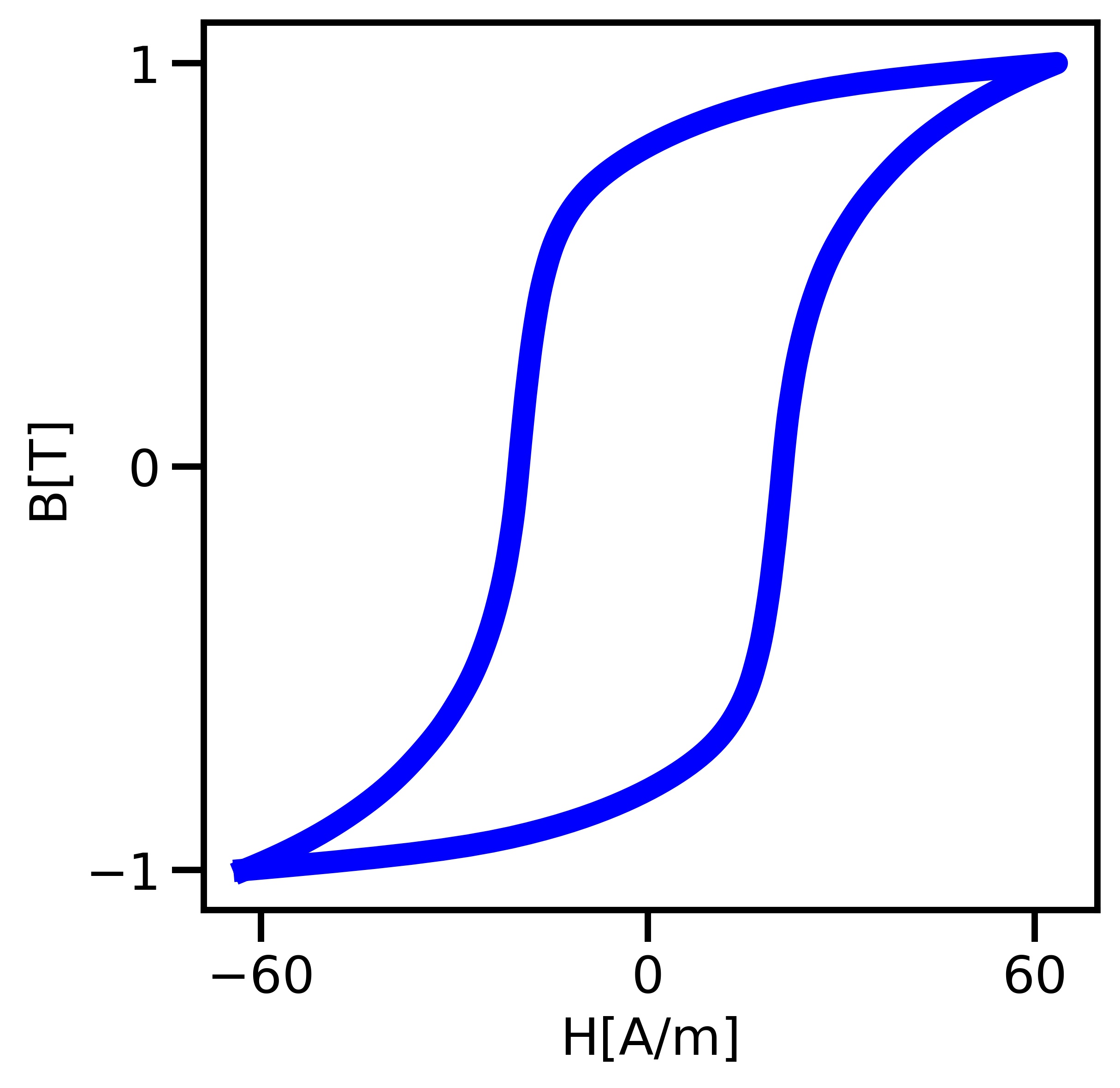}}}\hfill
      \subfigure[FORCs]{\label{fig:1b}{\includegraphics[height=4.5cm, width=4.5cm]{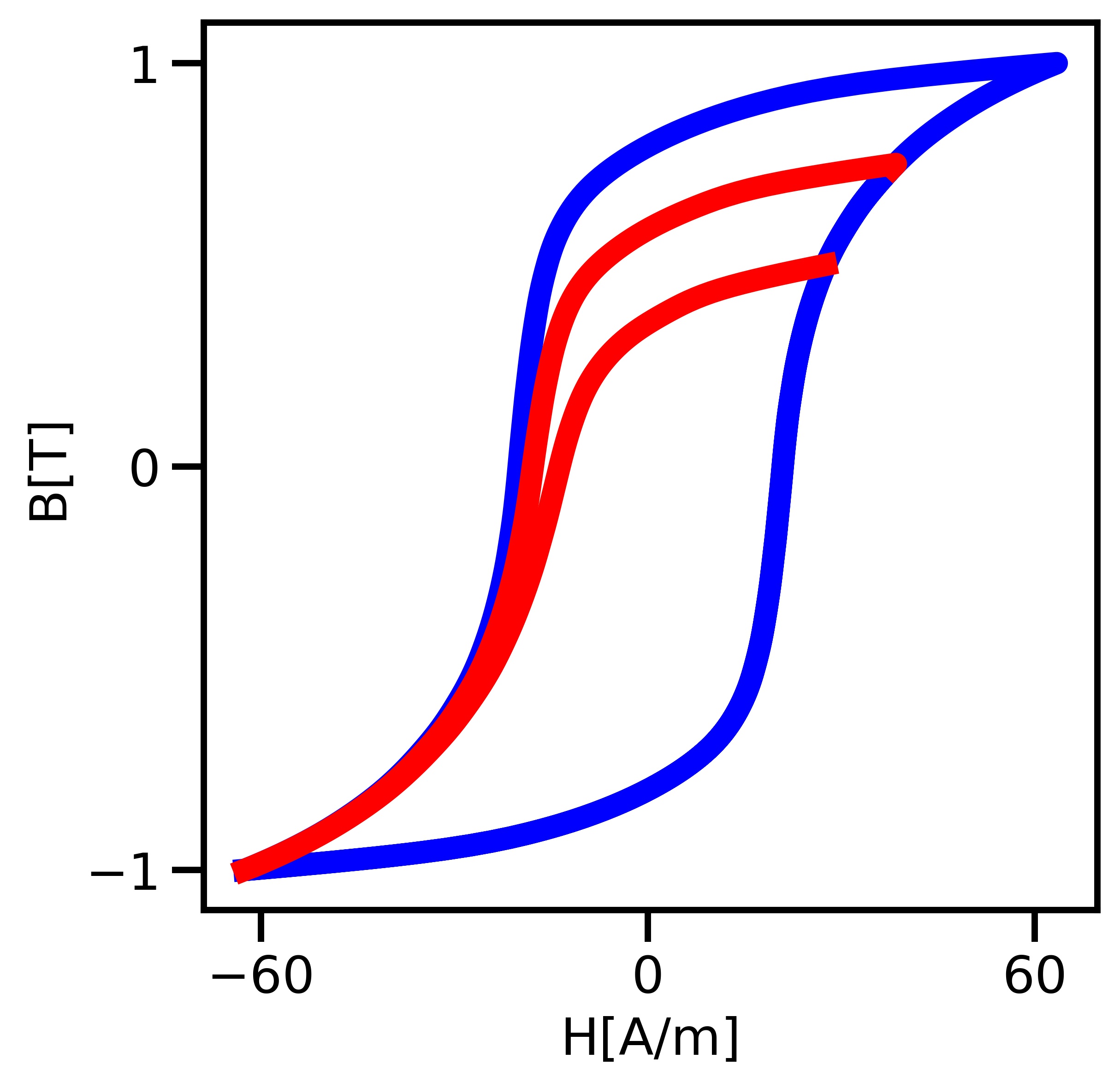}}} \hfill
      \subfigure[Minor loop]{\label{fig:1c}{\includegraphics[height=4.5cm, width=4.5cm]{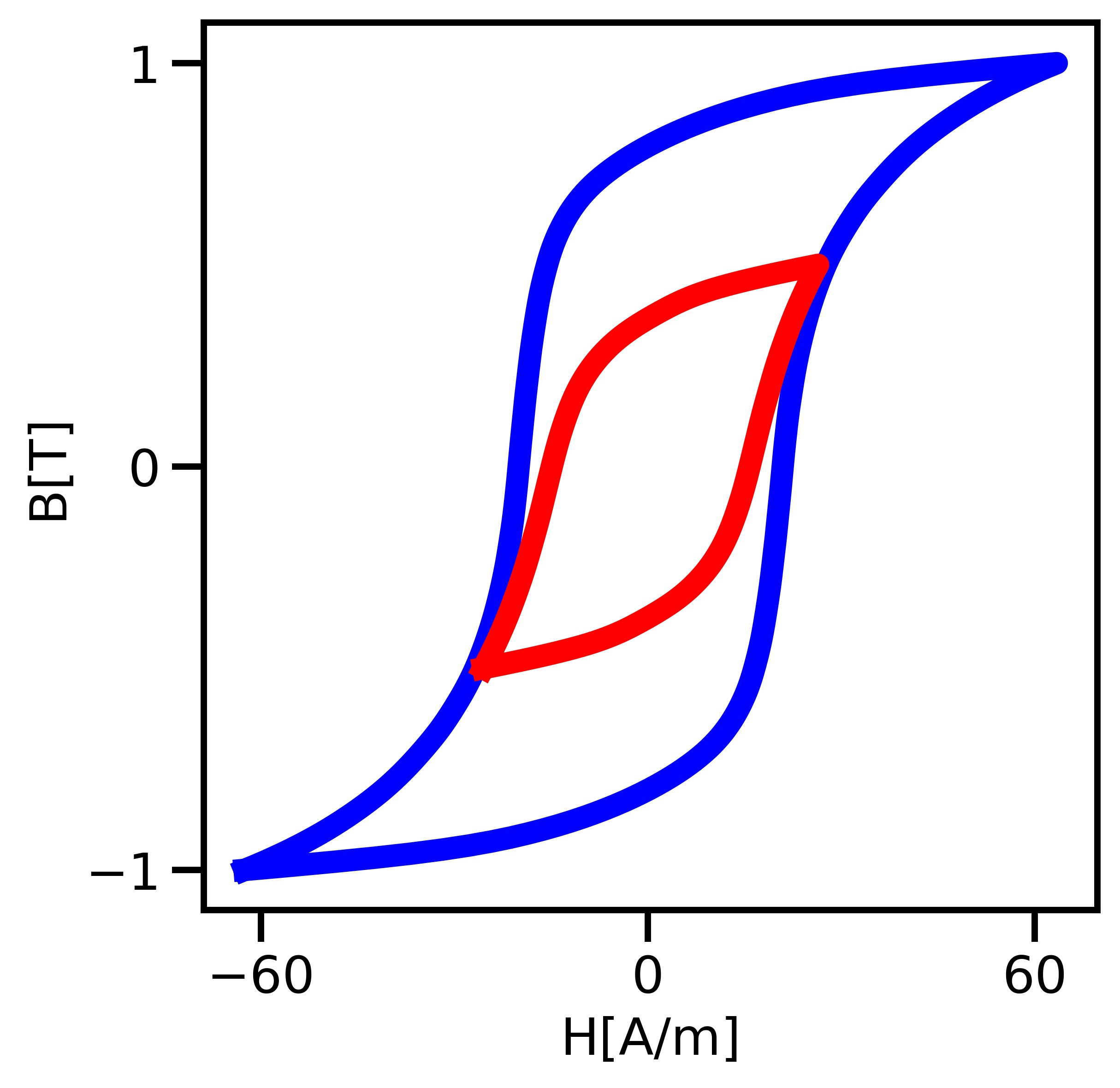}}} \hfill 
    \caption{$B-H$ magnetic hysteresis curves (FORC = first-order reversal curve)} 
    \label{fig1}
\end{figure}

Magnetic hysteresis pertains to a prevalent observed phenomenon in ferromagnetic and ferrimagnetic materials where the change in magnetization response \emph{lags behind} variations in the applied magnetic field. Specifically, the hysteresis effect is characterized by a delay in the magnetic flux density ($B$) to changes in the applied magnetic field strength ($H$), exhibiting \emph{history dependency}, nonlinearity and non-monotonicity \cite{bertotti2005science}. The relationship between $B$ and $H$ fields is represented as a hysteresis curve $\mathcal{C}$ ($B-H$ curve), which plays a pivotal role in comprehending hysteresis and governing the magnetization process during alterations in $H$ (Fig.~\ref{fig1}). The hysteresis loop offers insights into material behavior; for instance, the area of the hysteresis loop signifies the energy dissipated as heat during each cycle of magnetization and demagnetization.

Accurate hysteresis modeling is pivotal in enhancing the operational \emph{efficiency} of electrical machines. For instance, in engineering systems that involve the movement of cables, hysteresis requires more sophisticated control strategies to compensate for its effects \cite{vlachas2021local}. Similarly, the efficiency of electrical machines is intrinsically linked to the \emph{precise modeling} of the hysteresis characteristics exhibited by the steel materials employed \cite{ceylan2022novel}. Incorporating a robust model would avoid the costly manufacturing of multiple prototypes. Mathematically, the primary objective of hysteresis modeling is to predict the sequence of $B$ values that correspond to a given sequence of $H$ values. However, the relationship between $B$ and $H$ defies the mathematical definition of a single-valued function. Consequently, conventional function approximation techniques are \emph{not suitable} for modeling hysteresis as a function with domain $H$ and codomain $B$ \cite{chen2021diagonal}.

Traditionally, modeling the hysteresis behavior is based on fundamental principles of physics \cite{bertotti1998hysteresis}. However, in practical engineering scenarios, the manifestation of hysteresis behavior often stems from complex, large-scale effects and the multiphysical nature of the system, rendering deterministic models \emph{inaccurate} \cite{bertotti1998hysteresis, bertotti2005science}. Consequently, phenomenological models are employed, establishing connections between desired behaviors and specific underlying phenomena rooted in principles of thermodynamics or elasticity, for instance. Notable phenomenological models include the Preisach \cite{preisach1935magnetische, bertotti2005science}, Jiles-Atherton \cite{jiles1986theory}, and Bouc-Wen models \cite{bouc1967forced, wen1976method}. Generalizing these models across disciplines and incorporating them into systematic modeling approaches, fitting them to experimental data, and integrating them into other mathematical models \emph{pose significant challenges} \cite{chandra2023discovery}, such as sophisticated optimization techniques and increased computational burden \cite{lin2022identification}.

To mitigate these limitations of phenomenological models, feed-forward neural networks (FFNNs) are commonly used for modeling magnetic hysteresis \cite{serpico1998magnetic, makaveev2001modeling, quondam2023deep, farrokh2022hysteresis}. However, owing to the \emph{absence of a functional relationship} between $B$ and $H$ fields, the traditional FFNN approach with input $H$ and output $B$ is inadequate and suboptimal. Instead, studies \cite{makaveev2001modeling, quondam2023deep} propose $H$ and $B_{-1}$ as input and $B$ as output during training, where $B_{-1}$ denotes the previous $B$ value \cite{chen2021diagonal}. The notation $B_{-1}$ is discussed in detail in the 'Method' section. However, this approach is characterized by two notable limitations. First, it lacks the incorporation of \emph{sequential information} and fails to capture interdependencies among output values, hence not respecting the underlying physics of the problem. Second, this strategy exhibits \emph{limitations in generalizing} to new situations, as it relies on single-step prior information during training \cite{hwang2015recurrent}. Consequently, these models \emph{struggle to extrapolate} to scenarios beyond training data, limiting broader applications that require robust generalization.

To address the limitations faced by FFNNs, models centered on recurrent neural networks (RNNs) \cite{chen2021diagonal, saghafifar2002dynamic} have been employed, which provide a \emph{natural framework} for modeling the sequential hysteretic nature. However, the models that employ traditional RNNs, gated recurrent unit (GRU) \cite{cho2014learning}, long-short-term memory (LSTM) \cite{hochreiter1997long} and their variants exhibit \emph{limitations} regarding their ability to \emph{generalize effectively} to unseen $H$ variations, as we present in the \emph{current work}. Although these recurrent networks model the underlying relationship and predict hysteresis loops exceedingly well as an interpolation task, the \emph{primary objective} of achieving \emph{robust generalization remains inadequately addressed} \cite{serrano2023magnet}. An optimal recurrent-based technique should excel in \emph{both} interpolation tasks and demonstrate reasonable accuracy in generalization, effectively predicting $B$ sequences for unseen $H$ sequences.

A possible approach to accomplishing efficient generalization could be to enforce the recurrent architecture to \emph{incorporate the underlying dynamics}. An efficient way to represent time-varying dynamics involves representation through ordinary differential equations (ODEs) and dynamical systems, recognized for their capacity to model diverse, intricate and nonlinear phenomena across natural, scientific, and engineering domains \cite{brunton2022data}. This inclusion of inherent dynamics, which should effectively \emph{encapsulate crucial physical attributes} of the underlying magnetic material, motivates us to employ a system of ODEs to update the hidden states of the recurrent architecture, referred to as \emph{neural oscillators}. Recently, neural oscillators have shown significant success in machine learning and artificial intelligence and have been shown to handle the exploding and vanishing gradient problem effectively with high expressibility \cite{rusch2020coupled, rusch2021unicornn, rusch2021long, queiruga2021stateful, rusch2022graph}. The recent universal approximation property \cite{lanthaler2023neural} also supports our belief in modeling magnetic hysteresis with neural oscillators. 

Our neural oscillator, referred to as \emph{HystRNN} (hysteresis recurrent neural network), is influenced by the principles of the coupled-oscillator recurrent neural network (CoRNN) \cite{rusch2020coupled}, which integrates a second-order ordinary differential equation (ODE) based on mechanical system principles. CoRNN considers factors such as oscillation frequency and damping in the system. However, for magnetic hysteresis, these physical attributes are less significant. Instead, we focus on embedding the hysteric nature within the ODE formulation. We \emph{leverage phenomenological differential hysteresis models} to accomplish this, recognizing that models like Bouc-Wen \cite{bouc1967forced, wen1976method}, and Duhem \cite{oh2005semilinear} utilize the absolute value function to represent the underlying dynamics. By incorporating this function into our model, we can \emph{effectively capture} and control the shape of the hysteresis loop. This integration into the recurrent model is expected to \emph{facilitate robust generalization}, as it inherently captures the shape of the hysteresis loop, preserving \emph{symmetry} and \emph{structure}.

In this manuscript, we model \emph{nonoriented electrical steel} (NO27) to test the validity of the proposed method for magnetic materials. The hysteresis loops employed and modeled in this work are acquired using the Preisach model for an Epstein frame. This model has been adjusted to adhere to the IEC standard, and the core was assembled using 16 strips of NO27-1450H material. More details about the Preisach model are provided in supplementary material \textbf{SM} \S \textbf{B}. We train all our models on a \emph{major loop} (represented by a blue loop in Fig.~\ref{fig:1a}) and use the trained model to test two different generalization tasks: predicting \emph{first-order reversal curves} (FORCs, represented by red curves in Fig.~\ref{fig:1b}) and \emph{minor loops} (represented by a red loop in Fig.~\ref{fig:1c}). We perform experiments for \emph{four different $B$ fields} with applications relevant to modeling electrical machines.

The remainder of the manuscript is structured as follows. The section `Generalization in hysteresis' presents the \emph{challenge} of this manuscript and explains how the task amounts to \emph{generalization}. The section `Method' formulates the proposed HystRNN method and explains it in detail. The section `Numerical experiments' \emph{validates} the proposed method through a series of numerical experiments. Finally, the `Conclusions' section collates the key findings and implications of this study.

\section{Generalization in hysteresis}

Traditional supervised machine learning methodologies employed to model magnetic hysteresis train the model for $(H_i, B_i) \in \mathcal{C}_1$, where $1 \leq i \leq N$, $i \in \mathbb{Z}$ and $N$ is the number of training samples. The trained model is then tested on $(H_k, B_k) \in \mathcal{C}_2$, where $1 \leq k \leq M$, $k \in \mathbb{Z}$ and $M$ is the number of testing samples. Traditionally, $\mathcal{C}_2 \subset \mathcal{C}_1$, with $H_i \neq H_k$. However, this prediction reduces to an \emph{interpolation task} \cite{khan2022generalizable}. In \emph{contrast}, we are interested in training the model for $(H_i, B_i) \in \mathcal{C}_1$ and predicting a hysteresis trajectory for $(H_k, B_k) \in \mathcal{C}_2$, where $\mathcal{C}_2 \nsubseteq \mathcal{C}_1$, and $\mathcal{C}_2 \cap \mathcal{C}_1 = \phi$. Here, $\phi$ denotes the null set. Precisely, we train all our models on the major loop ($\mathcal{C_\mathrm{major}}$) as shown in Fig.~\ref{fig:1a}. Then the trained model is tested for two different scenarios. First the FORCs ($\mathcal{C_\mathrm{FORC}}$) shown in Fig.~\ref{fig:1b} and second the minor loops ($\mathcal{C_\mathrm{minor}}$) presented in Fig.~\ref{fig:1c}.

Modeling FORCs and minor hysteresis loops play a significant role in \emph{analyzing} magnetic materials. FORC modeling reveals intricate interactions, enabling the differentiation between magnetization components that can be reversed and those that cannot. This knowledge is pivotal in the optimization of magnetic devices like memory and sensor technologies \cite{kouhpanji2021first, gilbert2021forc, stancu2021characterization}. Minor hysteresis loop modeling complements this by providing insights into localized variations in magnetic behavior.

\begin{figure}[t]
    \centering
    \subfigure[$H$ vs time]{\label{2a}\includegraphics[width=0.48\textwidth]{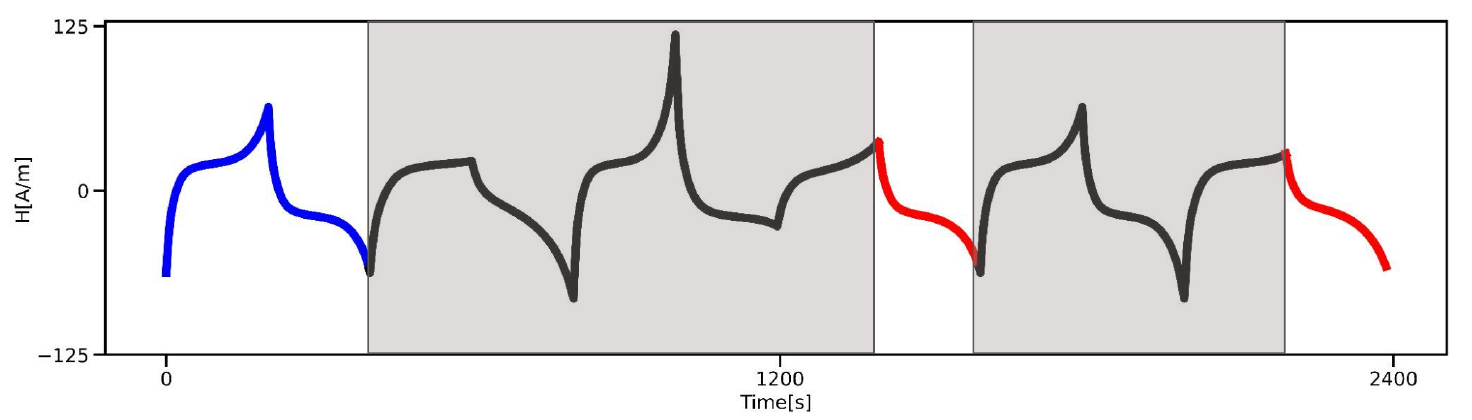}}\hfill
    \subfigure[$B$ vs time]{\label{2b}\includegraphics[width=0.48\textwidth]{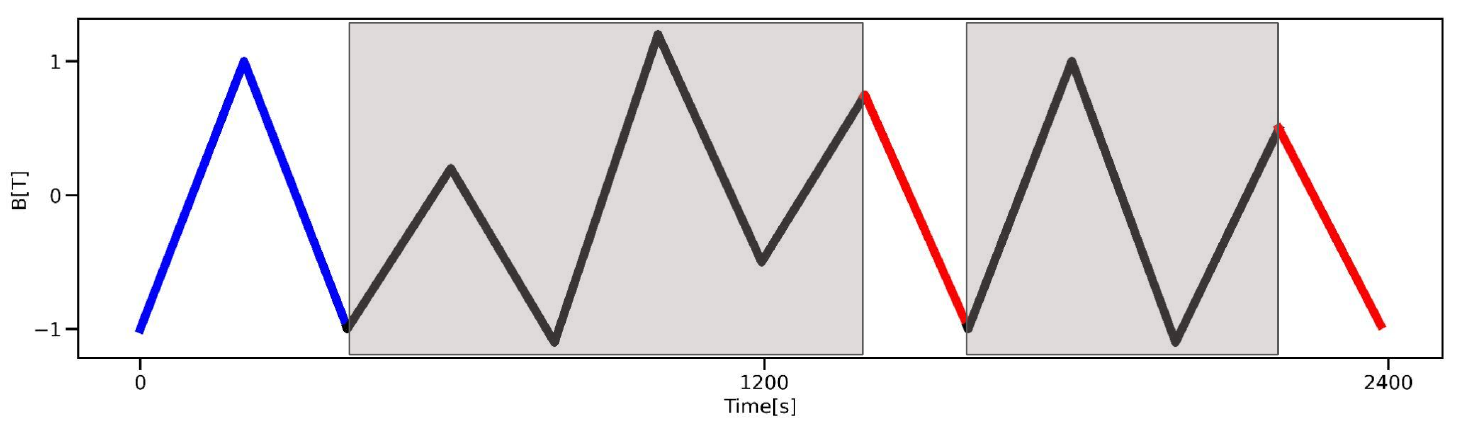}}\hfill
    \subfigure[$H$ vs time]{\label{2c}\includegraphics[width=0.48\textwidth]{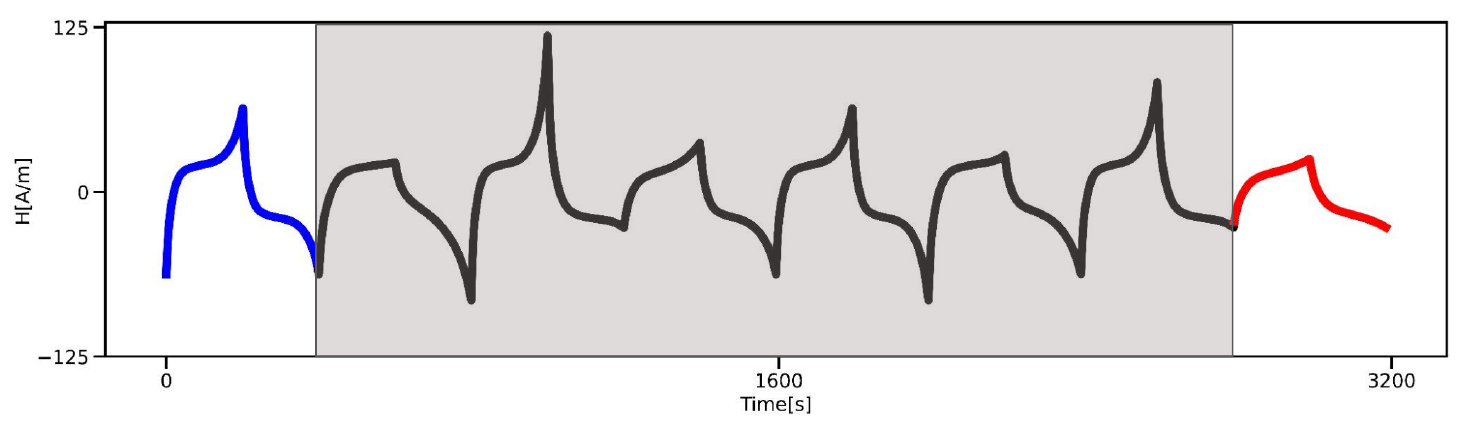}}\hfill
    \subfigure[$B$ vs time]{\label{2d}\includegraphics[width=0.48\textwidth]{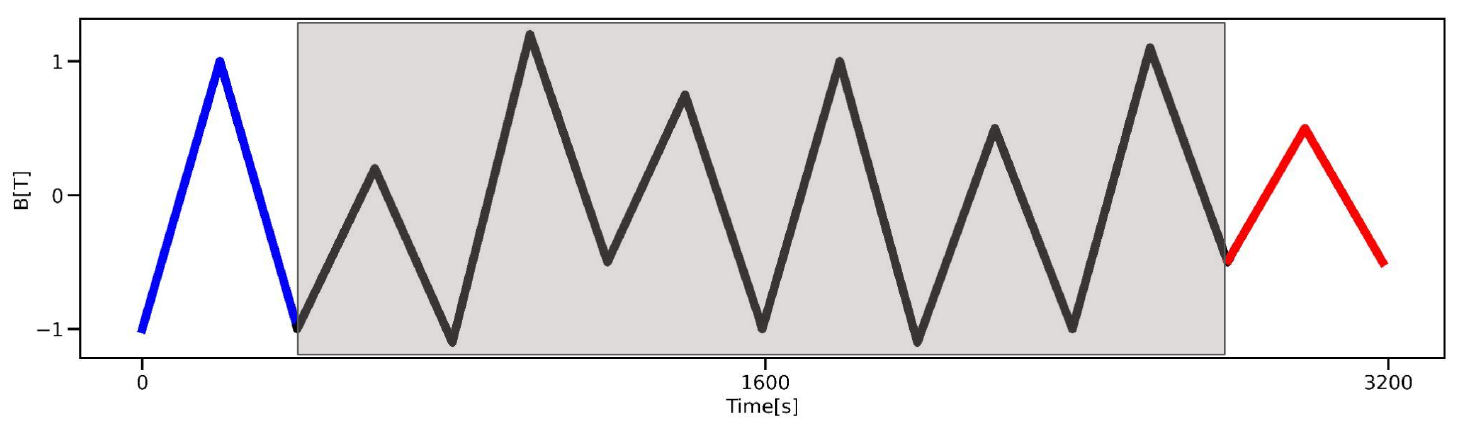}}\hfill
    \caption{Variation of magnetization and demagnetization in a magnetic material over time}
    \label{fig:2}
\end{figure}

More insights on how predicting $\mathcal{C_\mathrm{FORC}}$ and $\mathcal{C_\mathrm{minor}}$ entails to a generalization task is provided in Fig.~\ref{fig:2}. In Fig.~\ref{2a} and ~\ref{2b}, the time series of $H$ and $B$ fields are shown, respectively. The blue curve represents the training data ($H$ vs $B$ is $\mathcal{C_\mathrm{major}}$), and the red curve represents the region in which the prediction is sought ($\mathcal{C_\mathrm{FORC}}$ in this case). The black curve behind the shaded region signifies the \emph{history} that the material has gone through, which is the \emph{series of magnetization and demagnetization} that is \emph{unknown} at the time of testing. Hence, this task amounts to \emph{extrapolation in time or predicting in a generalized scenario}. Similarly, Fig.~\ref{2c} and ~\ref{2d} represent the case for $\mathcal{C_\mathrm{major}}$ and $\mathcal{C_\mathrm{minor}}$. 

\section{Method}
HystRNN utilizes a recurrent structure akin to RNNs, with the \emph{difference} being in the hidden state update. HystRNN \emph{employs ODEs} for updating the hidden states. The approach involves two inputs, $H$ and $B_{-1}$, which are mapped to $B$. The modeling process begins by collecting $N_e$ number of experimental data points ($H_i, B_{-1} := B_i$) for $\mathcal{C_\mathrm{major}}$, where $1 \leq i \leq N_e$, and $i \in \mathbb{Z}$. Subsequently, ($H_j, B_k$) and $B_j$ are taken as the input and output of HystRNN, respectively, where $2 \leq j \leq N_e$, $1 \leq k \leq N_e - 1$, $k = j-1$, and $j,k \in \mathbb{Z}$. The number of training points is denoted by $N = N_e -1$. While sharing certain similarities with some feedforward neural network (FFNN) architectures employed for modeling hysteresis, this training approach diverges by \emph{incorporating a recurrent relationship} that captures \emph{longer-time dynamics} and \emph{output dependencies}, which are absent in FFNNs. Next, the hidden states of HystRNN are updated using the following second-order ODE 

\begin{equation}
\label{eq:ode1}
\begin{aligned}
\by^{\prime \prime} = \sigma_1\left(\bW_1 \by +  \boldsymbol{\mathcal{W}_1} \by^{\prime} + \bV_1 \bu + \bb_1 \right) \\ + \sigma_2\left(\bW_2 |\by|^2 +  \boldsymbol{\mathcal{W}_2} |\by^{\prime}|^2 + \bV_2 |\bu|^2 + \bb_2 \right).
\end{aligned}
\end{equation}
Here, the hidden state of the HystRNN is denoted by $\by = \by (t) \in \R^m$. $\by^{\prime}$ indicates a time derivative, while $\by^{\prime \prime}$ indicates a second-order time derivative. $\bW_1, \bW_2, \boldsymbol{\mathcal{W}_1}, \boldsymbol{\mathcal{W}_2} \in \R^{m \times m}$, and $\bV_1, \bV_2 \in \R^{m \times n}$ are the weight matrices, $n = N \times 2$, and $t$ corresponds to the time at which the training data has been collected. $\bu = \bu(t) \in  \R^{n}$ is the input to HystRNN. 
$\bb_1, \bb_2 \in \R^m$ are the bias vectors. The activation functions $\sigma_{1,2}: \R \mapsto \R$ are taken to be $\sigma_{1, 2} (u) = \tanh(u)$. By setting $\bz = \by^{\prime}(t) \in \R^m$, ~\eqref{eq:ode1} becomes a system of first-order ODEs
\begin{equation}
\label{eq:ode}
\begin{aligned}
\by^{\prime} = \bz, \quad
\bz^{\prime}= \sigma_1\left(\bW_1 \by +  \boldsymbol{\mathcal{W}_1} \bz + \bV_1 \bu + \bb_1 \right) \\ + \sigma_2\left(\bW_2 |\by|^2 +  \boldsymbol{\mathcal{W}_2} |\bz|^2 + \bV_2 |\bu|^2 + \bb_2 \right).
\end{aligned}
\end{equation}
Discretizing the system of ODEs~\eqref{eq:ode} using an explicit scheme for $0 < \Dt < 1$ leads to 
\begin{equation}
\label{eq:brnn}
\begin{aligned}
\by_n &= \by_{n-1} + \Dt \bz_n,\\
\bz_n &= \bz_{n-1} + \Dt \sigma_1\left(\bW_1\by_{n-1} +  \boldsymbol{\mathcal{W}_1} \bz_{n-1} + \bV_1 \bu_{n} + \bb_1 \right) \\ &\quad \Dt \sigma_2\left(\bW_2|\by_{n-1}|^2 +  \boldsymbol{\mathcal{W}_2} |\bz_{n-1}|^2 + \bV_2 |\bu_{n}|^2 + \bb_2 \right).
\end{aligned}
\end{equation}
Finally, the output $\hat{B}$ is computed for each recurrent unit, where $\hat{B} \in \R^{n}$ with $\hat{B} =\mathcal{Q} \by_n$ and $\mathcal{Q} \in \R^{n\times m}$. 

\subsection{Related work}

\begin{figure}
    \centering
    \subfigure[LSTM]{\label{3a}{\includegraphics[height=4.5cm, width=4.5cm]{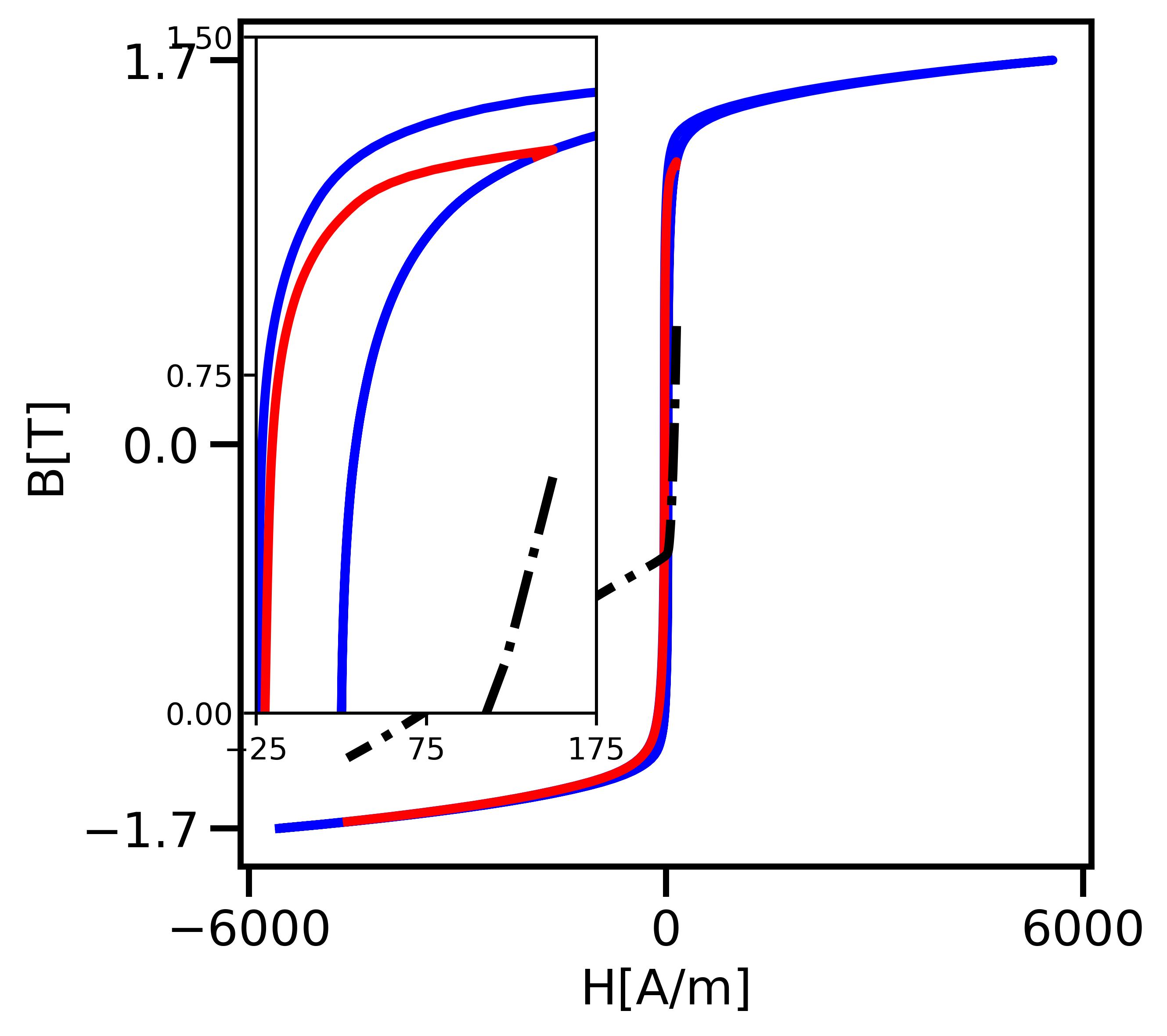}}}\hfill
      \subfigure[GRU]{\label{3b}{\includegraphics[height=4.5cm, width=4.5cm]{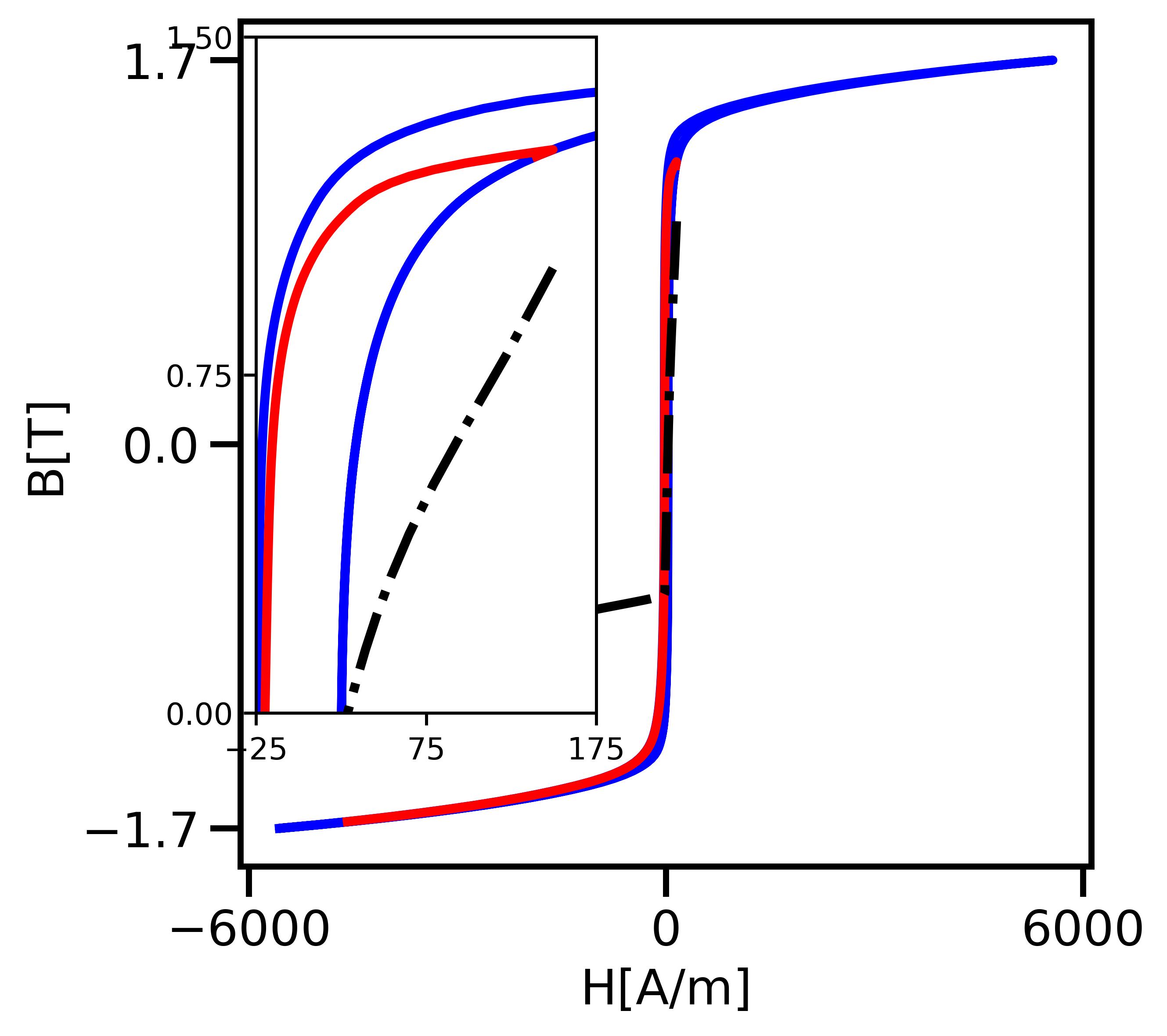}}} \hfill
      \subfigure[HystRNN]{\label{3c}{\includegraphics[height=4.5cm, width=4.5cm]{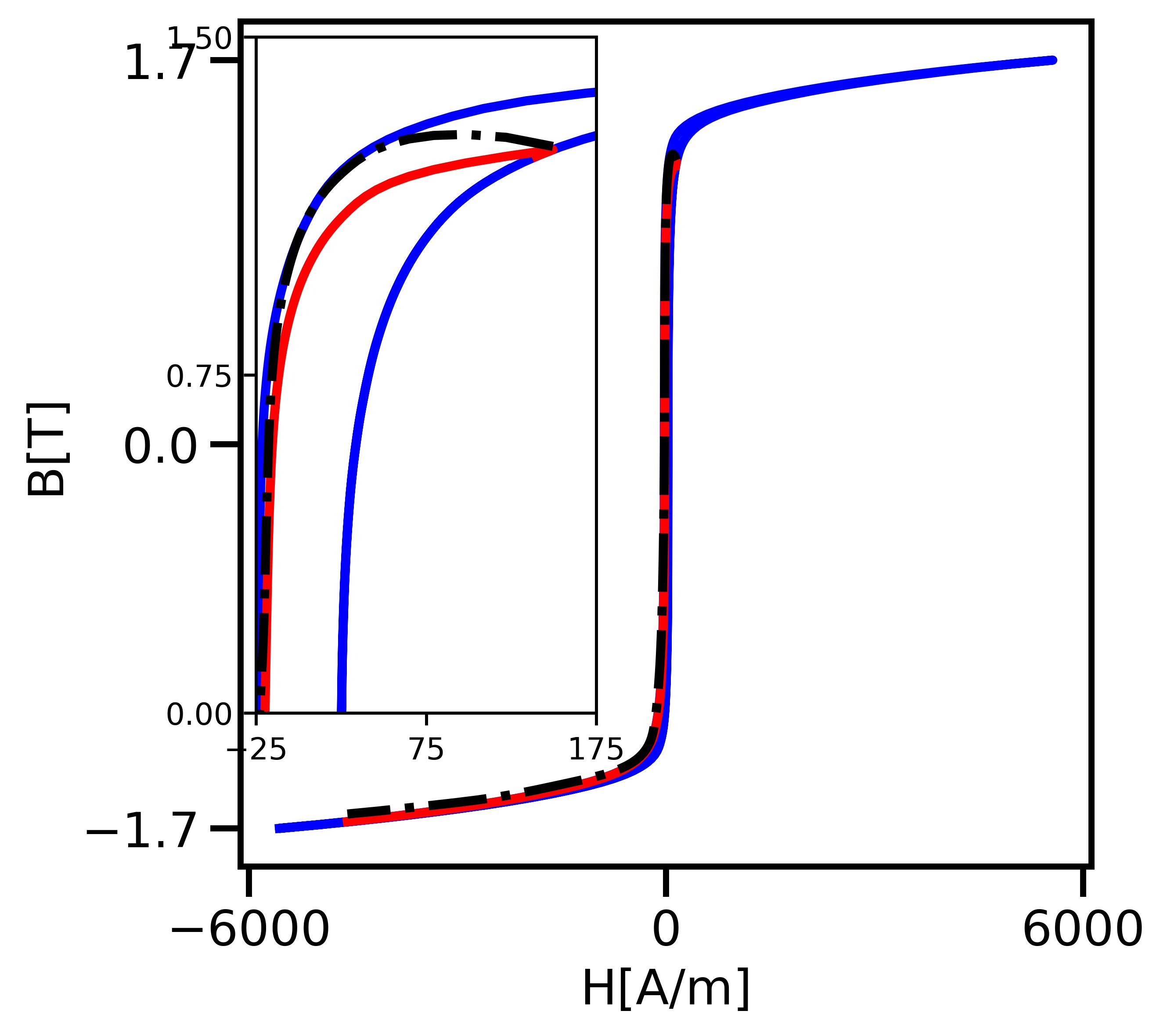}}} \\
      \subfigure[LSTM]{\label{3d}{\includegraphics[height=4.5cm, width=4.5cm]{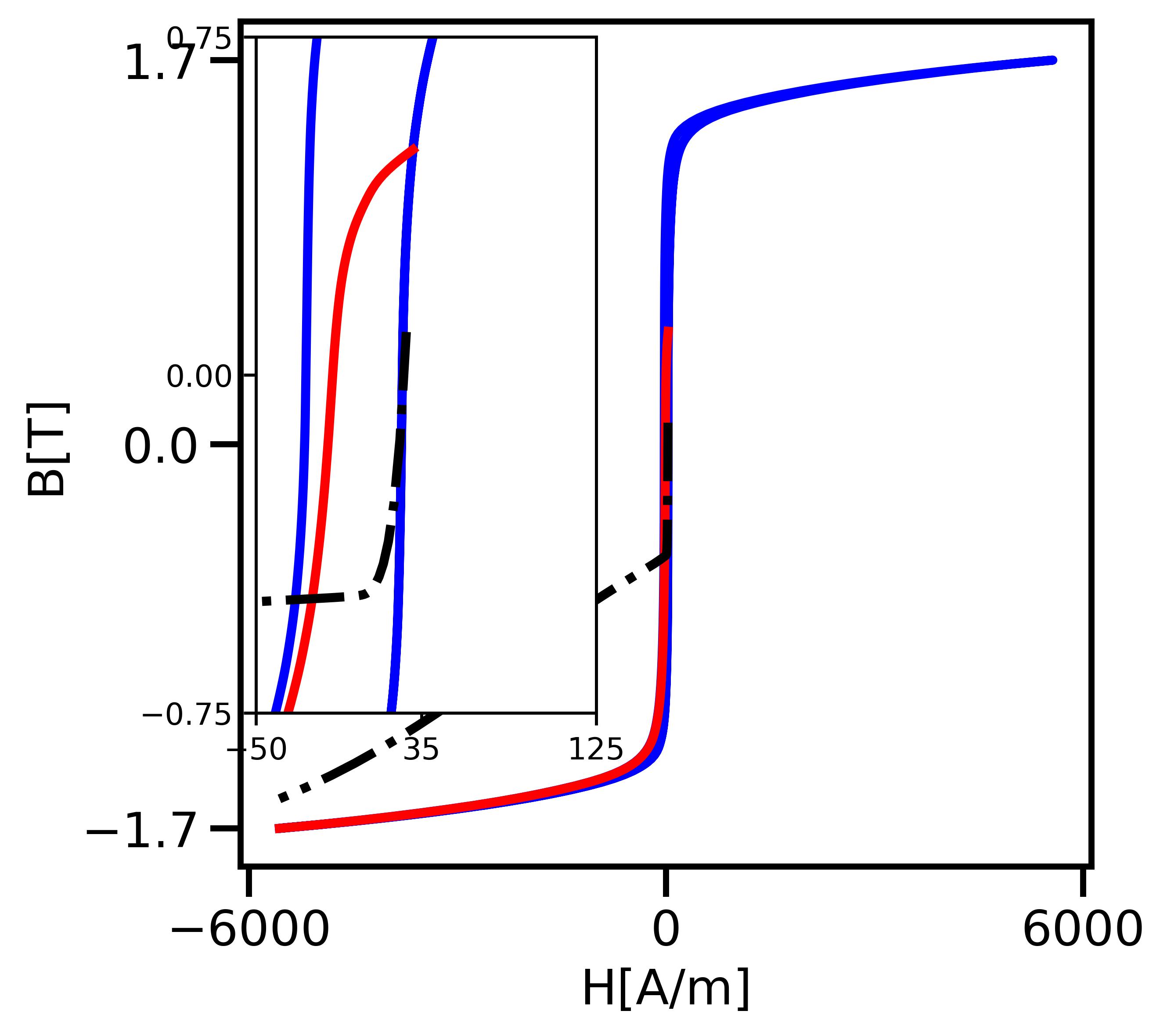}}}\hfill
      \subfigure[GRU]{\label{3e}{\includegraphics[height=4.5cm, width=4.5cm]{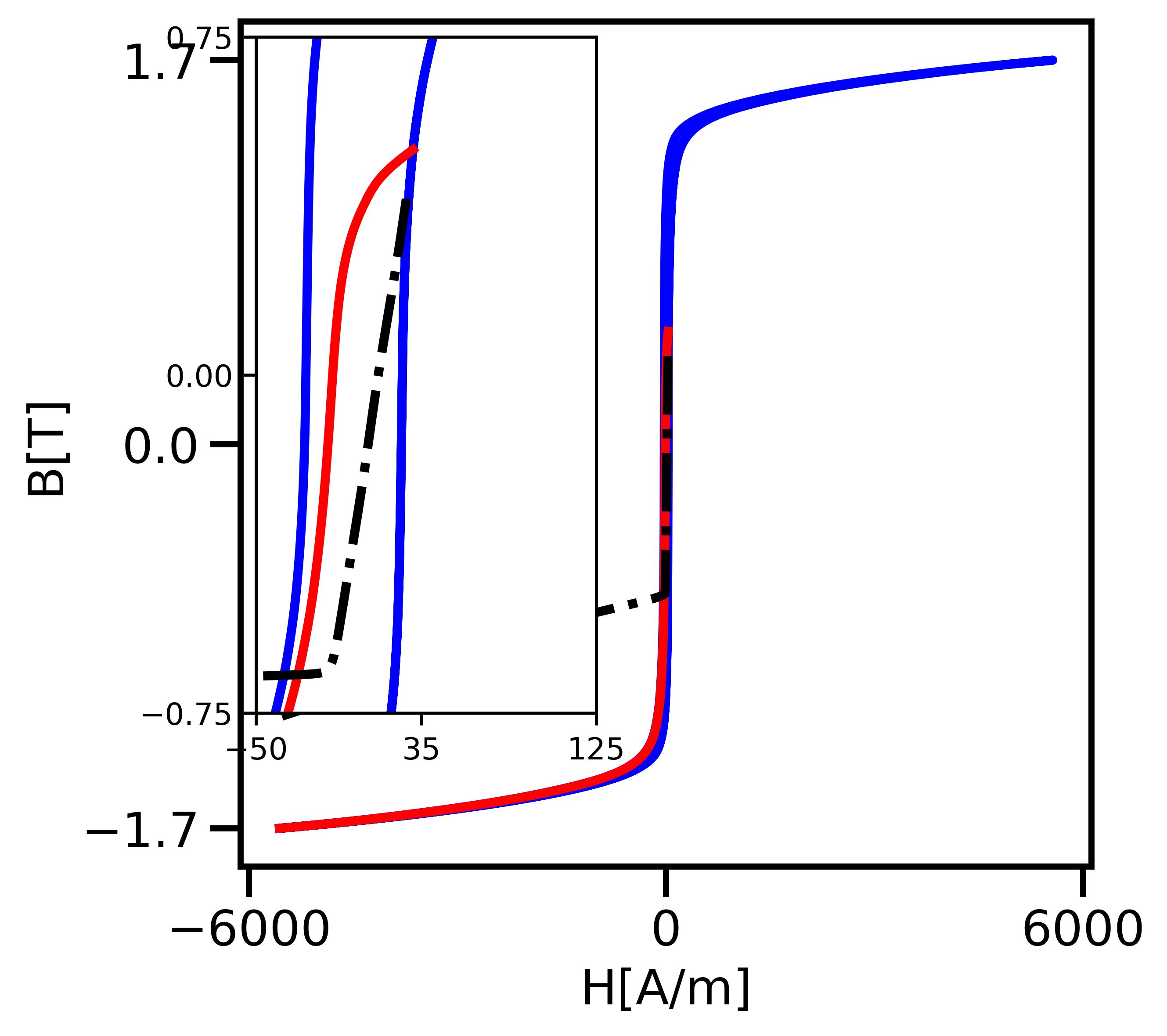}}} \hfill
      \subfigure[HystRNN]{\label{3f}{\includegraphics[height=4.5cm, width=4.5cm]{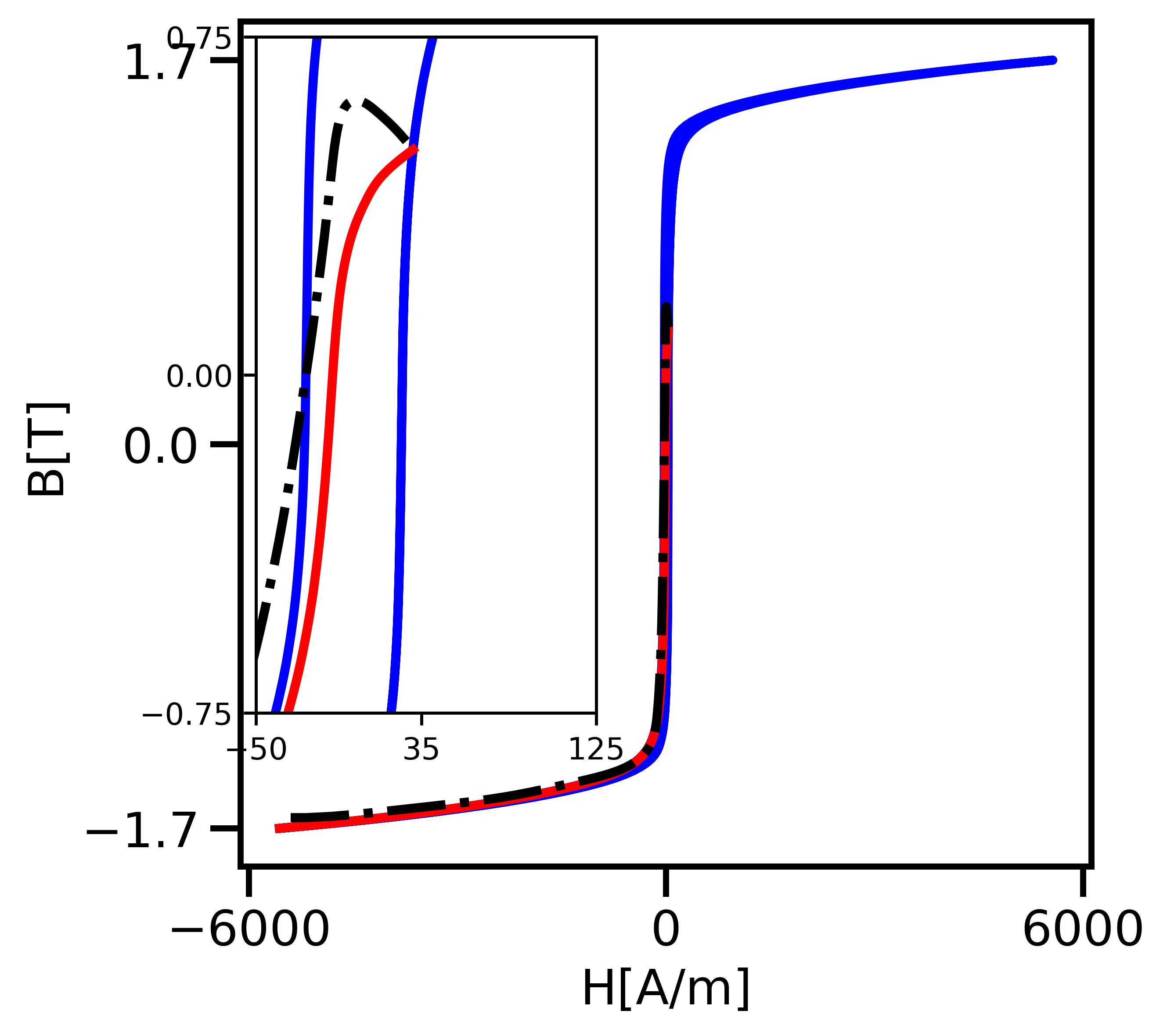}}} \\
      \subfigure[LSTM]{\label{3g}{\includegraphics[height=4.5cm, width=4.5cm]{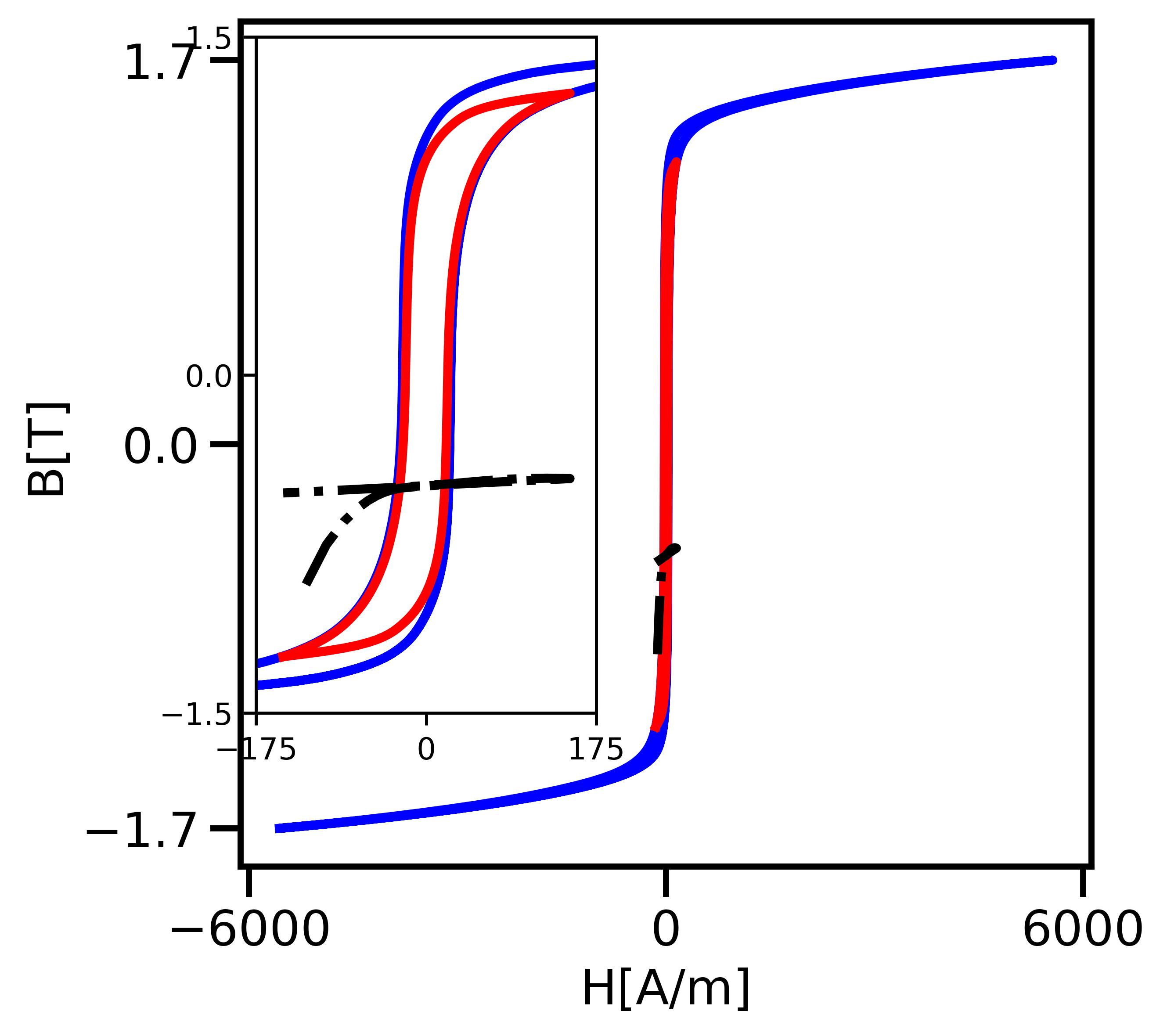}}}\hfill
      \subfigure[GRU]{\label{3h}{\includegraphics[height=4.5cm, width=4.5cm]{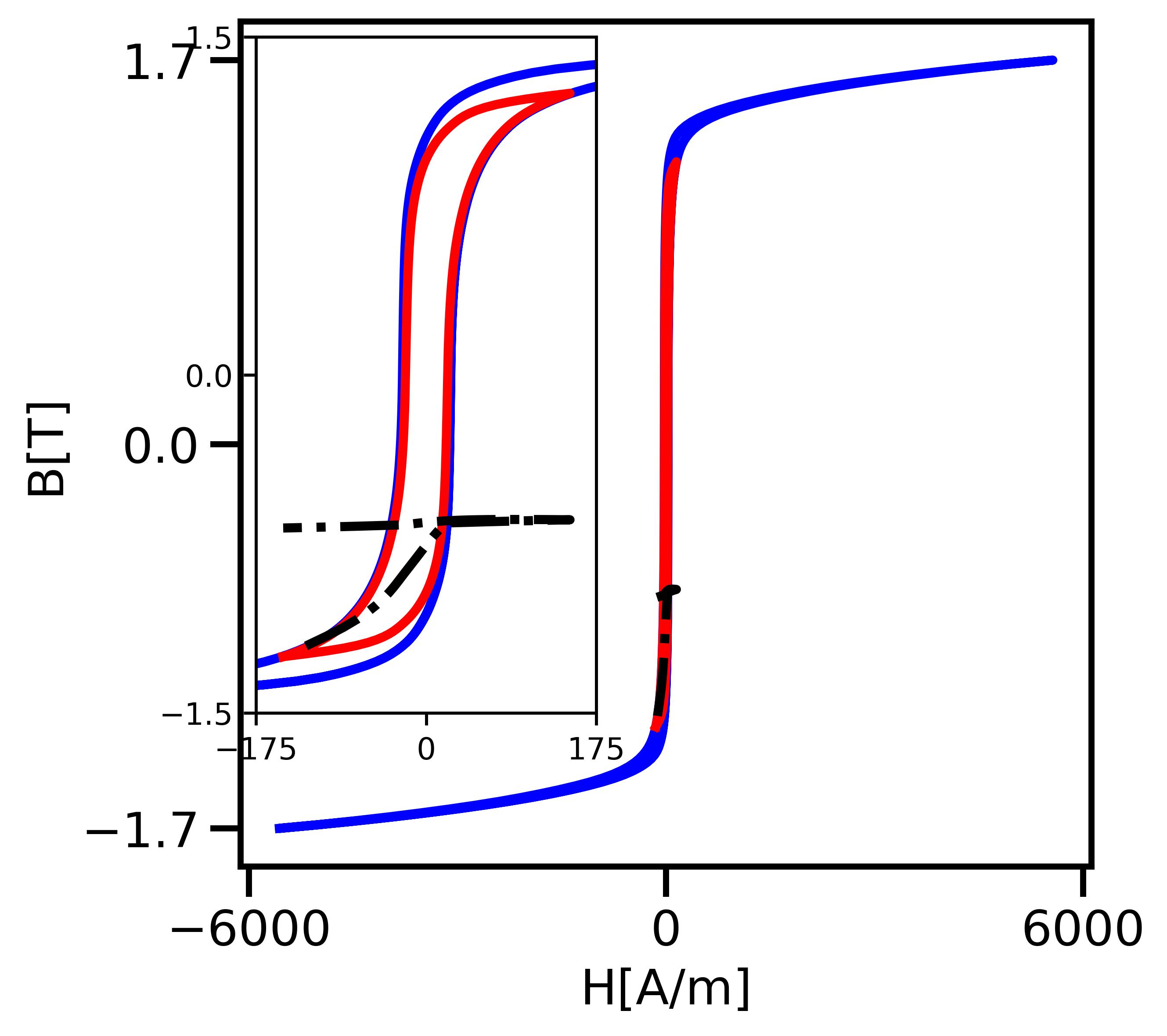}}} \hfill
      \subfigure[HystRNN]{\label{3i}{\includegraphics[height=4.5cm, width=4.5cm]{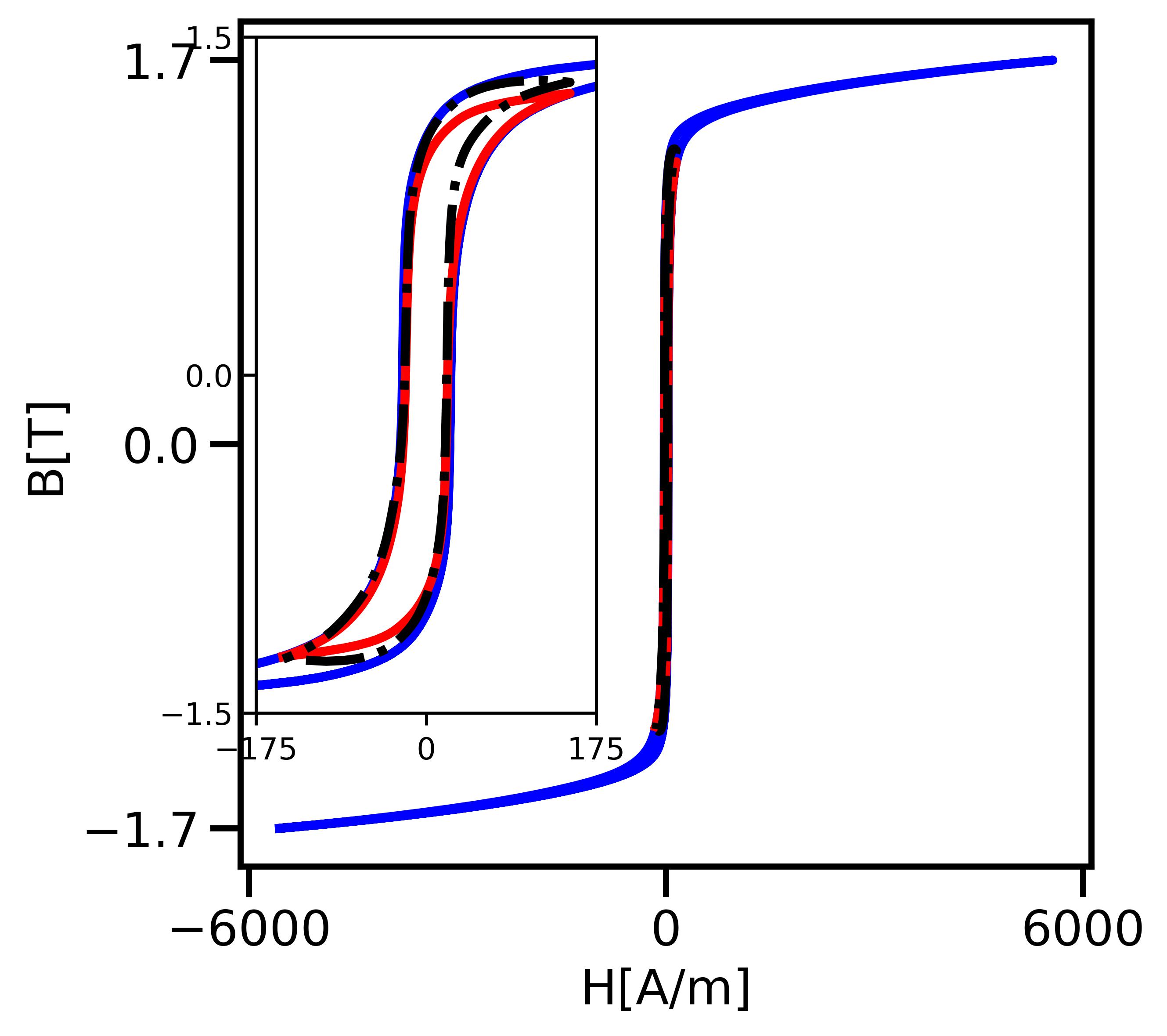}}} \\
      \subfigure[LSTM]{\label{3j}{\includegraphics[height=4.5cm, width=4.5cm]{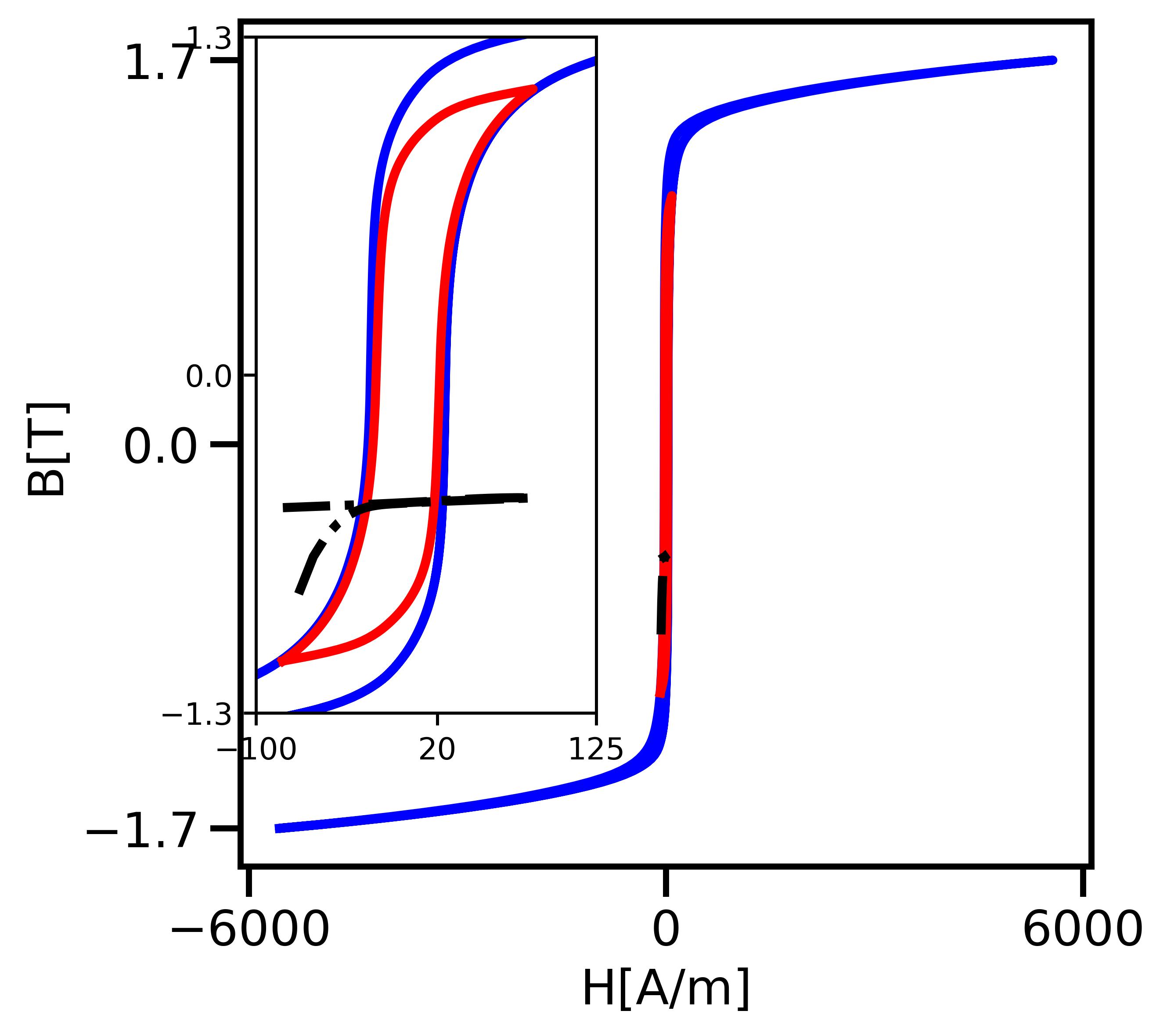}}}\hfill
      \subfigure[GRU]{\label{3k}{\includegraphics[height=4.5cm, width=4.5cm]{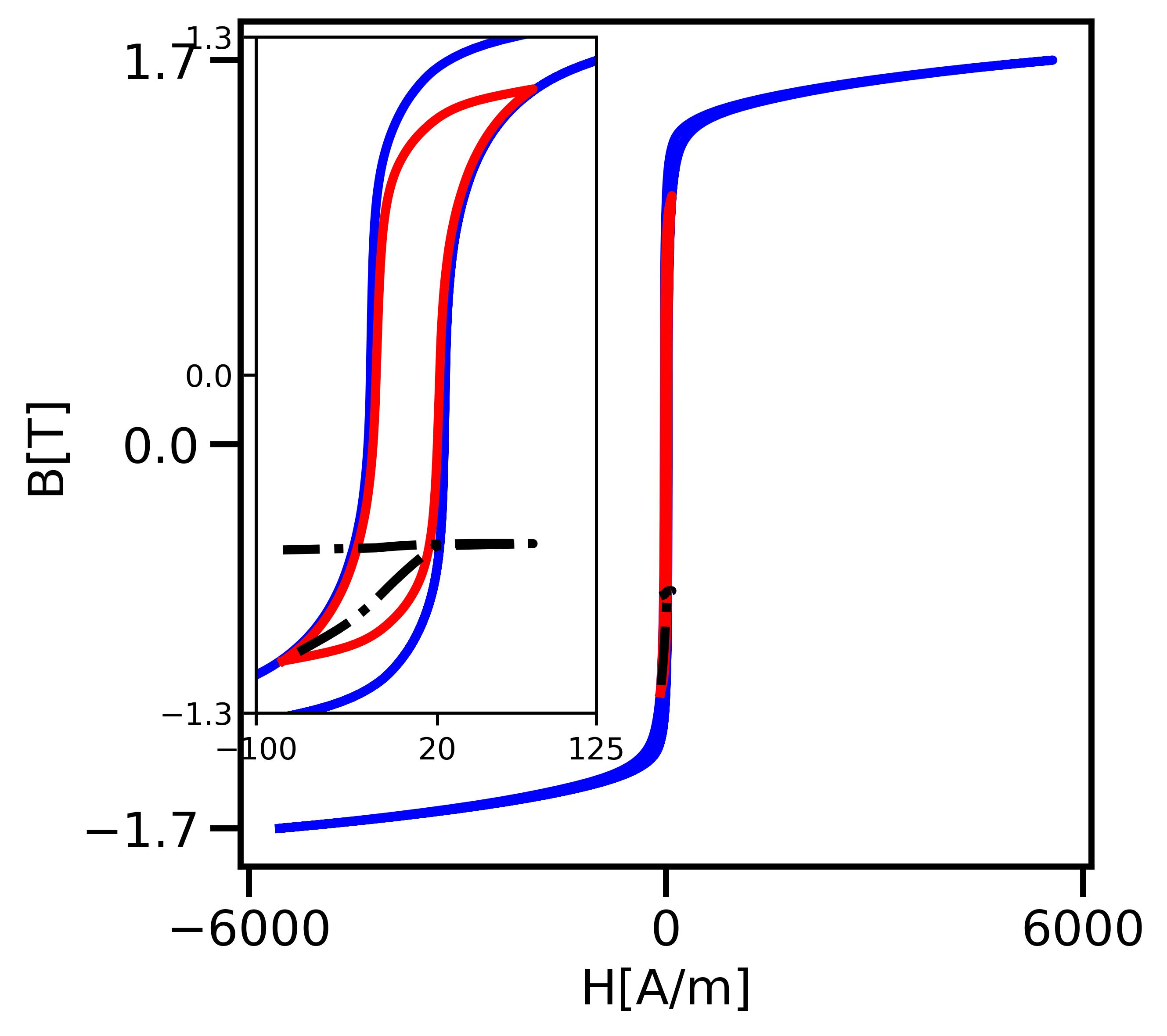}}} \hfill
      \subfigure[HystRNN]{\label{3l}{\includegraphics[height=4.5cm, width=4.5cm]{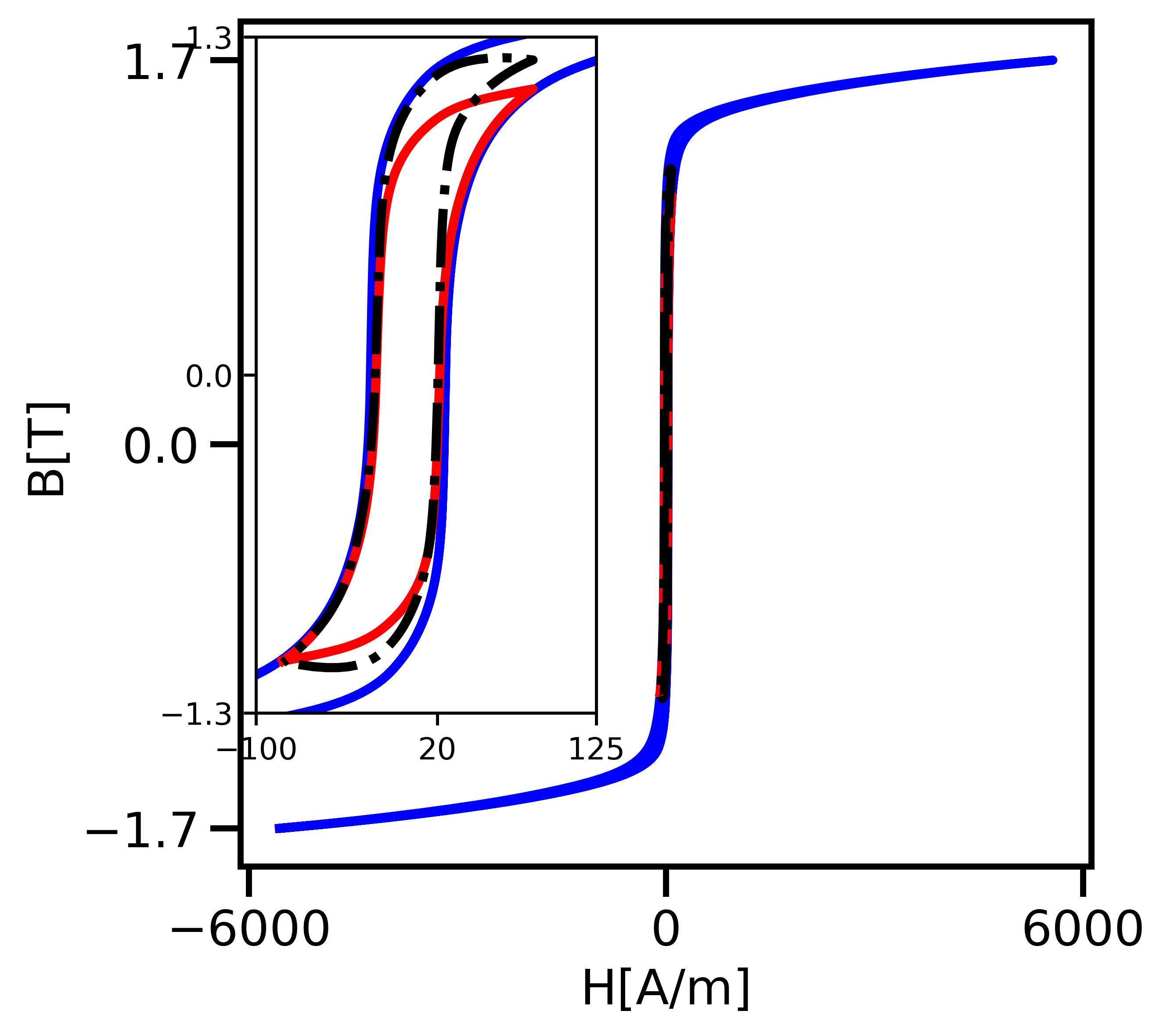}}} \hfill
    \caption{Experimental vs predicted hysteresis trajectories for experiment 1, where $\max(B)$ $=$ \SI{1.7}{\tesla}. The blue curve represents the training loop $\mathcal{C_\mathrm{major}}$. The red curve represents the ground truth for $\mathcal{C_\mathrm{FORC/minor}}$ and the black curve represents the prediction of the model. Top two rows: predictions for $\mathcal{C_\mathrm{FORC_1}}$ and $\mathcal{C_\mathrm{FORC_2}}$ respectively. Bottom two rows: predictions for $\mathcal{C_\mathrm{minor_1}}$ and $\mathcal{C_\mathrm{minor_2}}$ respectively. The colors are used consistently for the following figures.} 
    \label{fig3}
\end{figure}

\begin{table*}
\setlength{\tabcolsep}{2pt}
\centering
\scriptsize 
\caption{The generalization performance assessed using the metrics: L2-norm relative error, explained variance error, maximum error, and mean absolute error for the first experiment, where $\max(B)$ $=$ \SI{1.7}{\tesla}. For these metrics, higher (respectively, lower) values are favored for ($\uparrow$) (respectively, ($\downarrow$)). The implication of arrows remains consistent for all the following Tables.}\label{tbl:1}
\begin{tabular}{|c|c|c|c|c|c|c|c|c|c|c|c|c|c|c|c|c|}
\hline
\multirow{2}{*}{Test case} & \multicolumn{4}{c|}{L2-norm $(\downarrow)$} & \multicolumn{4}{c|}{Explained variance score $(\uparrow)$} & \multicolumn{4}{c|}{ Max error $(\downarrow)$} & \multicolumn{4}{c|}{ Mean absolute error $(\downarrow)$} \\ \cline{2-17}
          &RNN & LSTM & GRU   & HystRNN & RNN & LSTM & GRU & HystRNN & RNN & LSTM & GRU & HystRNN & RNN & LSTM & GRU&HystRNN  \\ \hline
$\mathcal{C_\mathrm{FORC_1}}$  & 5.0204 & 0.8525 & 0.7764 &\textbf{0.2198} & -0.0721  &  0.1081 & 0.2007  & \textbf{0.8252}  & 5.3597 & 2.4089 & 2.0075 &\textbf{1.2030}  & 2.9888 &1.1967 &1.5550 &\textbf{0.6149} \\ \hline
$\mathcal{C_\mathrm{FORC_2}}$ & 6.4877  & 0.5255 &  0.4701& \textbf{0.3085} &  -0.2545 & 0.1875 & 0.2395 &\textbf{0.8844}  & 5.3484 & 1.8428 & 1.8177 & \textbf{1.2371} & 3.6038 & 0.9327 & 0.8723 &\textbf{0.7613}  \\ \hline
$\mathcal{C_\mathrm{minor_1}}$ & 5.3506  & 1.4382 &  1.8028 & \textbf{0.0438}  & -0.1013 & 0.0298 & 0.0776  & \textbf{0.9839} & 2.7641 & 1.7098 & 1.8925 & \textbf{0.3108} & 1.4877 & 0.7142  & 0.7797 & \textbf{0.1258} \\ \hline
$\mathcal{C_\mathrm{minor_2}}$ & 12.3671  &  1.5785&  2.0563 & \textbf{0.0786}  & -2.7046 & 0.0248 & 0.0673 & \textbf{0.9661} & 3.7491 &  1.5726& 1.7486 & \textbf{0.3630} & 1.9703 & 0.6544  & 0.7341 &\textbf{0.1450}  \\ \hline
\end{tabular}
\end{table*}

Oscillator networks play a pervasive role across natural and engineering systems like pendulums in classical mechanics, among other instances. A \emph{notable trend} is emerging where RNN architectures are constructed based on ODEs and dynamical systems \cite{chen2018neural, rubanova2019latent, chang2019antisymmetricrnn, rusch2021unicornn}. Our study is \emph{closely associated} with CoRNN, where the oscillation and damping factors are integrated into model construction. \emph{In contrast}, our approach \emph{incorporates hysteretic terms} into the model. Another recent study \cite{kapoor2023neural}, demonstrates employing neural oscillators to extend the applicability of physics-informed machine learning. Our work shares similarities with this study, as both works aim to \emph{generalize scientific machine} learning and seek to predict quantities of interest beyond the scope of the training data \emph{without} relying on retraining or transfer learning methodologies. However, our work aims to predict trajectories of the hysteresis dynamics, whereas \cite{kapoor2023neural} predicts the solutions of partial differential equations in a generalized domain.

\subsection{Motivation}
The hidden state update in HystRNN is \emph{motivated} by the \emph{differential models of hysteresis} \cite{bouc1967forced, wen1976method, oh2005semilinear} that describe the phenomenon, incorporating an \emph{absolute value function} to model the hysteretic nonlinearity. Examples of such phenomenological models include, but are not limited to, the Bouc-Wen and Duhem models, presented in \textbf{SM} \S \textbf{F}. These absolute valued components play a crucial role in capturing hysteretic characteristics. These terms allow the models to account for the different responses during magnetization and demagnetization, as well as the \emph{effects of history} on the behavior of the system. The inclusion of absolute valued terms enhances the ability of the model to capture the intricate dynamics of hysteresis and provides a more realistic representation of the observed phenomena.

\section{Numerical Experiments}

\begin{table*}
\setlength{\tabcolsep}{2pt}
\centering
\scriptsize 
\caption{The generalization performance assessed using the metrics: L2-norm relative error, explained variance error, maximum error, and mean absolute error for the second experiment, where $\max(B)$ $=$ \SI{1.25}{\tesla}.}\label{tbl:2}
\begin{tabular}{|c|c|c|c|c|c|c|c|c|c|c|c|c|c|c|c|c|}
\hline
\multirow{2}{*}{Test case} & \multicolumn{4}{c|}{L2-norm $(\downarrow)$} & \multicolumn{4}{c|}{Explained variance score $(\uparrow)$} & \multicolumn{4}{c|}{ Max error $(\downarrow)$} & \multicolumn{4}{c|}{ Mean absolute error $(\downarrow)$} \\ \cline{2-17}
          &RNN & LSTM & GRU   & HystRNN & RNN & LSTM & GRU & HystRNN & RNN & LSTM & GRU & HystRNN & RNN & LSTM & GRU&HystRNN  \\ \hline
$\mathcal{C_\mathrm{FORC_1}}$  &5.3234  & 1.2308  & 0.6863 &\textbf{0.0109} &  0.1625 & 0.3566  & 0.4230  & \textbf{0.9891}  &2.6134  & 1.4922 & 1.0994 &\textbf{0.2370}  & 1.6253 & 0.6871& 0.5433 &\textbf{0.0563} \\ \hline
$\mathcal{C_\mathrm{FORC_2}}$ & 6.4035  &0.9569  & 0.5098 & \textbf{0.0115} & 0.1576 & 0.3834 & 0.4693 &\textbf{0.9924}  & 2.6084 & 1.2985 & 0.9055 & \textbf{0.2520} & 1.7484 & 0.5796 & 0.4587 &\textbf{0.0583}  \\ \hline
$\mathcal{C_\mathrm{minor_1}}$ & 7.5971  & 1.6295 & 0.8069  & \textbf{0.0320}  & 0.1497 & 0.2977 & 0.3268  & \textbf{0.9844} & 2.3788 & 1.3344 & 0.9456 & \textbf{0.1882} & 1.4996 &  0.5974 & 0.4396 & \textbf{0.0916} \\ \hline
$\mathcal{C_\mathrm{minor_2}}$ & 11.8293  & 2.1787 & 0.8859  & \textbf{0.1283}  & 0.0616 & 0.2769  & 0.3091 & \textbf{0.9267} & 2.2669 & 1.1784 & 0.7925 & \textbf{0.2923} &  1.5243&  0.5639 & 0.3653 &\textbf{0.1443}  \\ \hline
\end{tabular}
\end{table*}

\begin{table*}
\setlength{\tabcolsep}{2pt}
\centering
\scriptsize 
\caption{The generalization performance assessed using the metrics: L2-norm relative error, explained variance error, maximum error, and mean absolute error for the third experiment, where $\max(B)$ $=$ \SI{1.3}{\tesla}.}\label{tbl:3}
\begin{tabular}{|c|c|c|c|c|c|c|c|c|c|c|c|c|c|c|c|c|}
\hline
\multirow{2}{*}{Test case} & \multicolumn{4}{c|}{L2-norm $(\downarrow)$} & \multicolumn{4}{c|}{Explained variance score $(\uparrow)$} & \multicolumn{4}{c|}{ Max error $(\downarrow)$} & \multicolumn{4}{c|}{ Mean absolute error $(\downarrow)$} \\ \cline{2-17}
          &RNN & LSTM & GRU   & HystRNN & RNN & LSTM & GRU & HystRNN & RNN & LSTM & GRU & HystRNN & RNN & LSTM & GRU&HystRNN  \\ \hline
$\mathcal{C_\mathrm{FORC_1}}$  & 6.0907 &0.9692   & 0.6625  &\textbf{0.0342} & 0.0534  & 0.3290  & 0.3649  & \textbf{0.9705}  & 3.2184 & 1.5661 & 1.2194 &\textbf{0.3878}  & 1.9992 & 0.7105  & 0.6247 &\textbf{0.1296} \\ \hline
$\mathcal{C_\mathrm{FORC_2}}$ & 7.2661  & 0.6720 & 0.4775 & \textbf{0.0377} & 0.0295 & 0.3691 & 0.4183 &\textbf{0.9785}  & 3.2126 & 1.2972 & 0.9456 & \textbf{0.4055} & 2.1660 & 0.5806 & 0.5249 &\textbf{0.1371}  \\ \hline
$\mathcal{C_\mathrm{minor_1}}$ & 10.2305  & 1.6042 & 0.9009  & \textbf{0.0301}  & -0.0216 & 0.2699 &  0.2909 & \textbf{0.9774} & 2.7703 & 1.3222 &  0.9947& \textbf{0.1857} & 1.7520 & 0.5923  & 0.4619 & \textbf{0.0855} \\ \hline
$\mathcal{C_\mathrm{minor_2}}$ & 18.2528  &  2.3069& 1.0498  & \textbf{0.1580}  & -0.1629 & 0.2528  & 0.2785 & \textbf{0.8780} & 2.5696 & 1.1249& 0.8045 & \textbf{0.3066} & 1.7901 & 0.5446  & 0.3673 &\textbf{0.1491}  \\ \hline
\end{tabular}
\end{table*}

\begin{table*}
\setlength{\tabcolsep}{2pt}
\centering
\scriptsize 
\caption{The generalization performance assessed using the metrics: L2-norm relative error, explained variance error, maximum error, and mean absolute error for the fourth experiment, where $\max(B)$ $=$ \SI{1.5}{\tesla}.}\label{tbl:4}
\begin{tabular}{|c|c|c|c|c|c|c|c|c|c|c|c|c|c|c|c|c|}
\hline
\multirow{2}{*}{Test case} & \multicolumn{4}{c|}{L2-norm $(\downarrow)$} & \multicolumn{4}{c|}{Explained variance score $(\uparrow)$} & \multicolumn{4}{c|}{ Max error $(\downarrow)$} & \multicolumn{4}{c|}{ Mean absolute error $(\downarrow)$} \\ \cline{2-17}
          &RNN & LSTM & GRU   & HystRNN & RNN & LSTM & GRU & HystRNN & RNN & LSTM & GRU & HystRNN & RNN & LSTM & GRU&HystRNN  \\ \hline
$\mathcal{C_\mathrm{FORC_1}}$  & 0.7673  & 0.8556  & 0.9330 &\textbf{0.1007} & 0.2957  &  0.1675 &  0.3407 & \textbf{0.9017}  & 1.7979 & 1.9572 & 1.8971 &\textbf{0.7491}  & 0.9147  & 0.9642 & 0.9957 &\textbf{0.3258} \\ \hline
$\mathcal{C_\mathrm{FORC_2}}$ & 0.4968  & 0.5942 & 0.5268 & \textbf{0.1164} & 0.3331 & 0.2209 & 0.3718 &\textbf{0.9330}  & 1.2239 & 1.3492 & 1.3040 & \textbf{0.7725} & 0.7094 & 0.7686 & 0.7253 &\textbf{0.3304}  \\ \hline
$\mathcal{C_\mathrm{minor_1}}$ & 1.3719  & 1.3723 &  4.0550 & \textbf{0.0784}  & 0.2318 & 0.0903 & 0.0640  & \textbf{0.9224} & 1.1112 & 1.1767 &  1.7688& \textbf{0.2394} & 0.4981  &  0.5022 & 0.9177 & \textbf{0.1306} \\ \hline
$\mathcal{C_\mathrm{minor_2}}$ &  1.7923 & 1.6721 &  6.0749 & \textbf{0.2257}  & 0.2291 & 0.0907  & 0.0157 & \textbf{0.7743} & 0.9550 & 0.9988& 1.5820 & \textbf{0.2991} & 0.4396 &  0.4261 & 0.911 &\textbf{0.1717}  \\ \hline
\end{tabular}
\end{table*}

A series of numerical experiments encompassing \emph{four distinct scenarios} is conducted, in which we systematically vary the upper limit of the magnetic field $B$. The selection of diverse maximum $B$ field values corresponds to the specific usage context of the material. As a result, these experiments are geared towards demonstrating the viability of the proposed methodology across a spectrum of electrical machines, all constrained by their respective permissible maximum $B$ values.

\begin{figure}
    \centering
    \subfigure[LSTM]{\label{4a}{\includegraphics[height=4.5cm, width=4.5cm]{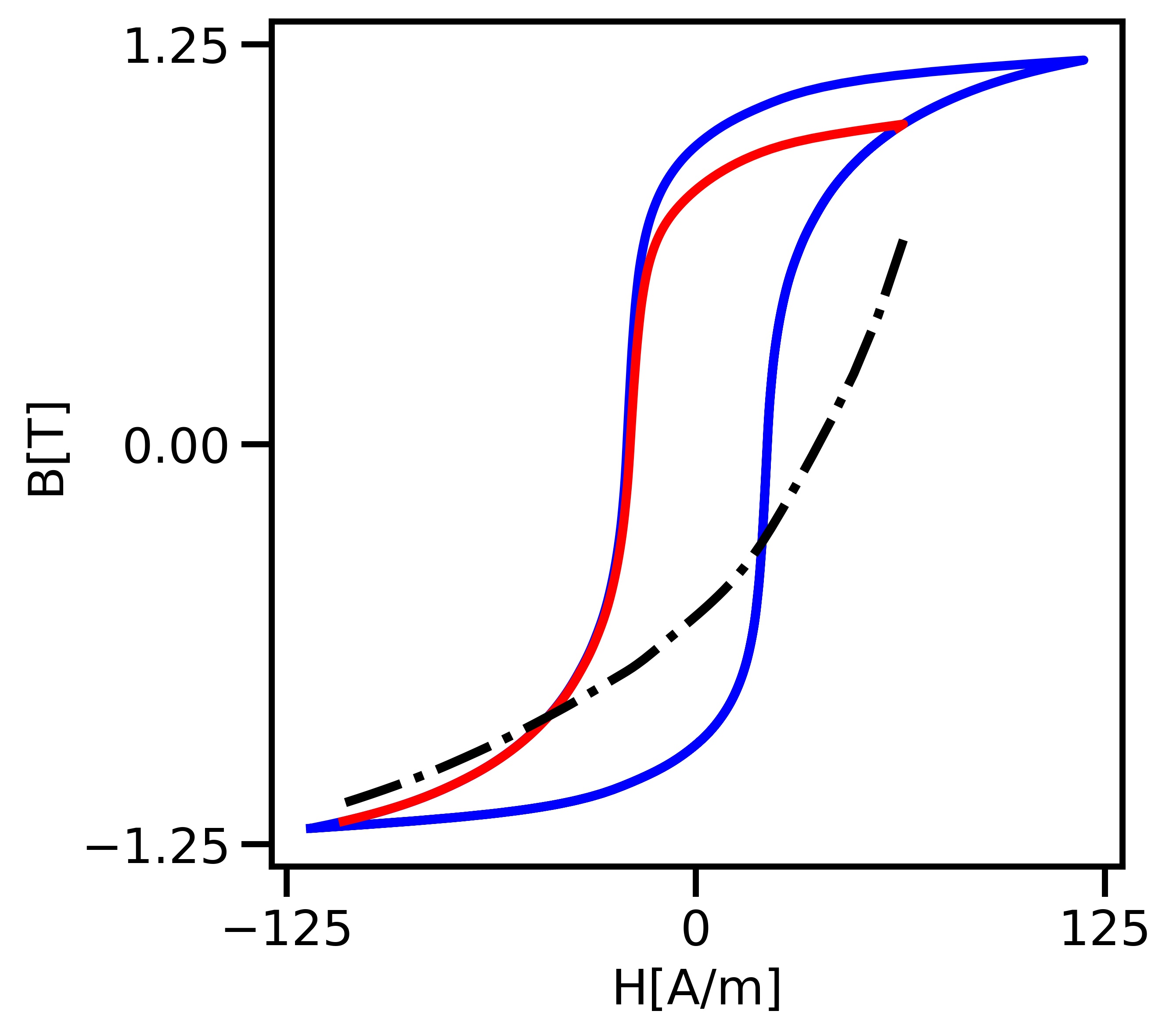}}}\hfill
      \subfigure[GRU]{\label{4b}{\includegraphics[height=4.5cm, width=4.5cm]{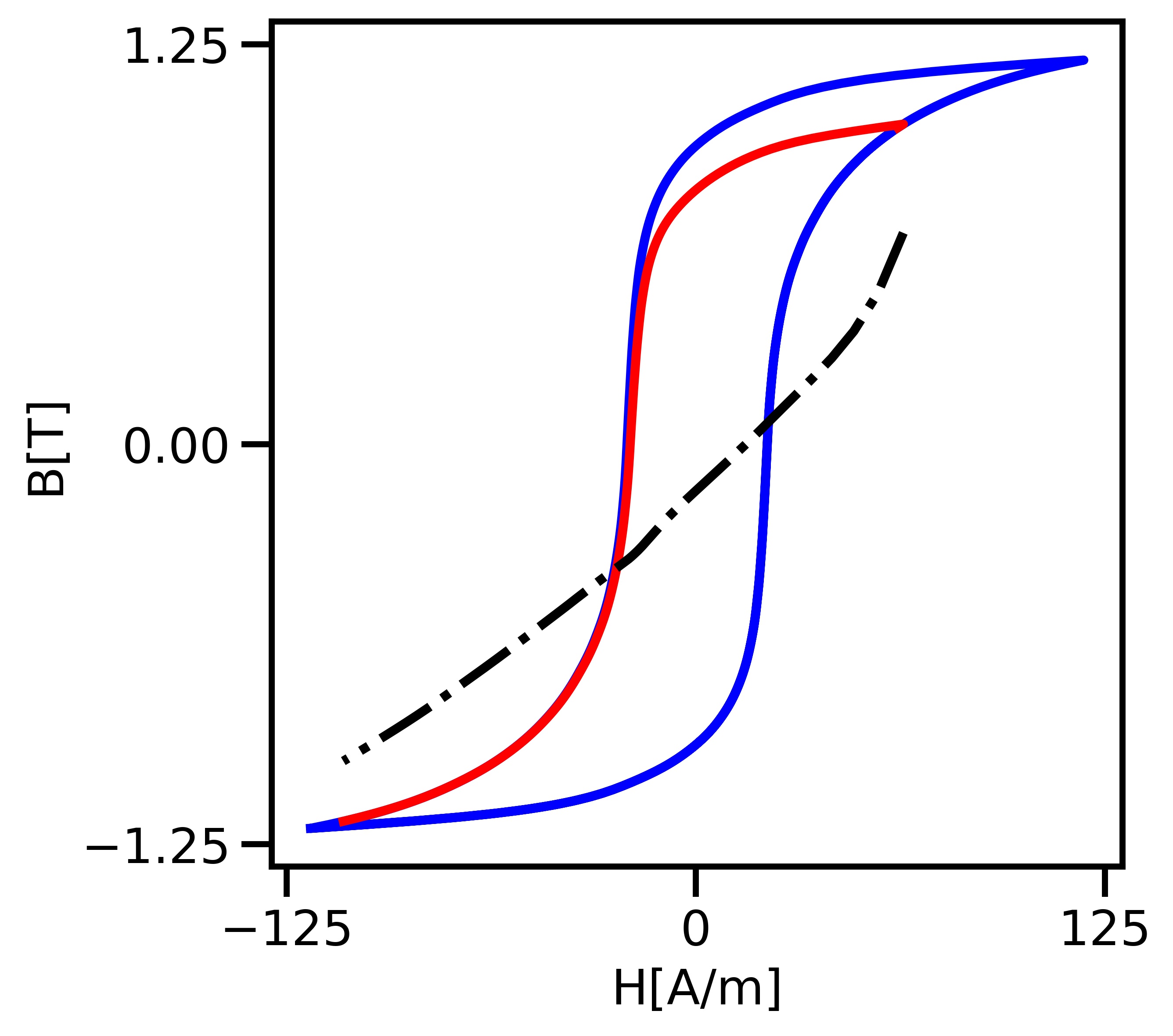}}} \hfill
      \subfigure[HystRNN]{\label{4c}{\includegraphics[height=4.5cm, width=4.5cm]{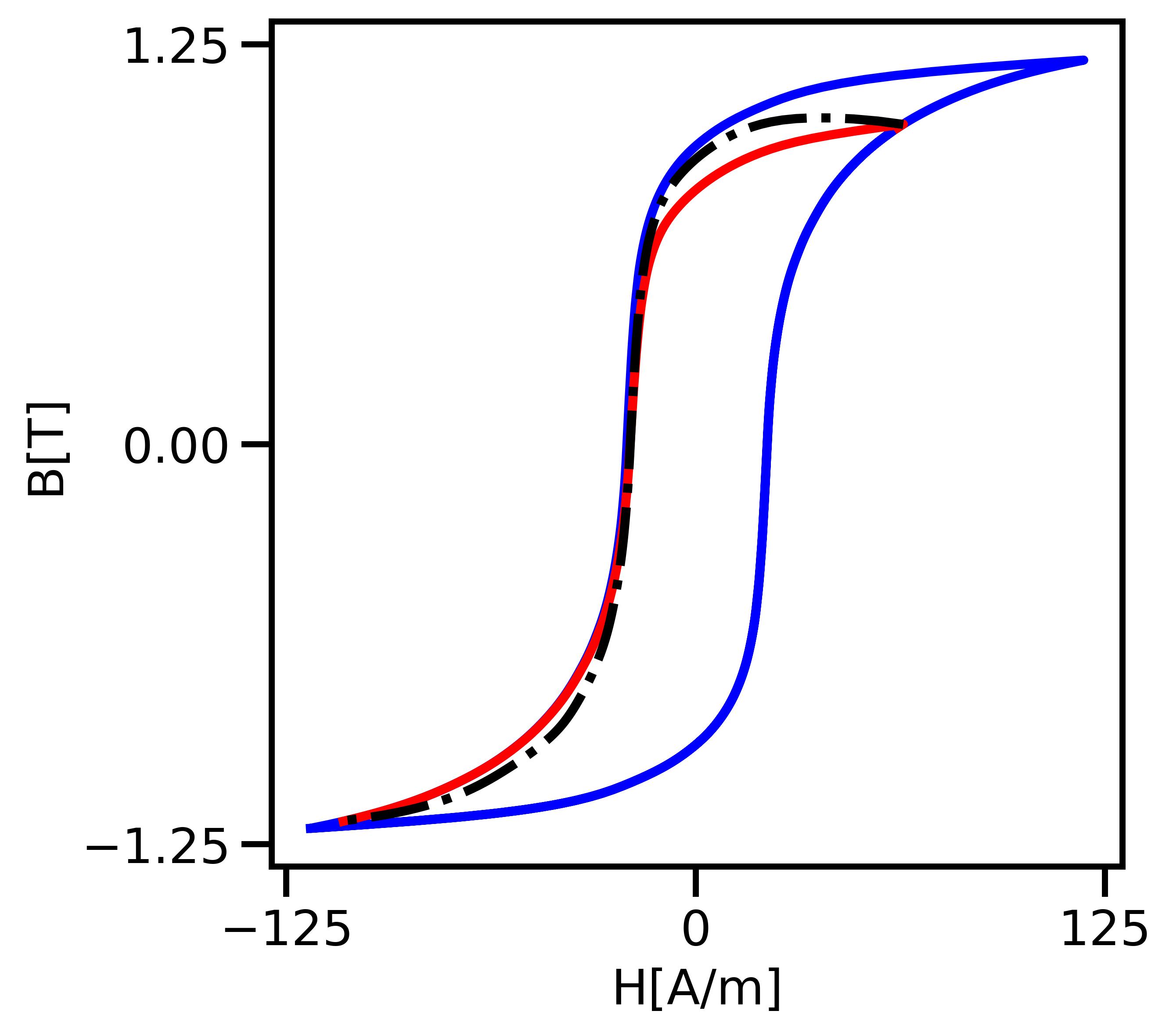}}} \\
      \subfigure[LSTM]{\label{4d}{\includegraphics[height=4.5cm, width=4.5cm]{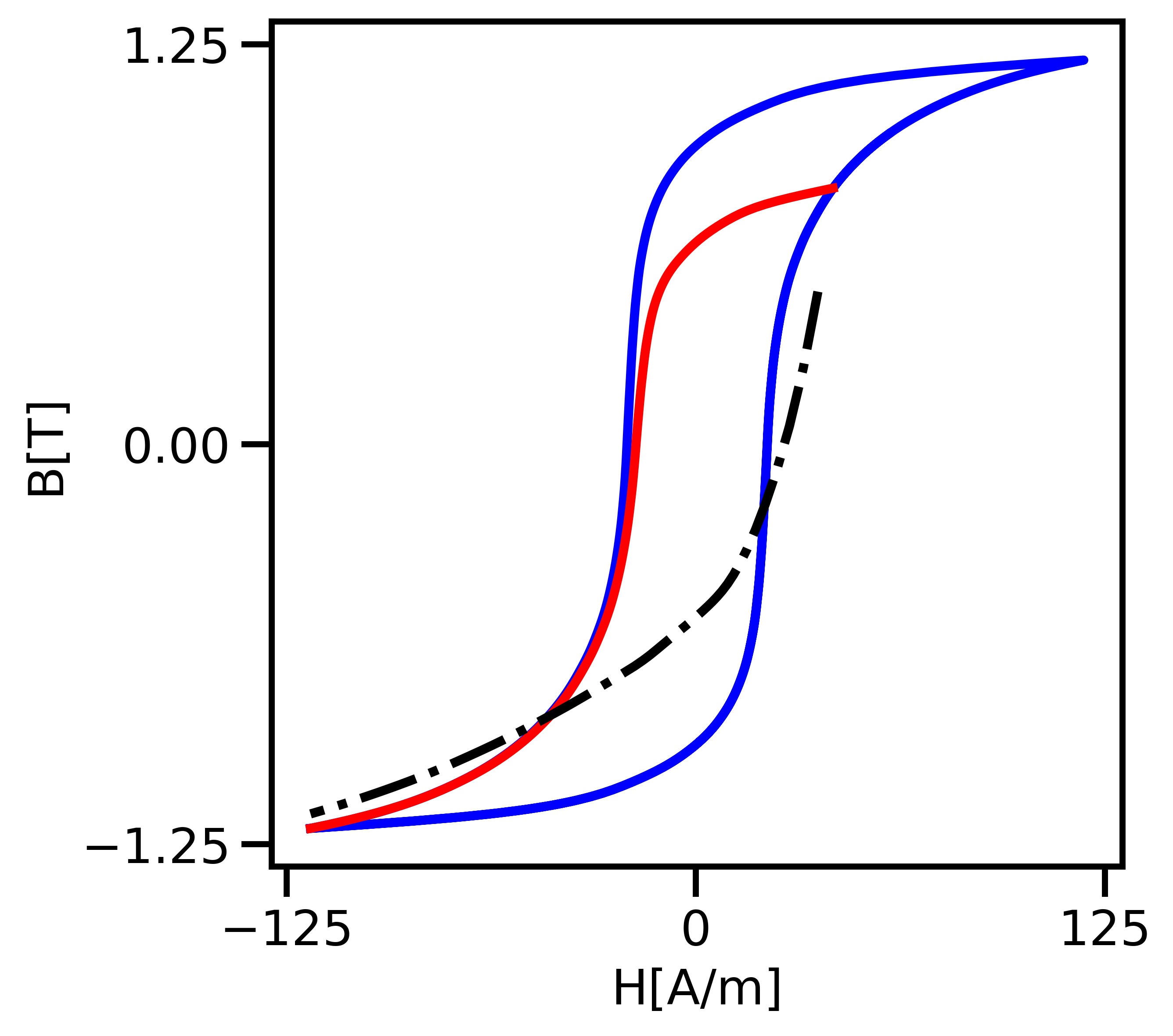}}}\hfill
      \subfigure[GRU]{\label{4e}{\includegraphics[height=4.5cm, width=4.5cm]{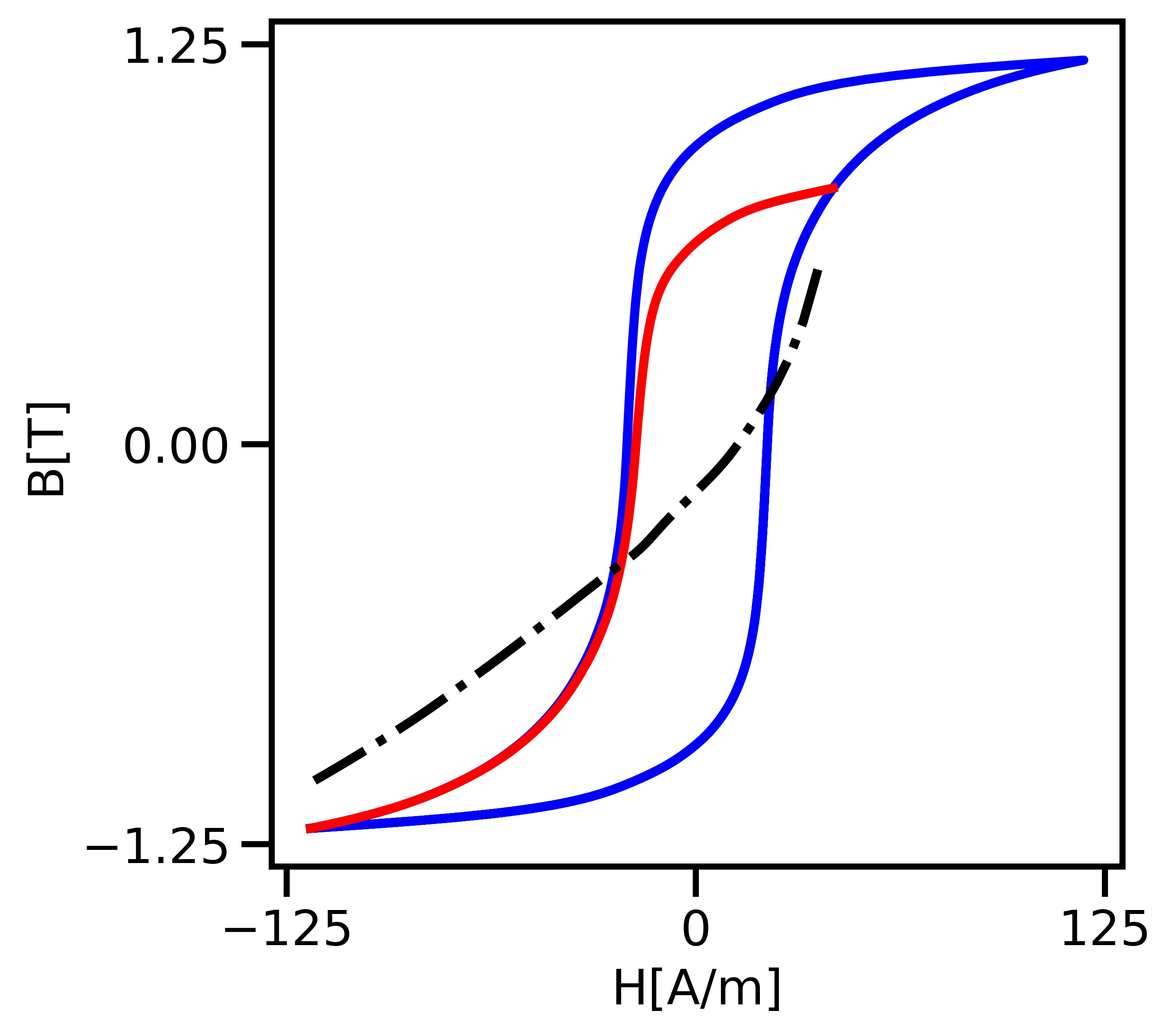}}} \hfill
      \subfigure[HystRNN]{\label{4f}{\includegraphics[height=4.5cm, width=4.5cm]{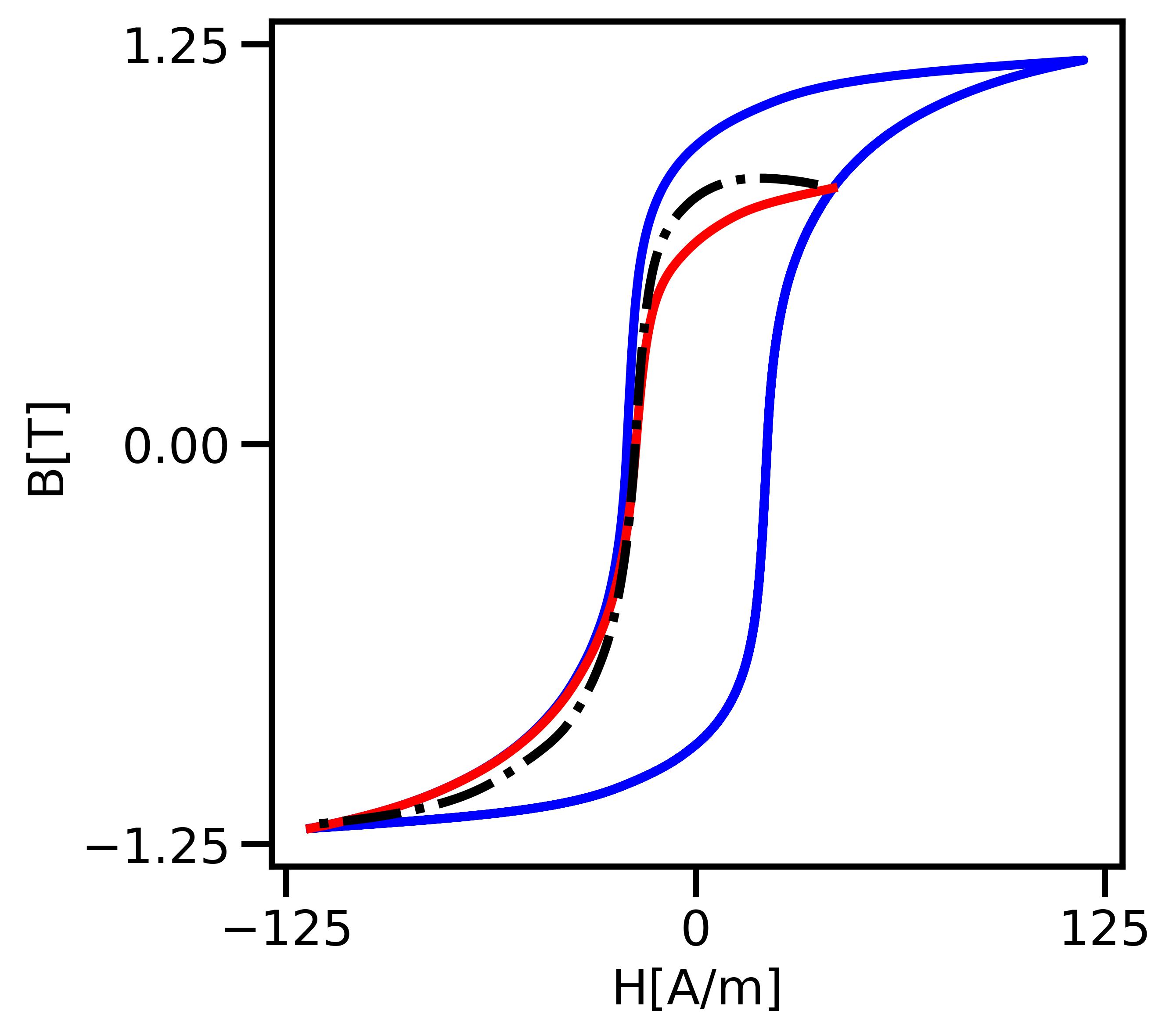}}} \\
      \subfigure[LSTM]{\label{4g}{\includegraphics[height=4.5cm, width=4.5cm]{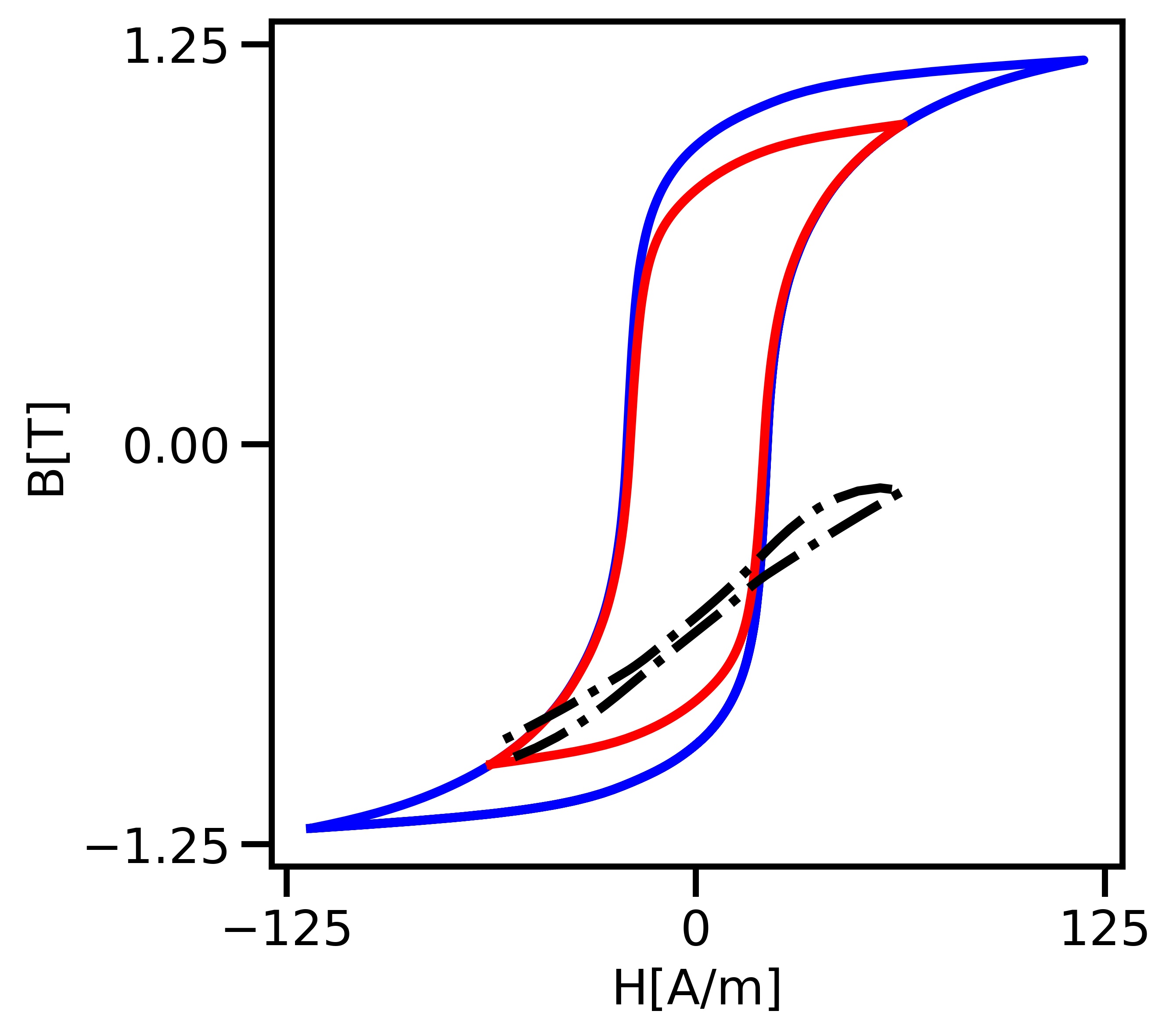}}}\hfill
      \subfigure[GRU]{\label{4h}{\includegraphics[height=4.5cm, width=4.5cm]{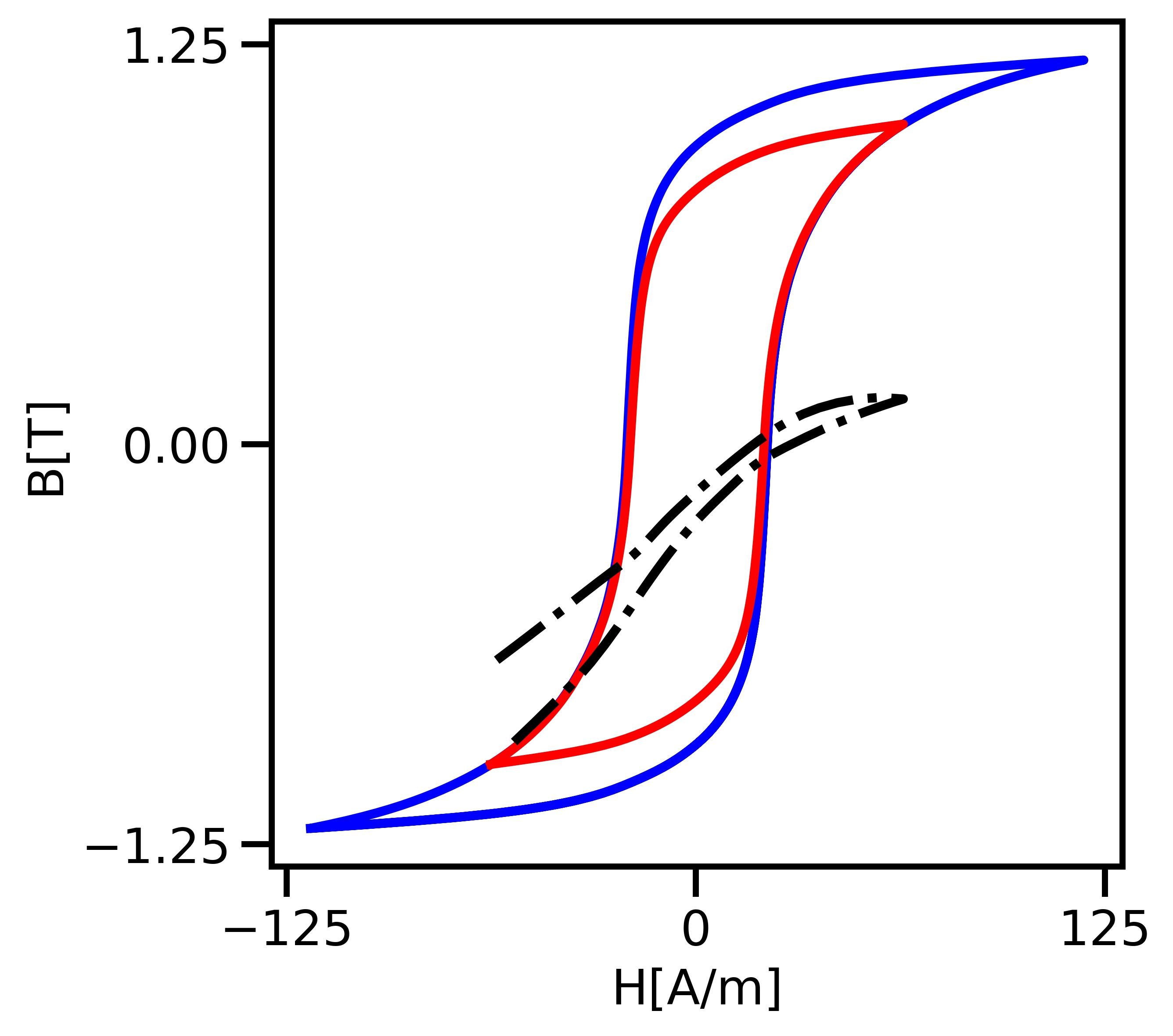}}} \hfill
      \subfigure[HystRNN]{\label{4i}{\includegraphics[height=4.5cm, width=4.5cm]{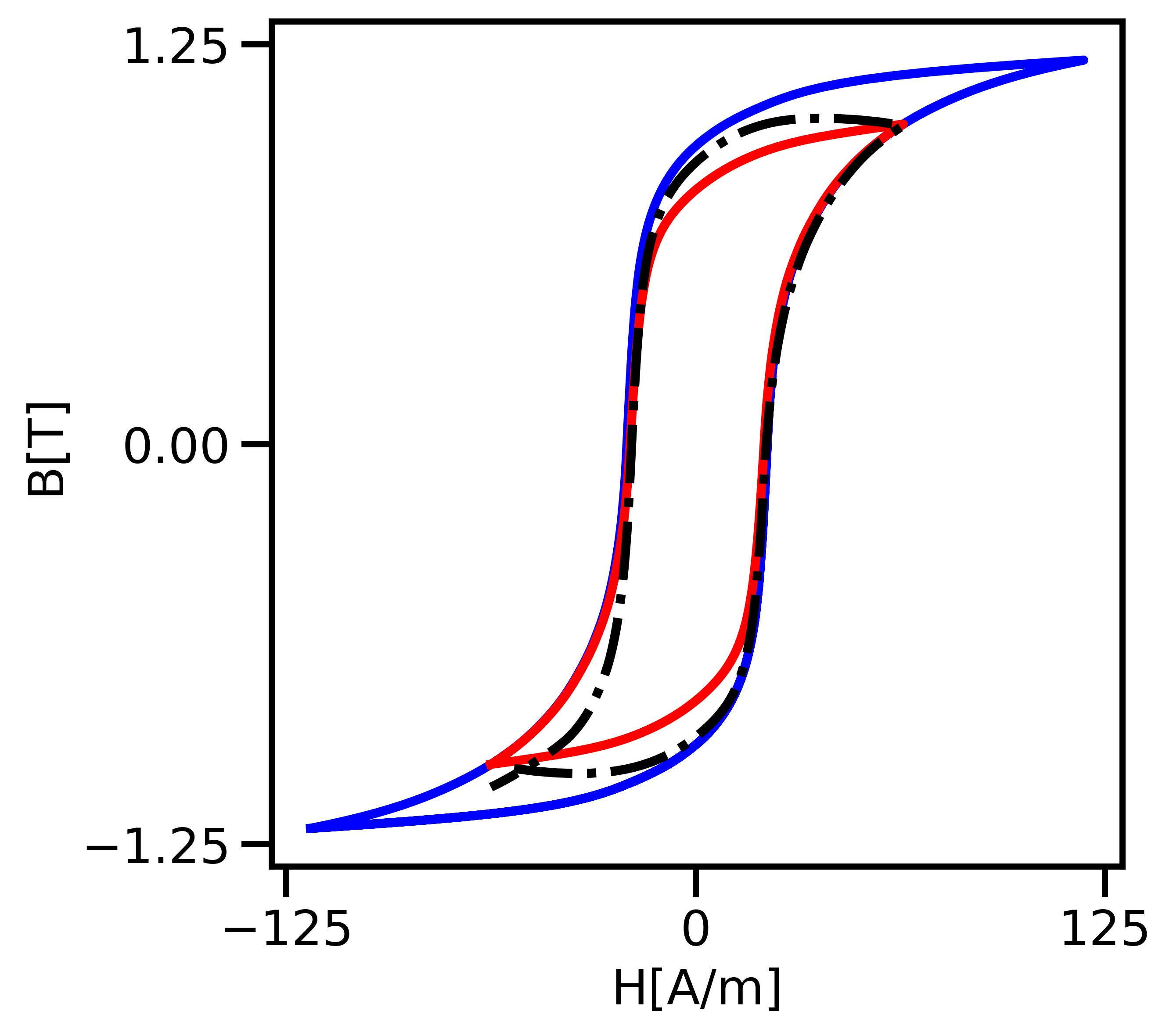}}} \\
      \subfigure[LSTM]{\label{4j}{\includegraphics[height=4.5cm, width=4.5cm]{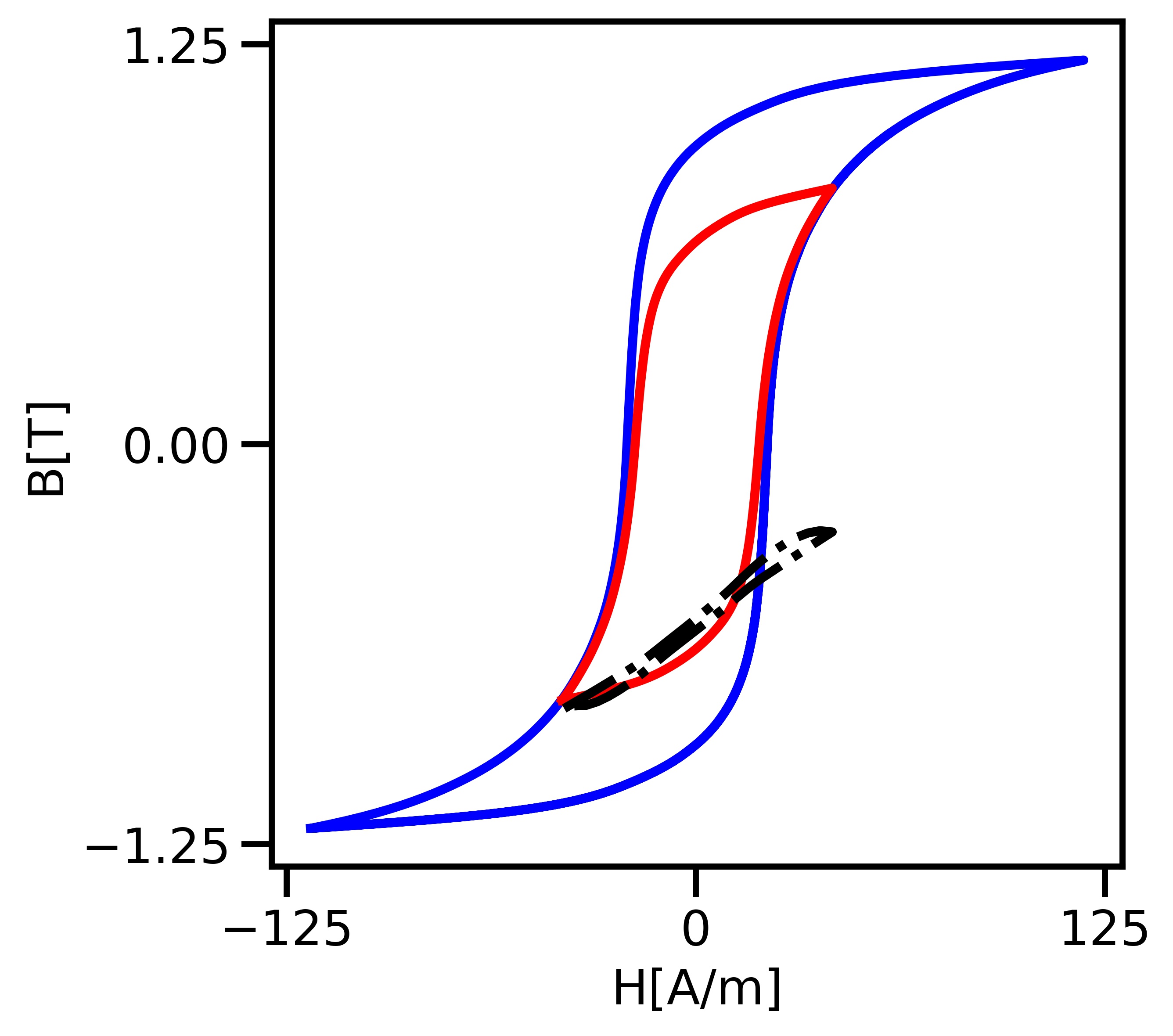}}}\hfill
      \subfigure[GRU]{\label{4k}{\includegraphics[height=4.5cm, width=4.5cm]{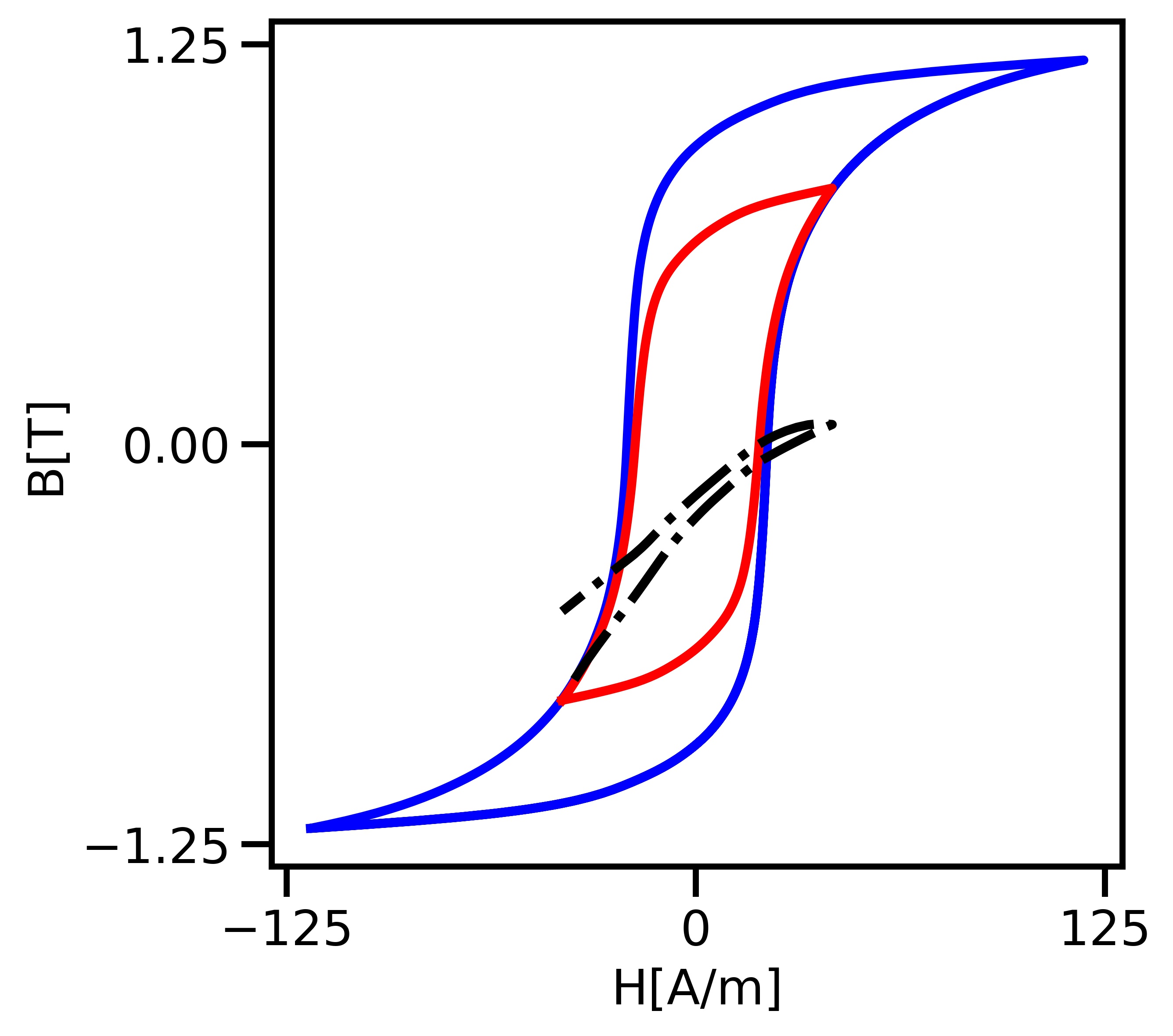}}} \hfill
      \subfigure[HystRNN]{\label{4l}{\includegraphics[height=4.5cm, width=4.5cm]{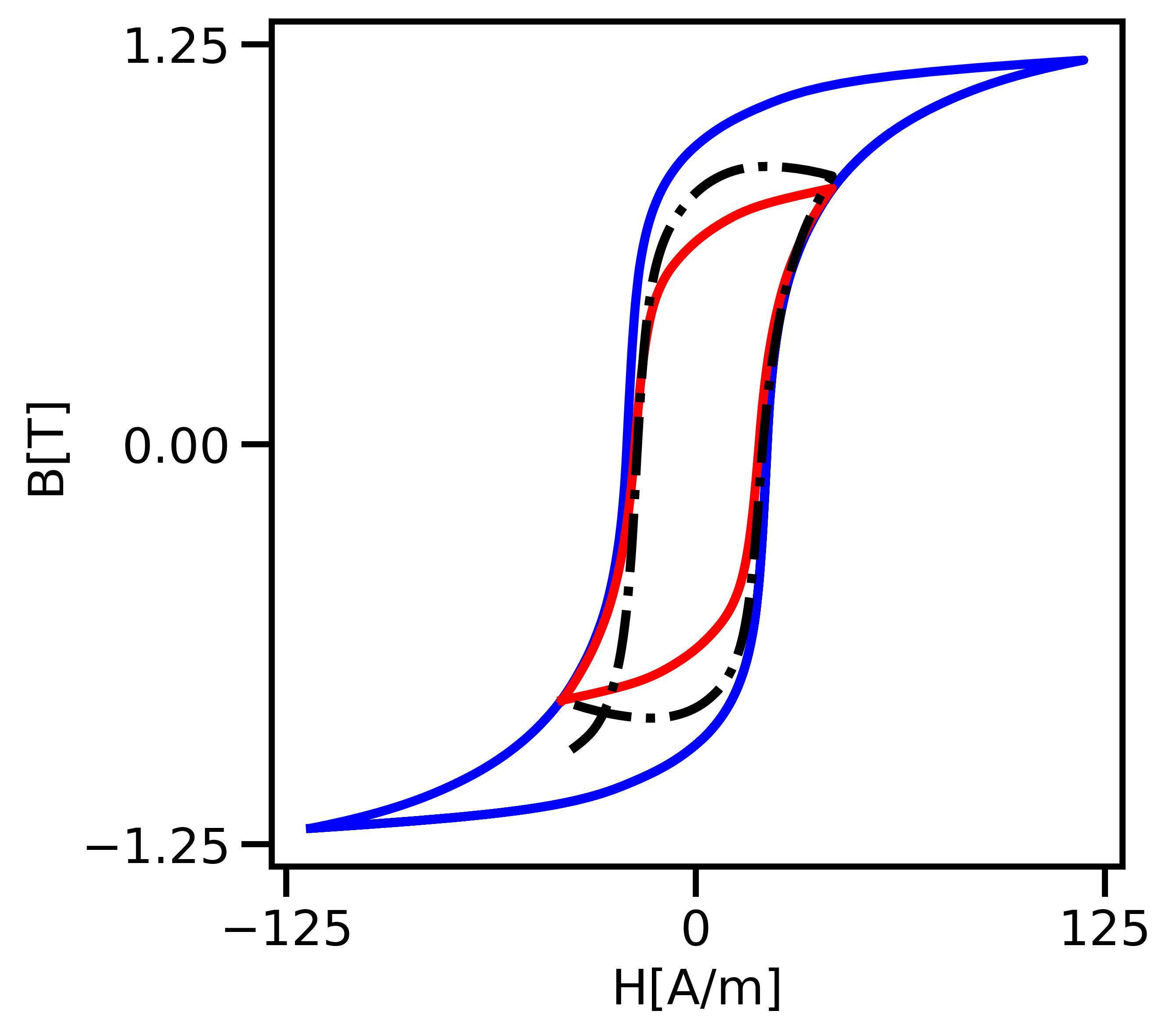}}} \hfill
    \caption{Experimental vs predicted hysteresis trajectories for experiment 2, where $\max(B)$ $=$ \SI{1.25}{\tesla}. Top two rows: predictions for $\mathcal{C_\mathrm{FORC_1}}$ and $\mathcal{C_\mathrm{FORC_2}}$ respectively. Bottom two rows: predictions for $\mathcal{C_\mathrm{minor_1}}$ and $\mathcal{C_\mathrm{minor_2}}$ respectively.} 
    \label{fig4}
\end{figure}

Precisely, we opt for the maximum $B$ values of \SI{1.7}{\tesla}, \SI{1.5}{\tesla}, \SI{1.3}{\tesla}, and \SI{1.25}{\tesla}. \emph{In each of these instances, we execute a total of four experiments}. After training for $\mathcal{C_\mathrm{major}}$ until the $B$ field reaches its saturation point, the experiments involve predicting $\mathcal{C_\mathrm{FORC}}$ and $\mathcal{C_\mathrm{minor}}$. Notably, the data set is generated from the Preisach model, which characterizes the behavior of non-oriented NO27 steel. It is imperative to preprocess the data before feeding it into the HystRNN model. We employ a normalization step using the min-max scaling technique, as elucidated in \textbf{SM} \S \textbf{C}.

For all the numerical experiments, the software and hardware environments used for performing the experiments are as follows: \textsc{Ubuntu} 20.04.6 LTS, \textsc{Python} 3.9.7, \textsc{Numpy} 1.20.3, \textsc{Scipy} 1.7.1, \textsc{Matplotlib} 3.4.3, \textsc{PyTorch} 1.12.1, \textsc{CUDA} 11.7, and \textsc{NVIDIA} Driver 515.105.01, i7 CPU, and \textsc{NVIDIA GeForce RTX 3080}.

\subsection{Baselines}
We introduce the concept of employing neural oscillators constructed within the framework of recurrent networks for modeling hysteresis. This choice is driven by the \emph{intrinsic sequentiality and memory dependence} characteristics of hysteresis. Additionally, recurrent architectures have been successfully employed for interpolation-type hysteresis modeling tasks across various fields \cite{chen2021diagonal, saghafifar2002dynamic}. Consequently, motivated by these rationales, we conduct a comparative analysis of HystRNN with traditional recurrent networks such as RNN, LSTM, and GRU. We subject these models to a comprehensive comparison, specifically focusing on their performance in the context of generalization. Our experiments involve delving into the potential of these models by predicting outcomes for untrained $H$ sequences, thereby exploring their capabilities for \emph{generalization beyond trained data}.

\begin{figure}[t]
    \centering
    \subfigure[LSTM]{\label{5a}{\includegraphics[height=4.5cm, width=4.5cm]{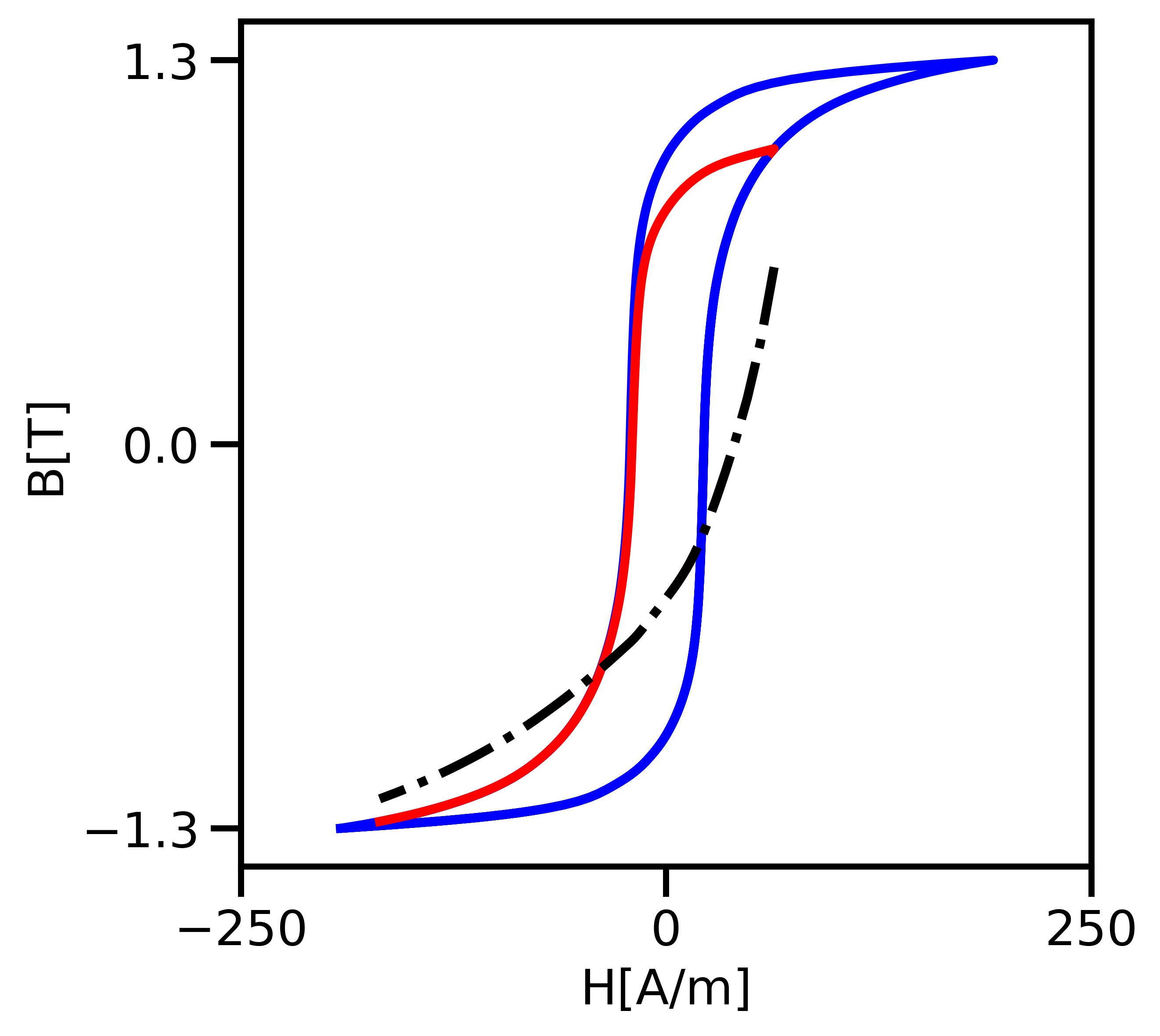}}}\hfill
      \subfigure[GRU]{\label{5b}{\includegraphics[height=4.5cm, width=4.5cm]{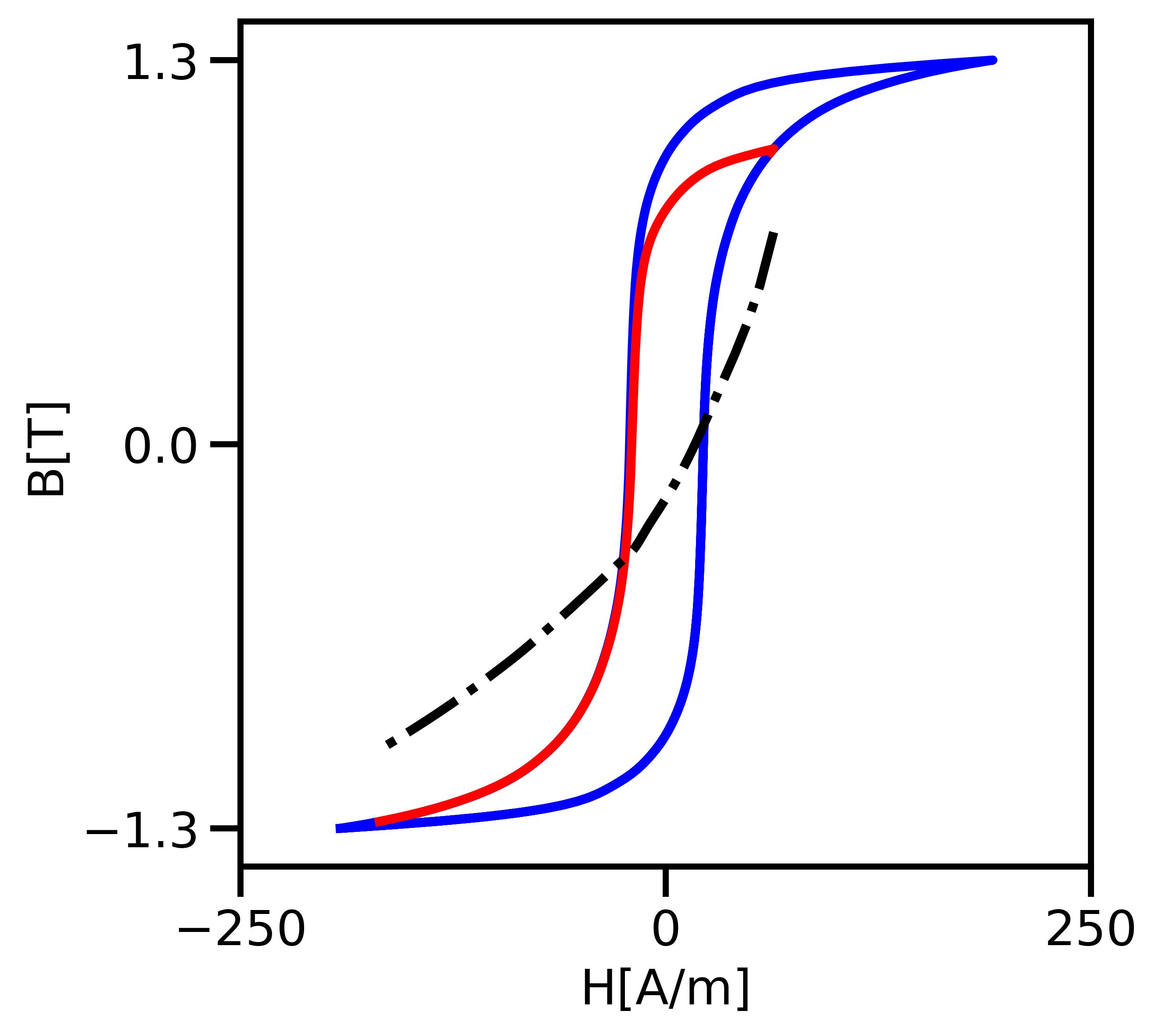}}} \hfill
      \subfigure[HystRNN]{\label{5c}{\includegraphics[height=4.5cm, width=4.5cm]{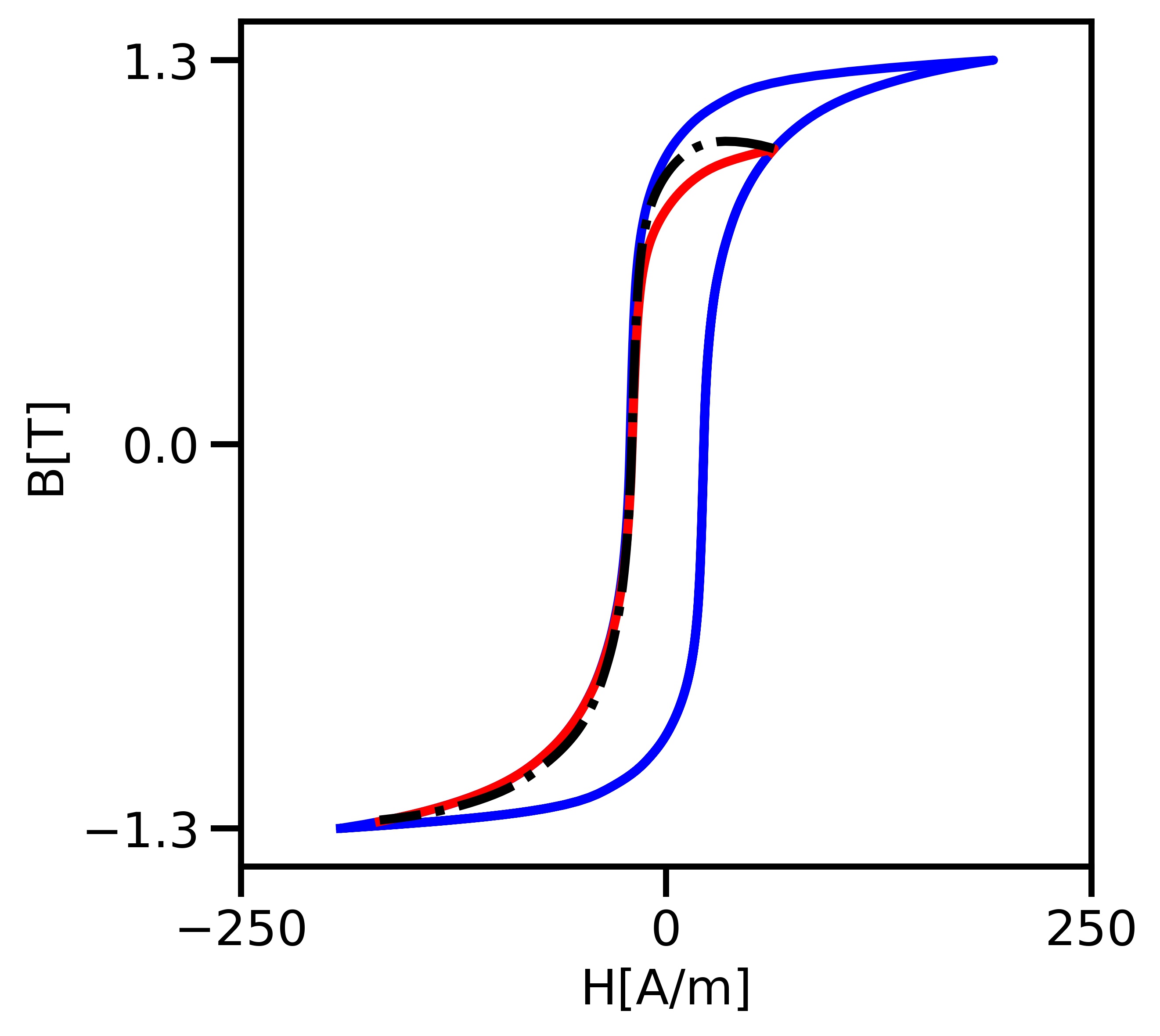}}} \\
      \subfigure[LSTM]{\label{5d}{\includegraphics[height=4.5cm, width=4.5cm]{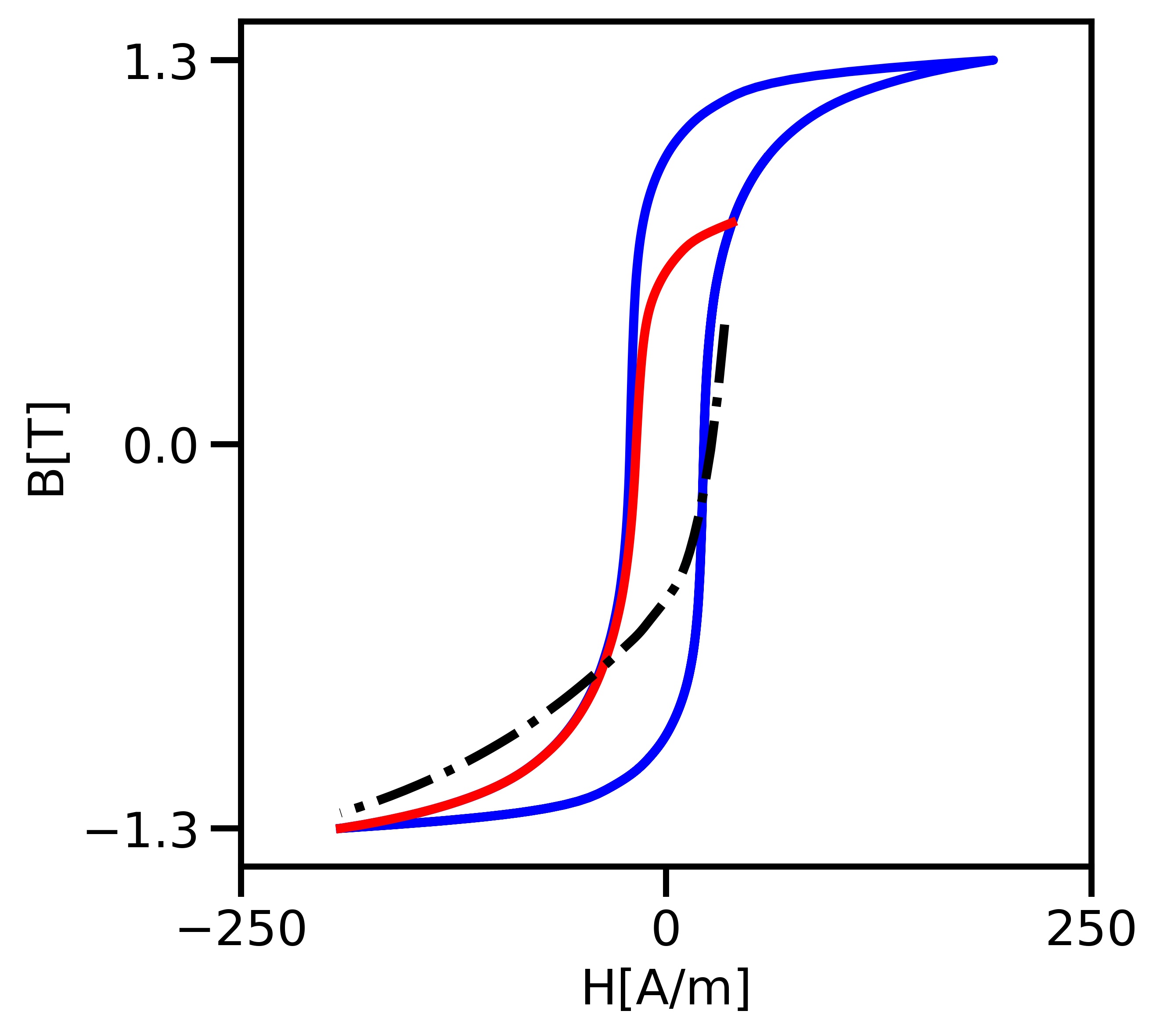}}}\hfill
      \subfigure[GRU]{\label{5e}{\includegraphics[height=4.5cm, width=4.5cm]{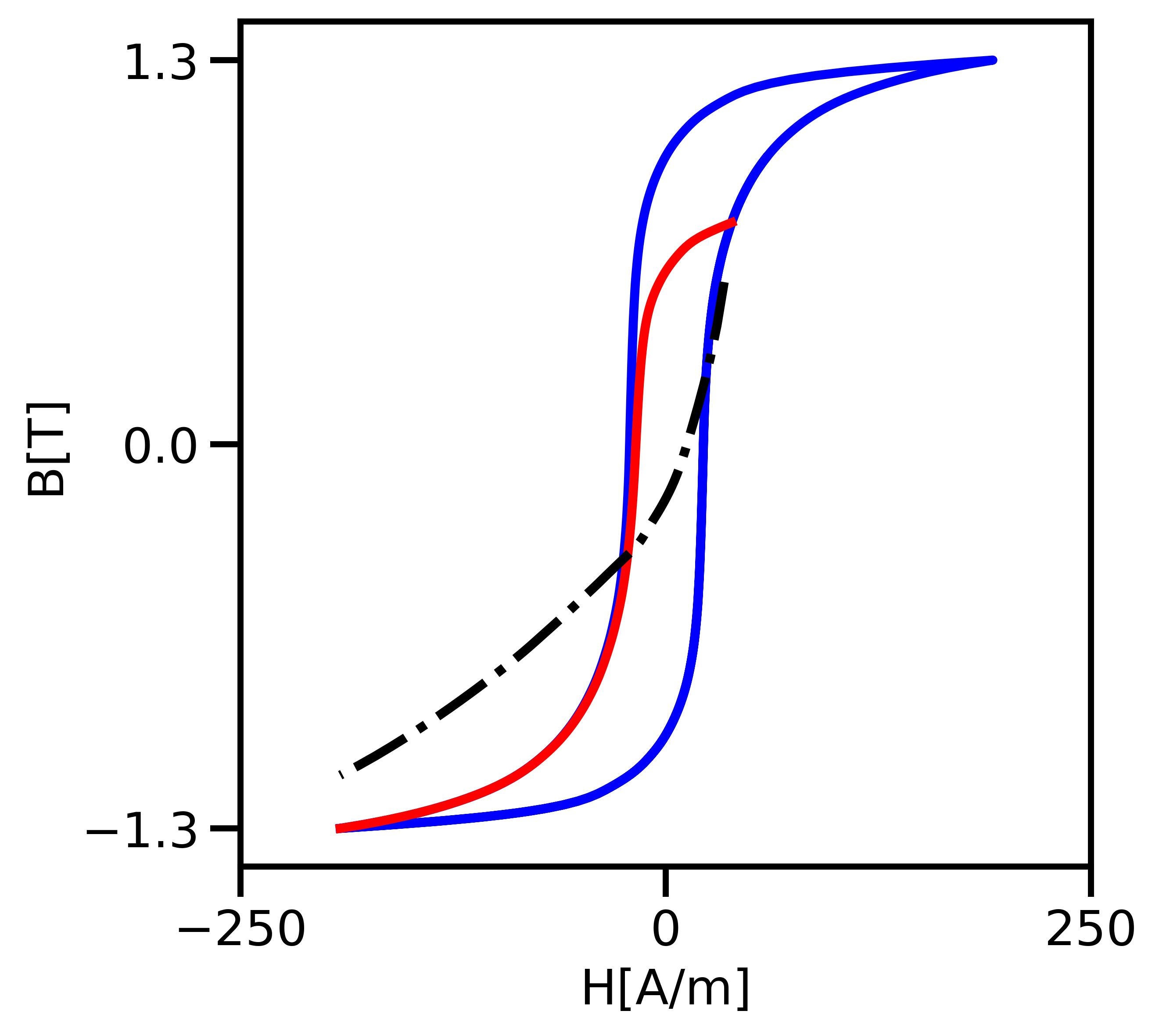}}} \hfill
      \subfigure[HystRNN]{\label{5f}{\includegraphics[height=4.5cm, width=4.5cm]{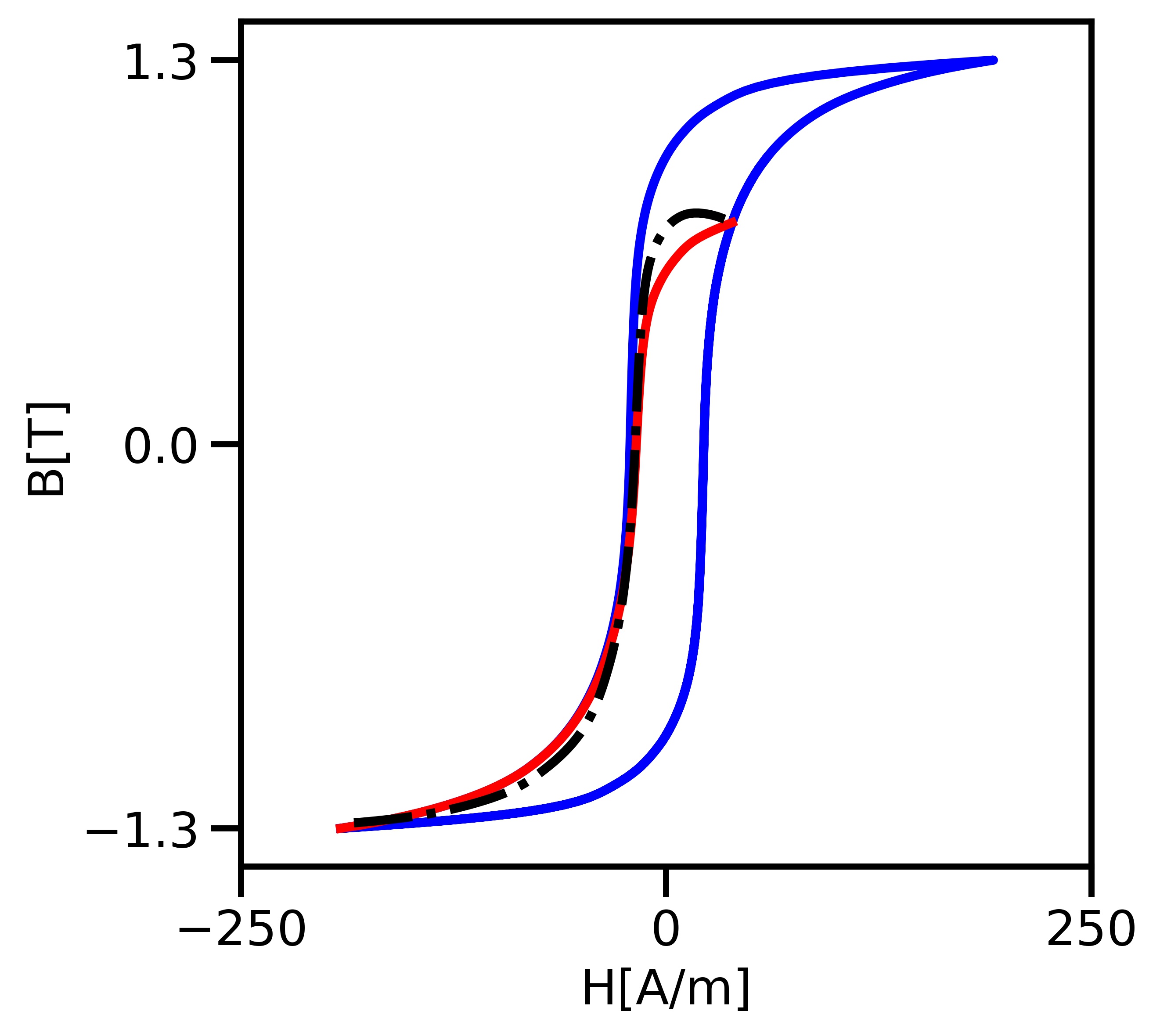}}} \\
      \subfigure[LSTM]{\label{5g}{\includegraphics[height=4.5cm, width=4.5cm]{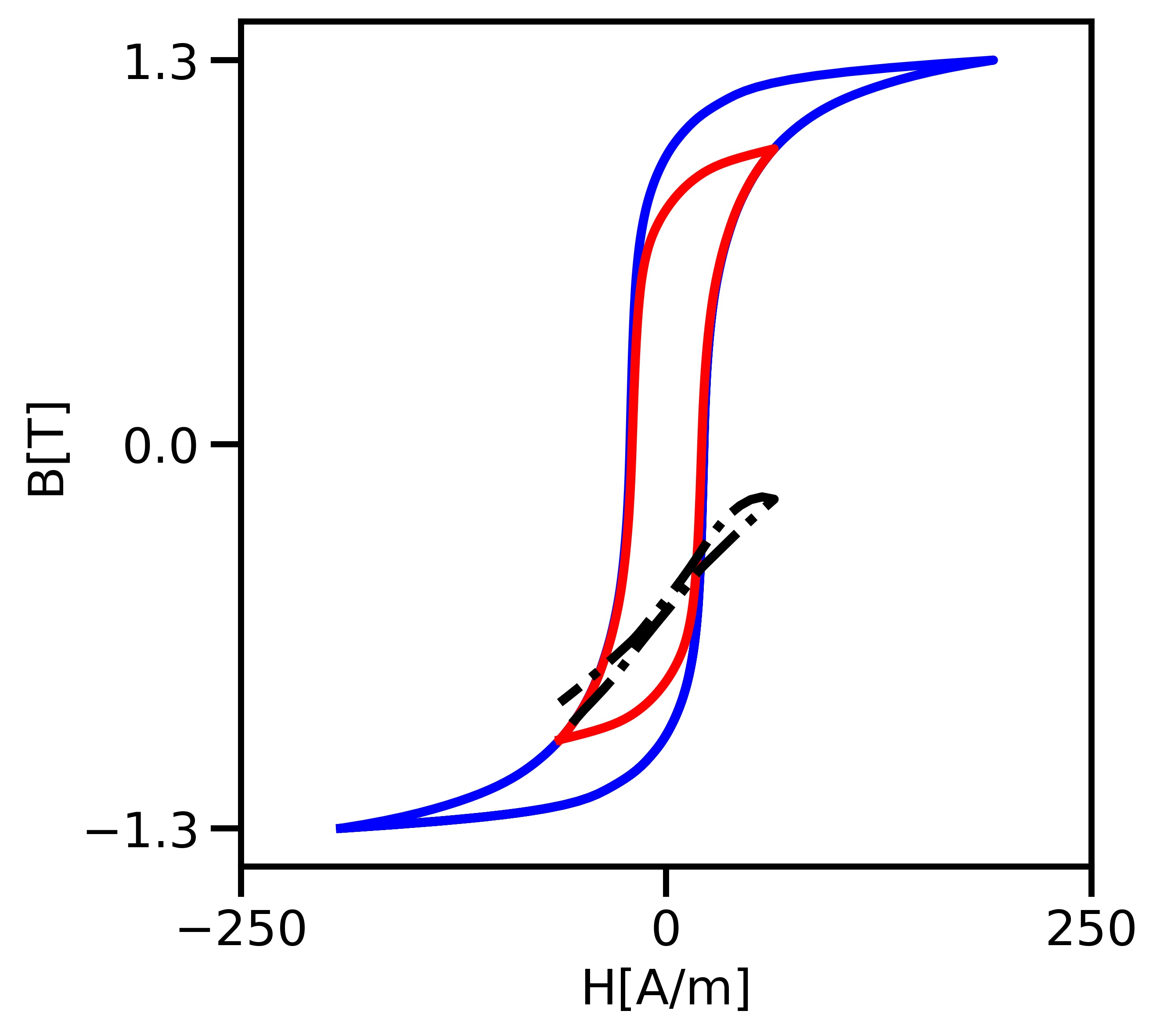}}}\hfill
      \subfigure[GRU]{\label{5h}{\includegraphics[height=4.5cm, width=4.5cm]{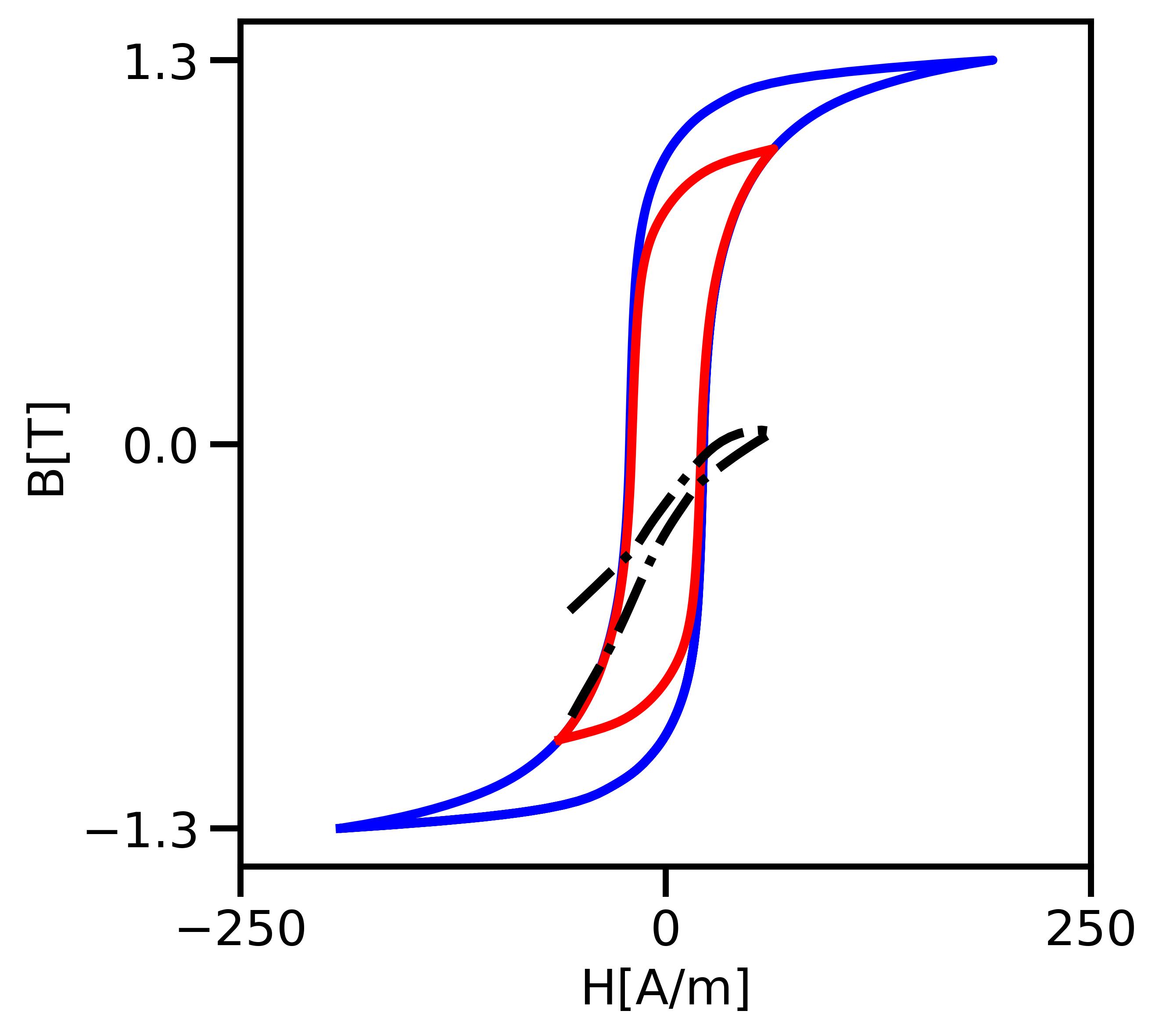}}} \hfill
      \subfigure[HystRNN]{\label{5i}{\includegraphics[height=4.5cm, width=4.5cm]{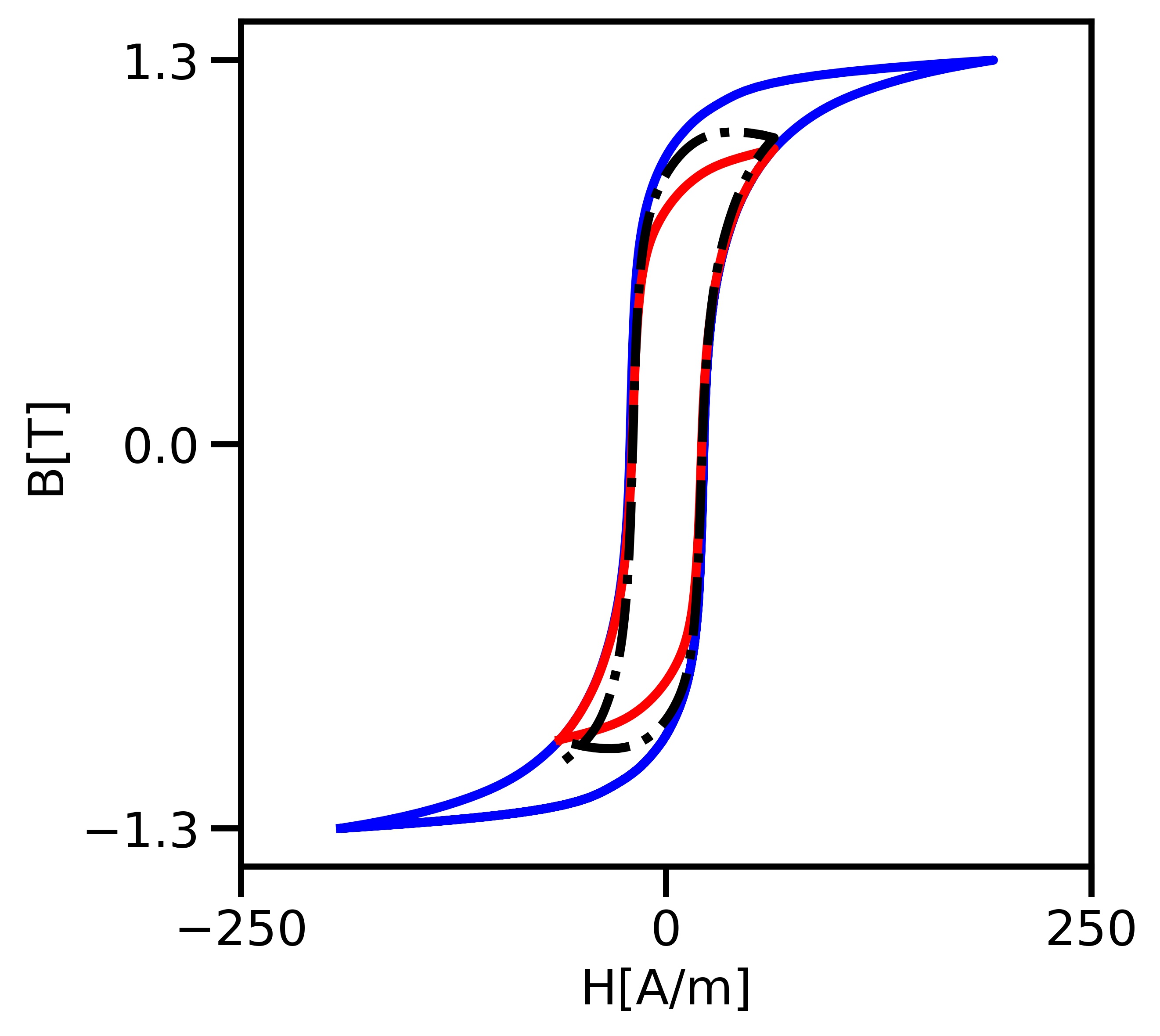}}} \\
      \subfigure[LSTM]{\label{5j}{\includegraphics[height=4.5cm, width=4.5cm]{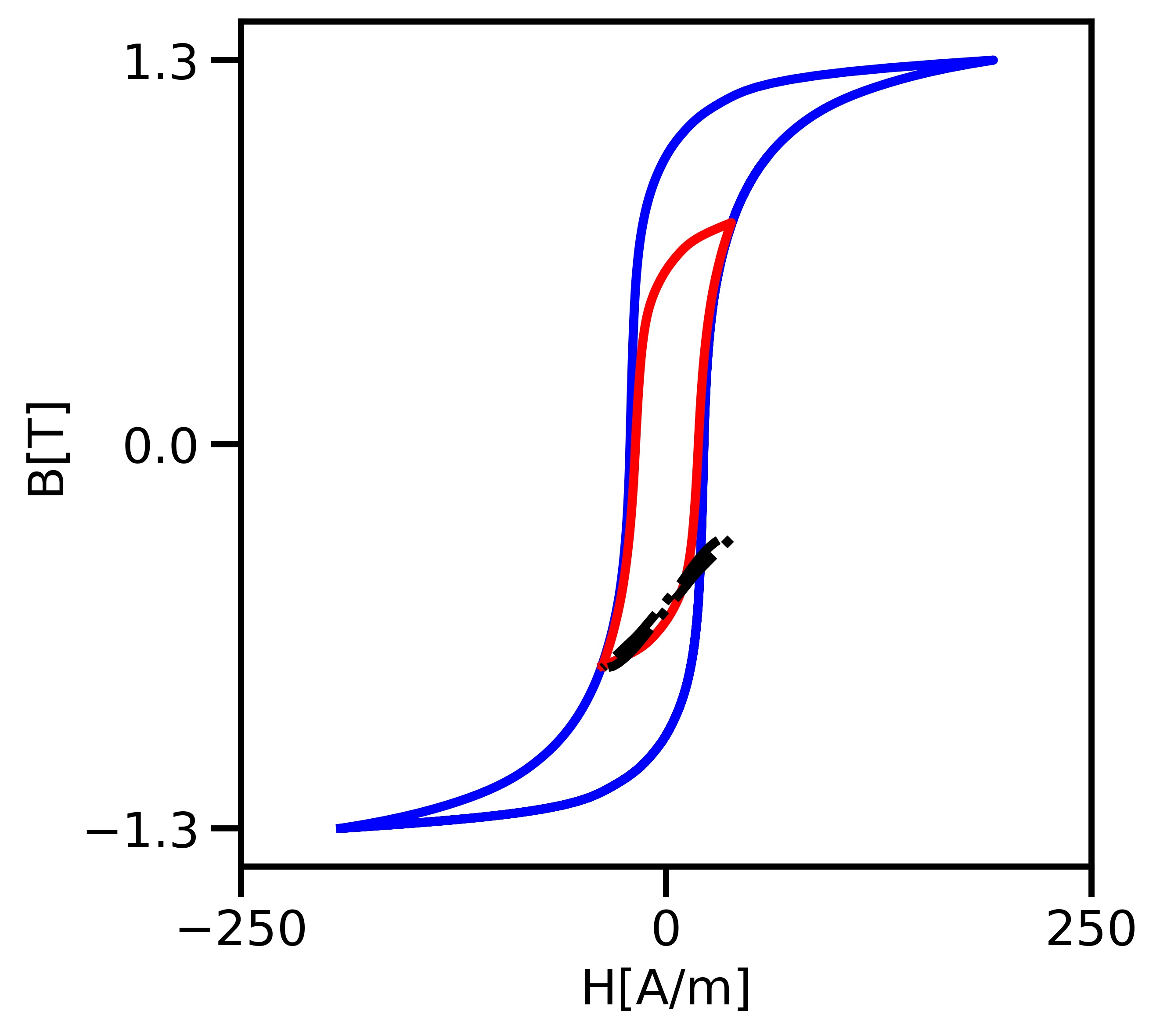}}}\hfill
      \subfigure[GRU]{\label{5k}{\includegraphics[height=4.5cm, width=4.5cm]{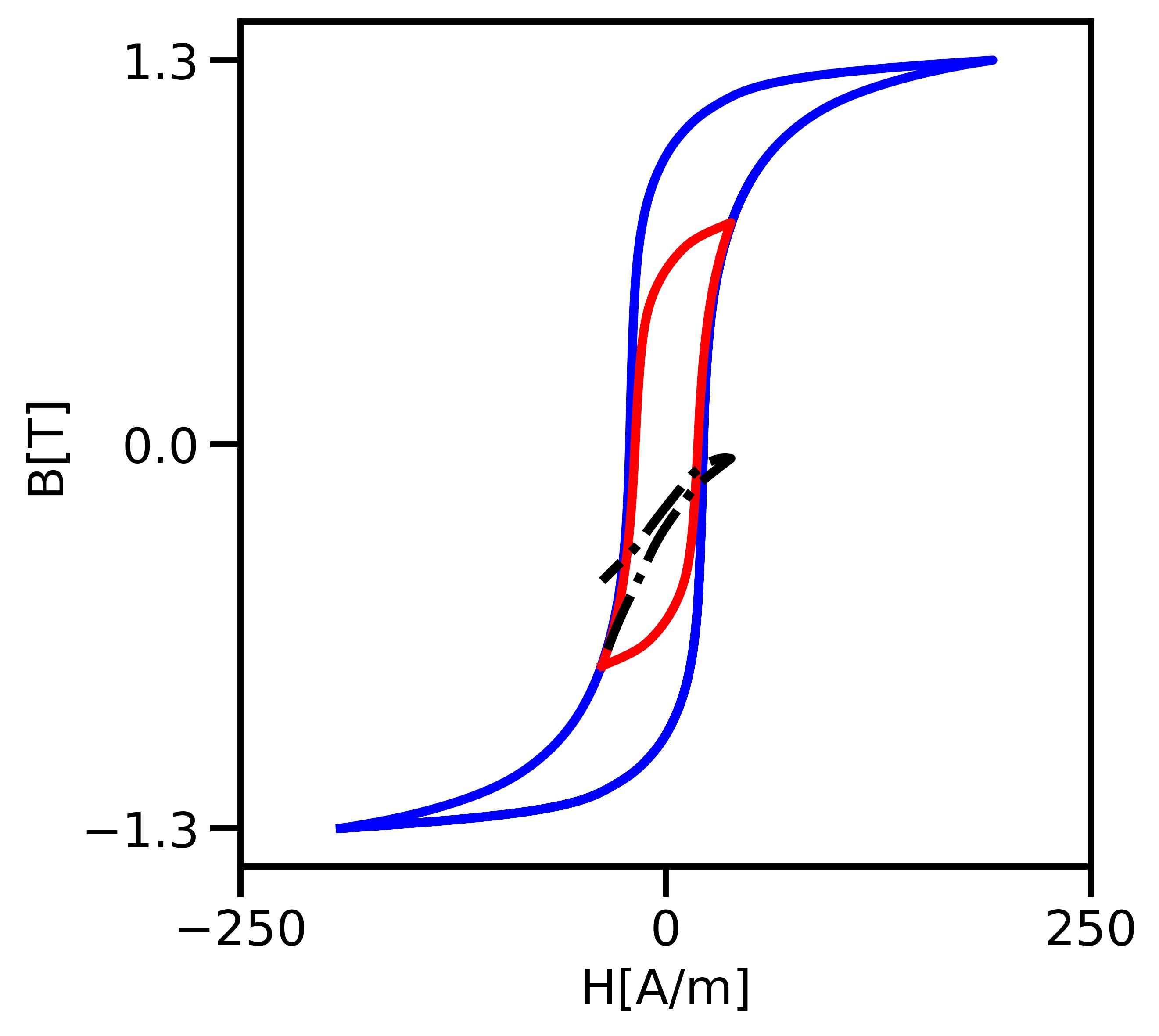}}} \hfill
      \subfigure[HystRNN]{\label{5l}{\includegraphics[height=4.5cm, width=4.5cm]{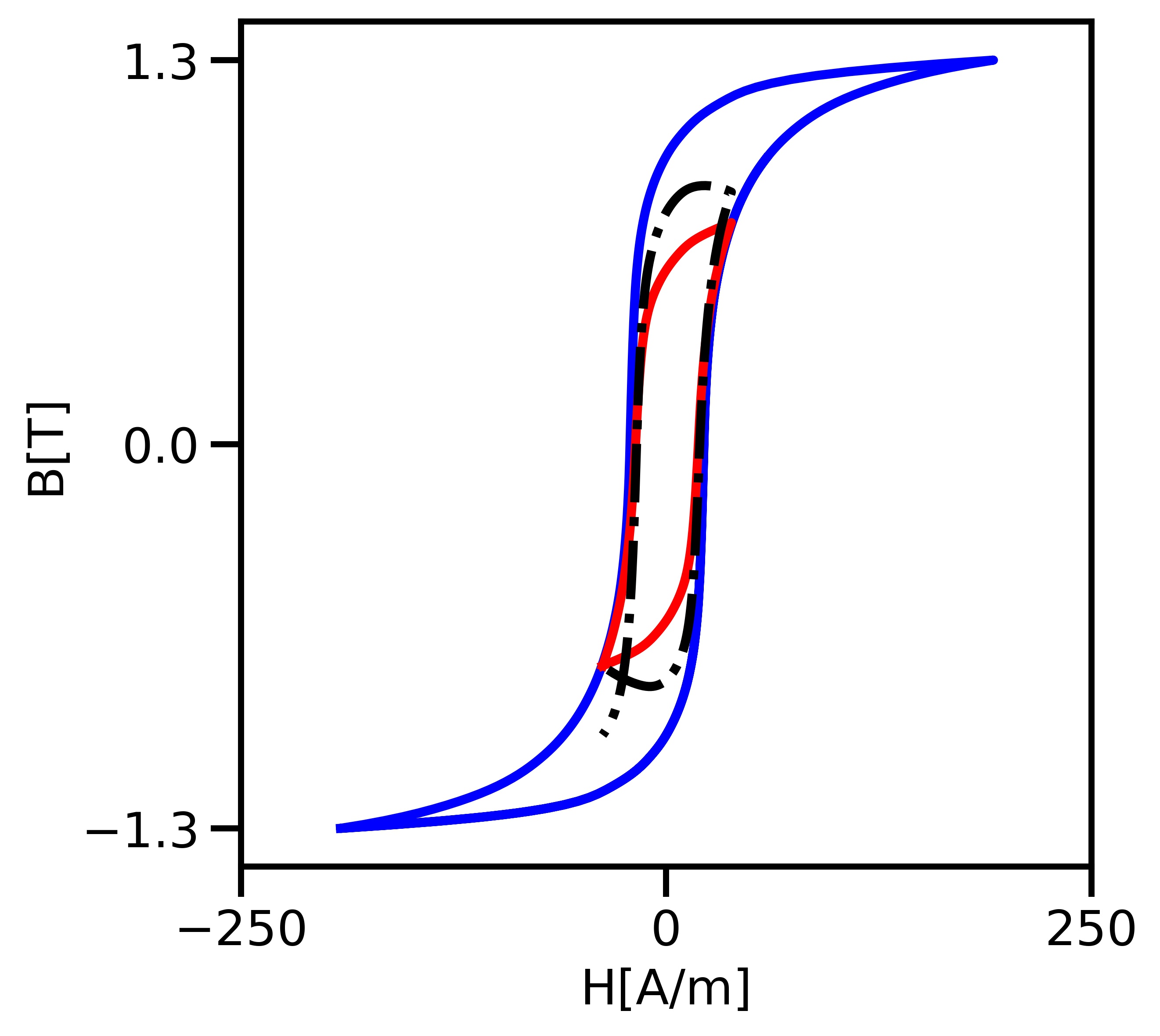}}} \hfill
    \caption{Experimental vs predicted hysteresis trajectories for experiment 3, where $\max(B)$ $=$ \SI{1.3}{\tesla}. Top two rows: predictions for $\mathcal{C_\mathrm{FORC_1}}$ and $\mathcal{C_\mathrm{FORC_2}}$ respectively. Bottom two rows: predictions for $\mathcal{C_\mathrm{minor_1}}$ and $\mathcal{C_\mathrm{minor_2}}$ respectively.} 
    \label{fig5}
\end{figure}

\begin{figure}[t]
    \centering
    \subfigure[LSTM]{\label{6a}{\includegraphics[height=4.5cm, width=4.5cm]{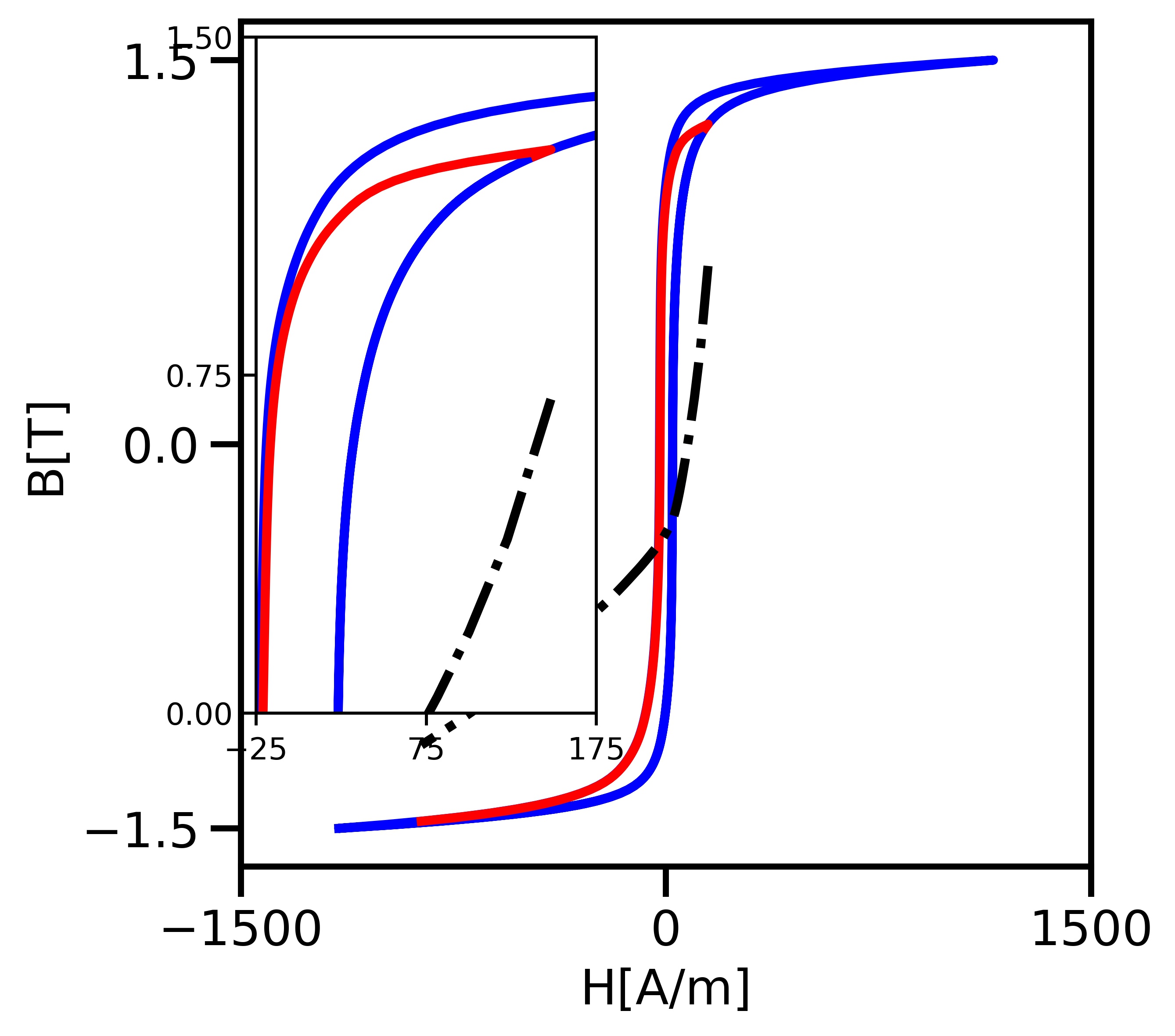}}}\hfill
      \subfigure[GRU]{\label{6b}{\includegraphics[height=4.5cm, width=4.5cm]{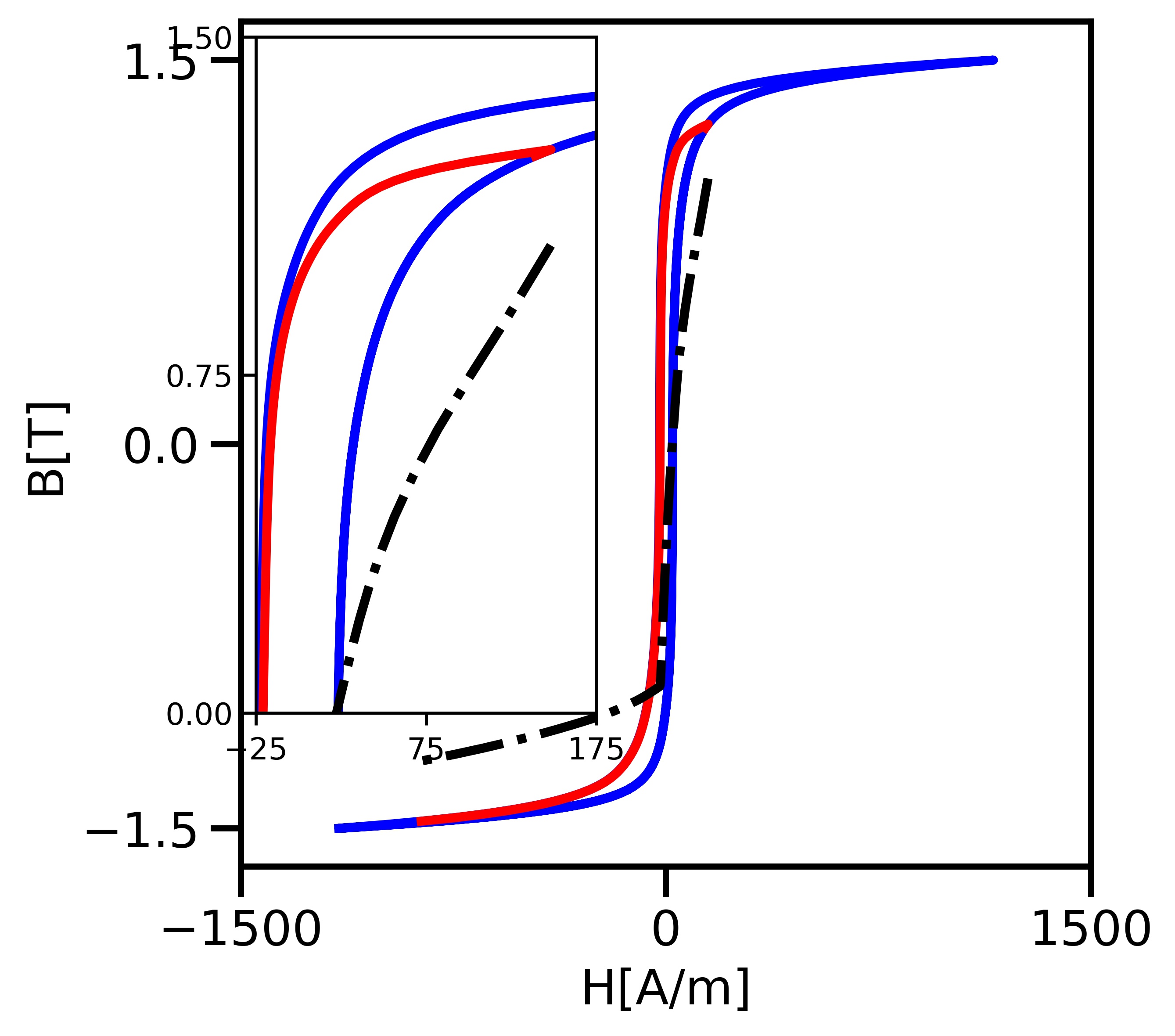}}} \hfill
      \subfigure[HystRNN]{\label{6c}{\includegraphics[height=4.5cm, width=4.5cm]{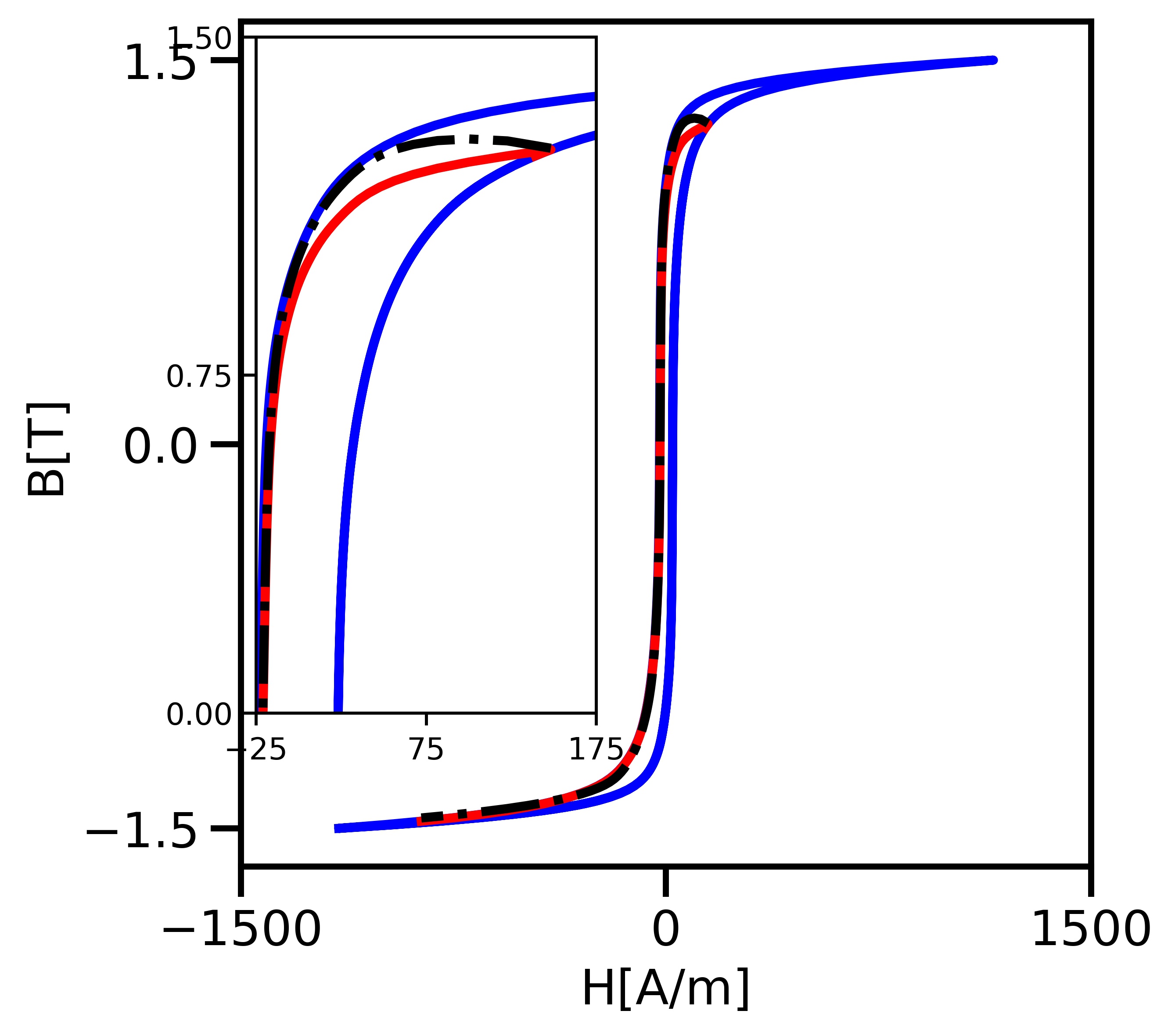}}} \\
      \subfigure[LSTM]{\label{6d}{\includegraphics[height=4.5cm, width=4.5cm]{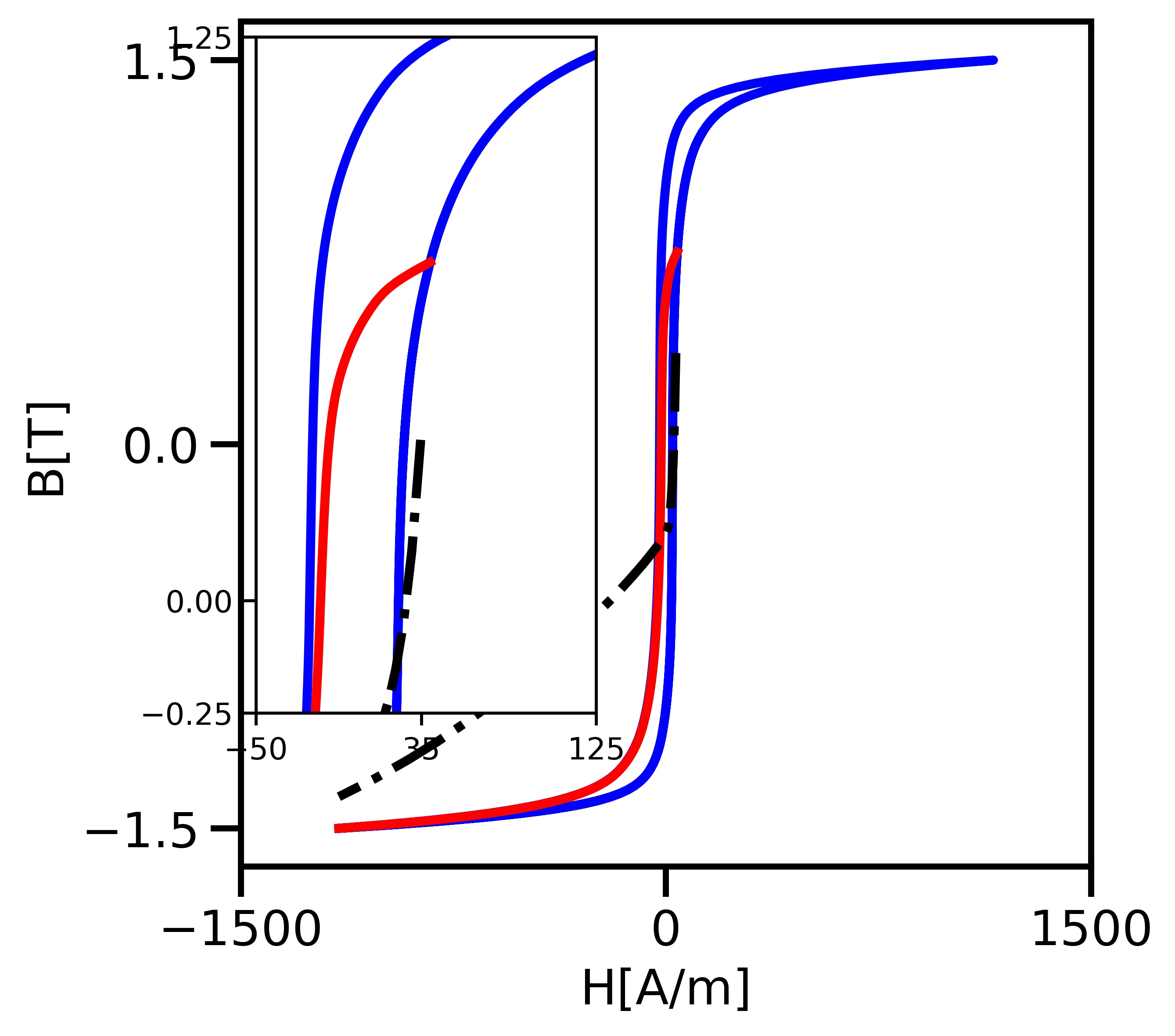}}}\hfill
      \subfigure[GRU]{\label{6e}{\includegraphics[height=4.5cm, width=4.5cm]{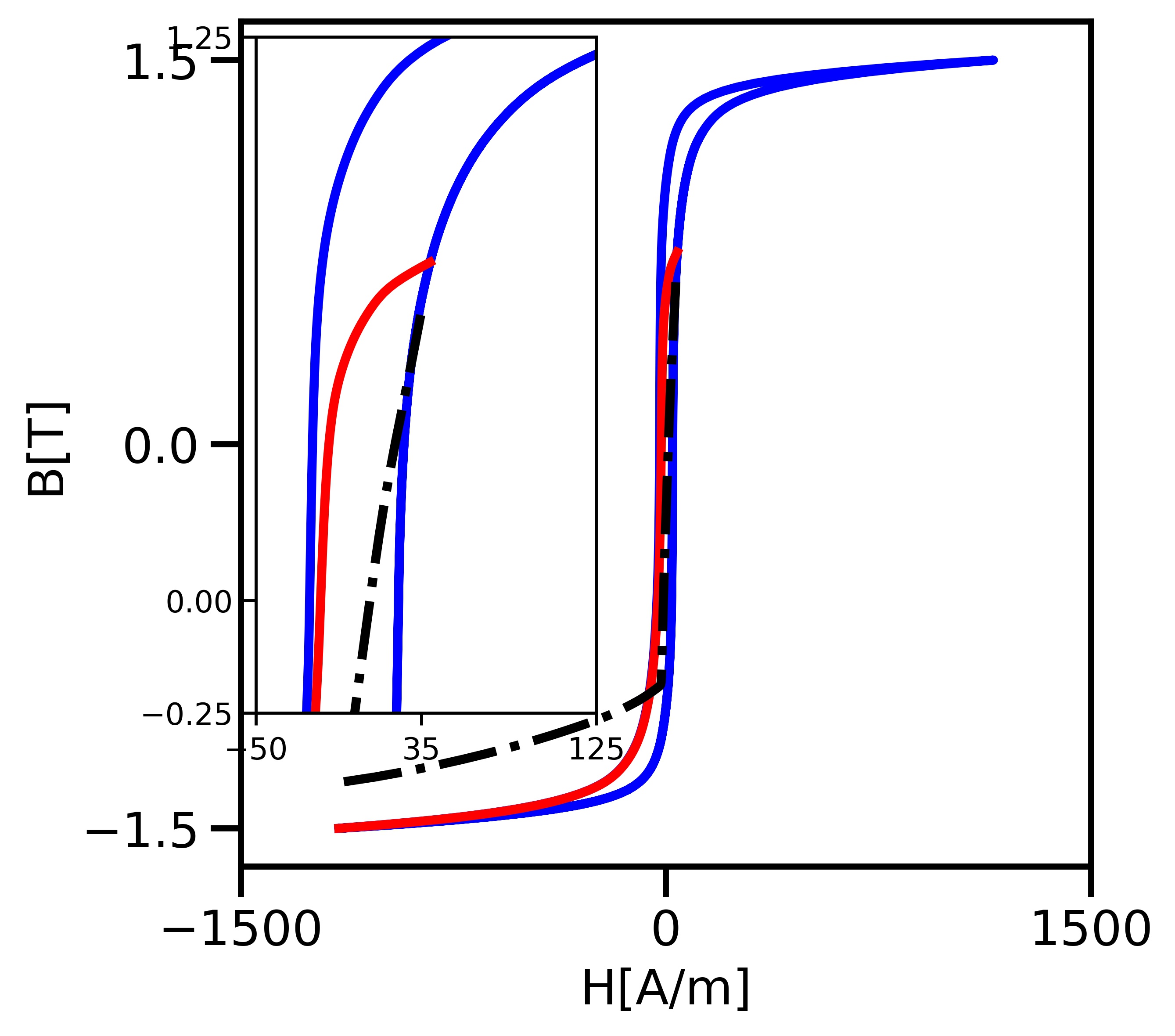}}} \hfill
      \subfigure[HystRNN]{\label{6f}{\includegraphics[height=4.5cm, width=4.5cm]{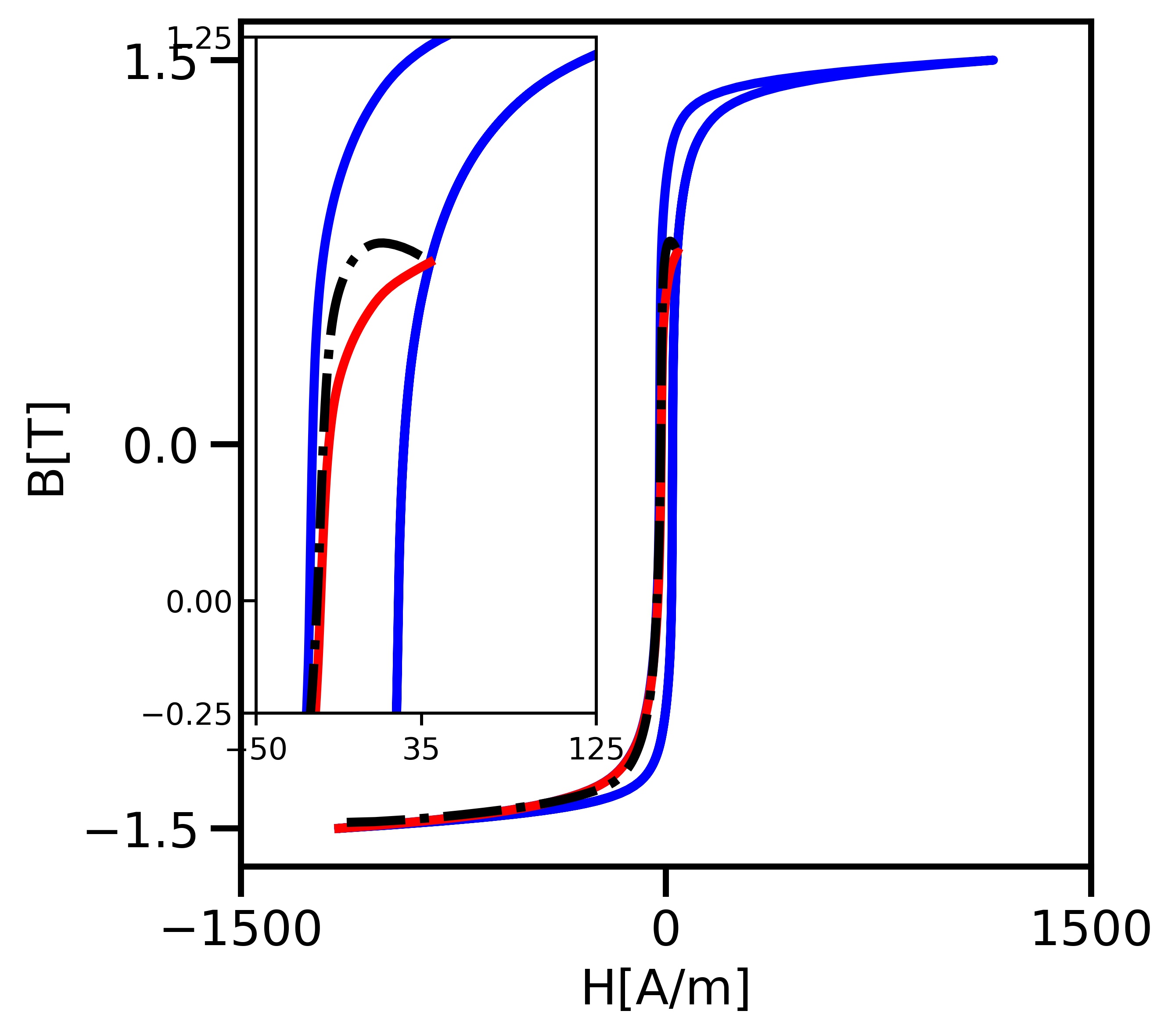}}} \\
      \subfigure[LSTM]{\label{6g}{\includegraphics[height=4.5cm, width=4.5cm]{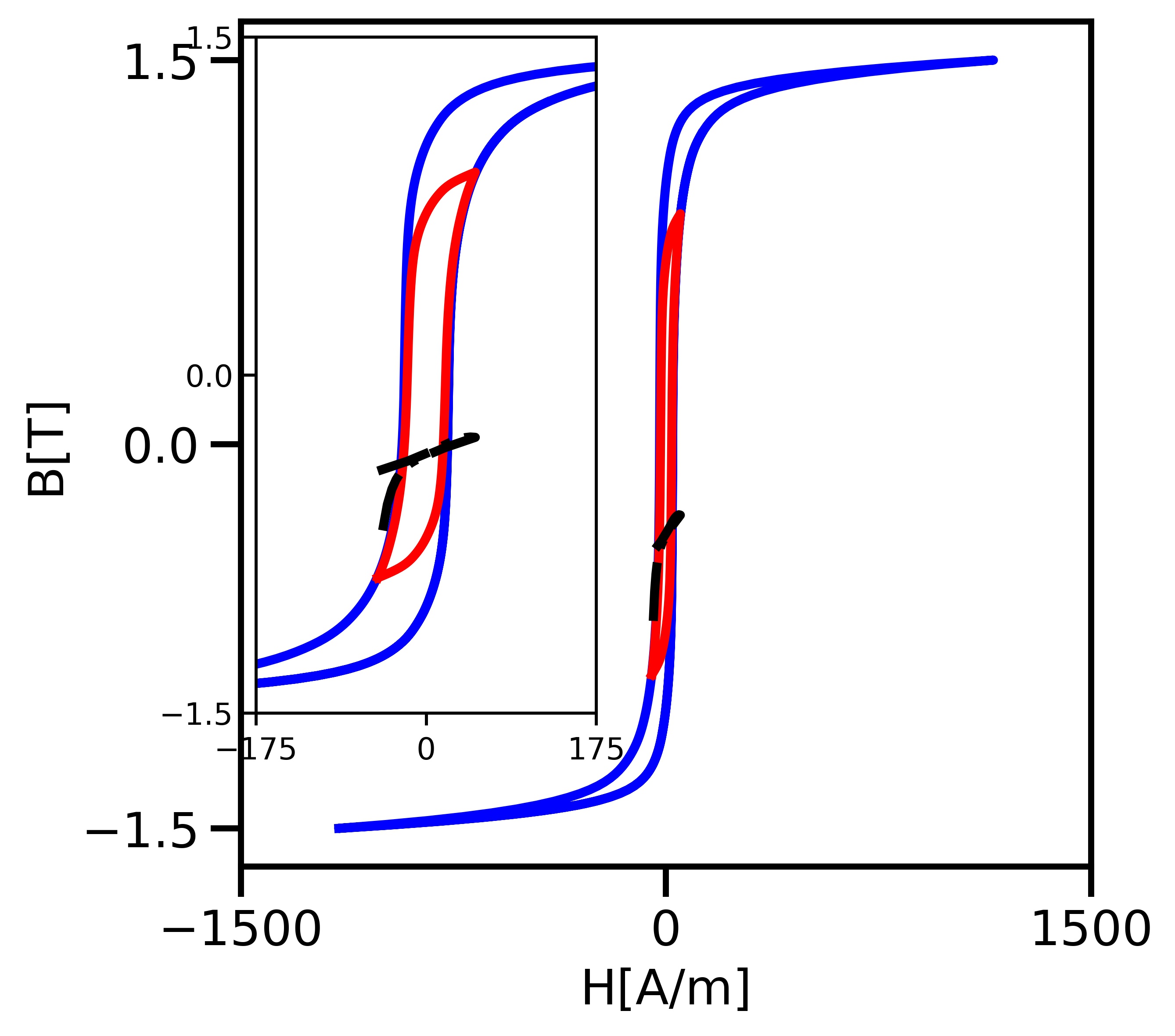}}}\hfill
      \subfigure[GRU]{\label{6h}{\includegraphics[height=4.5cm, width=4.5cm]{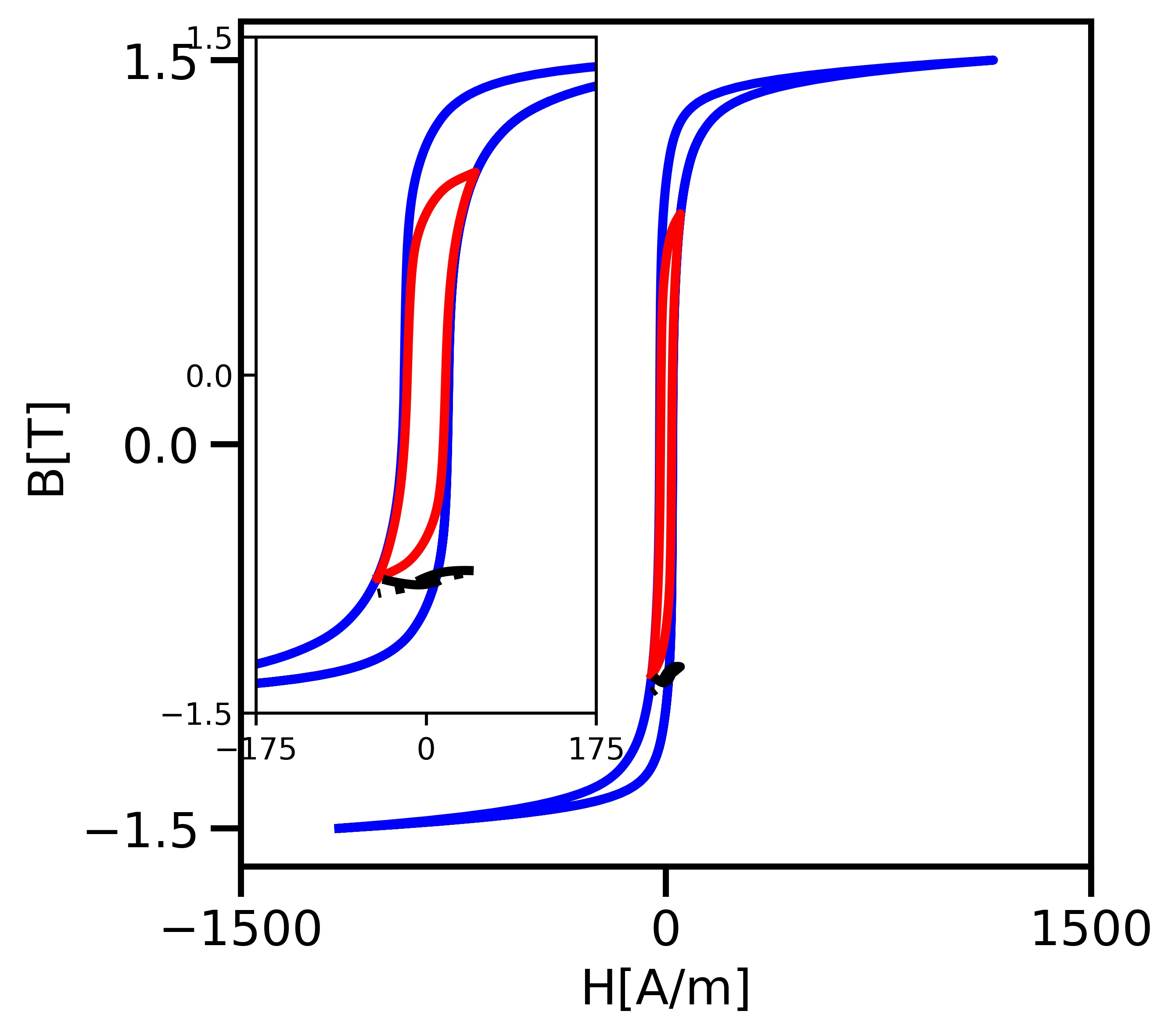}}} \hfill
      \subfigure[HystRNN]{\label{6i}{\includegraphics[height=4.5cm, width=4.5cm]{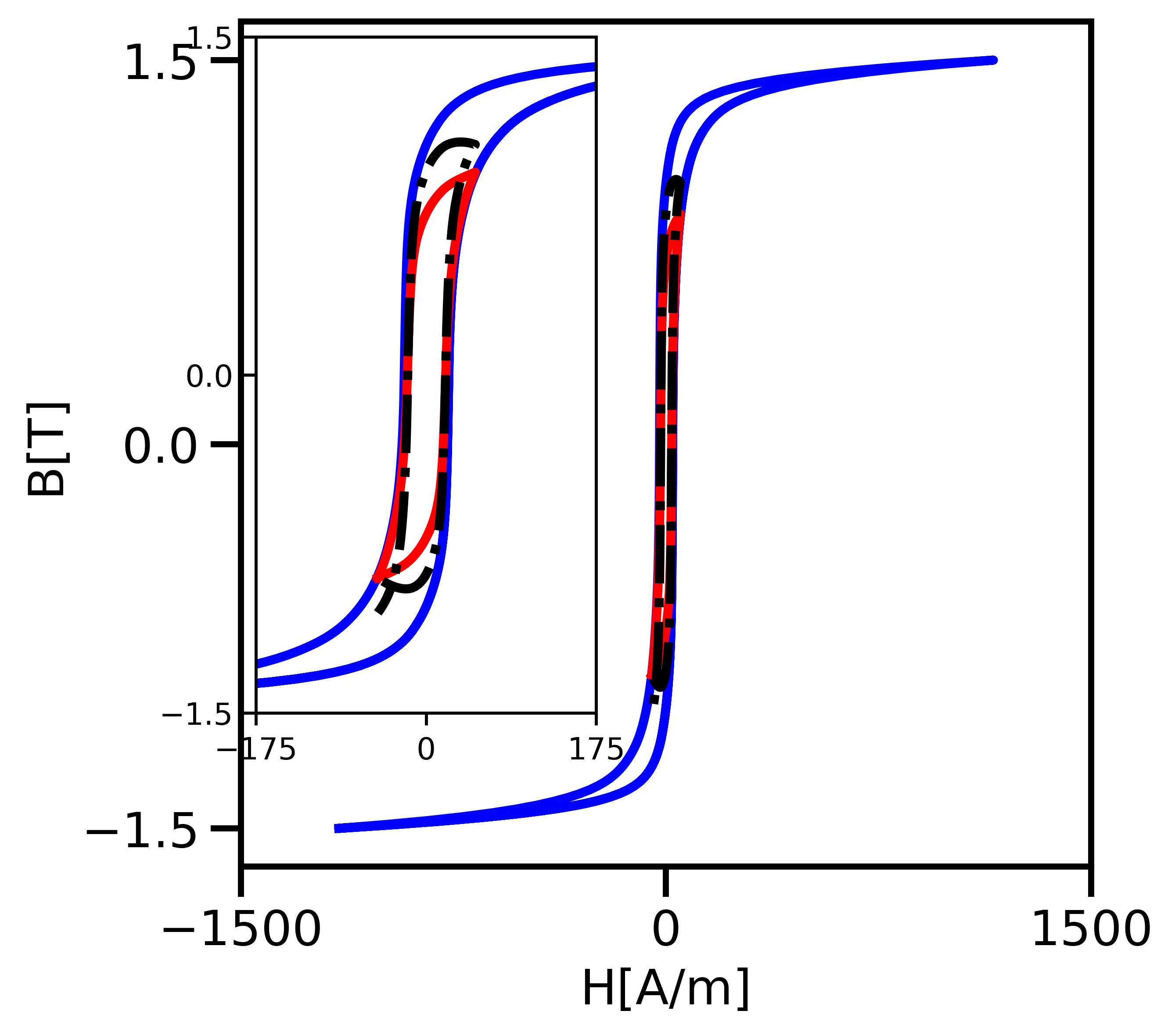}}} \\
      \subfigure[LSTM]{\label{6j}{\includegraphics[height=4.5cm, width=4.5cm]{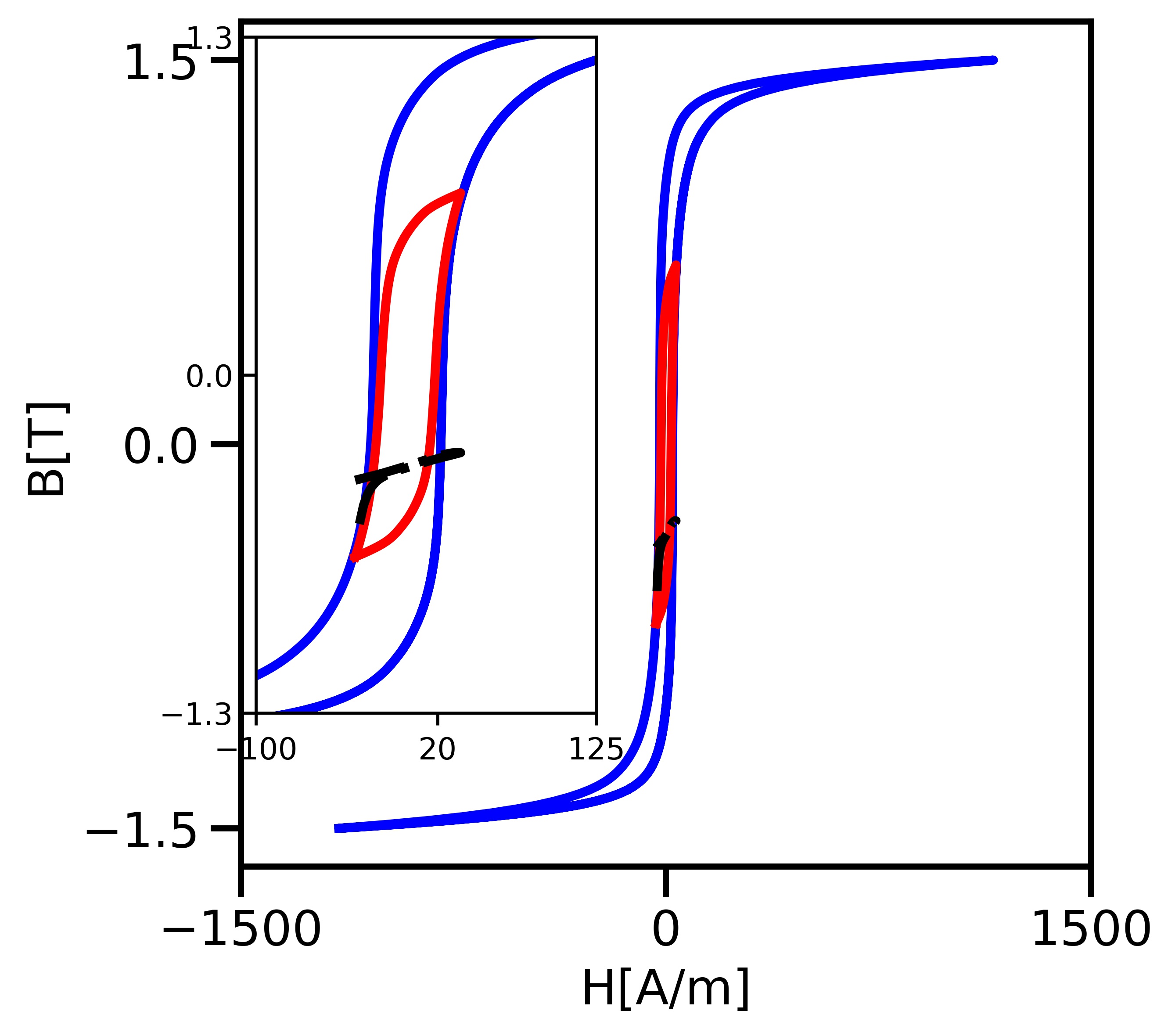}}}\hfill
      \subfigure[GRU]{\label{6k}{\includegraphics[height=4.5cm, width=4.5cm]{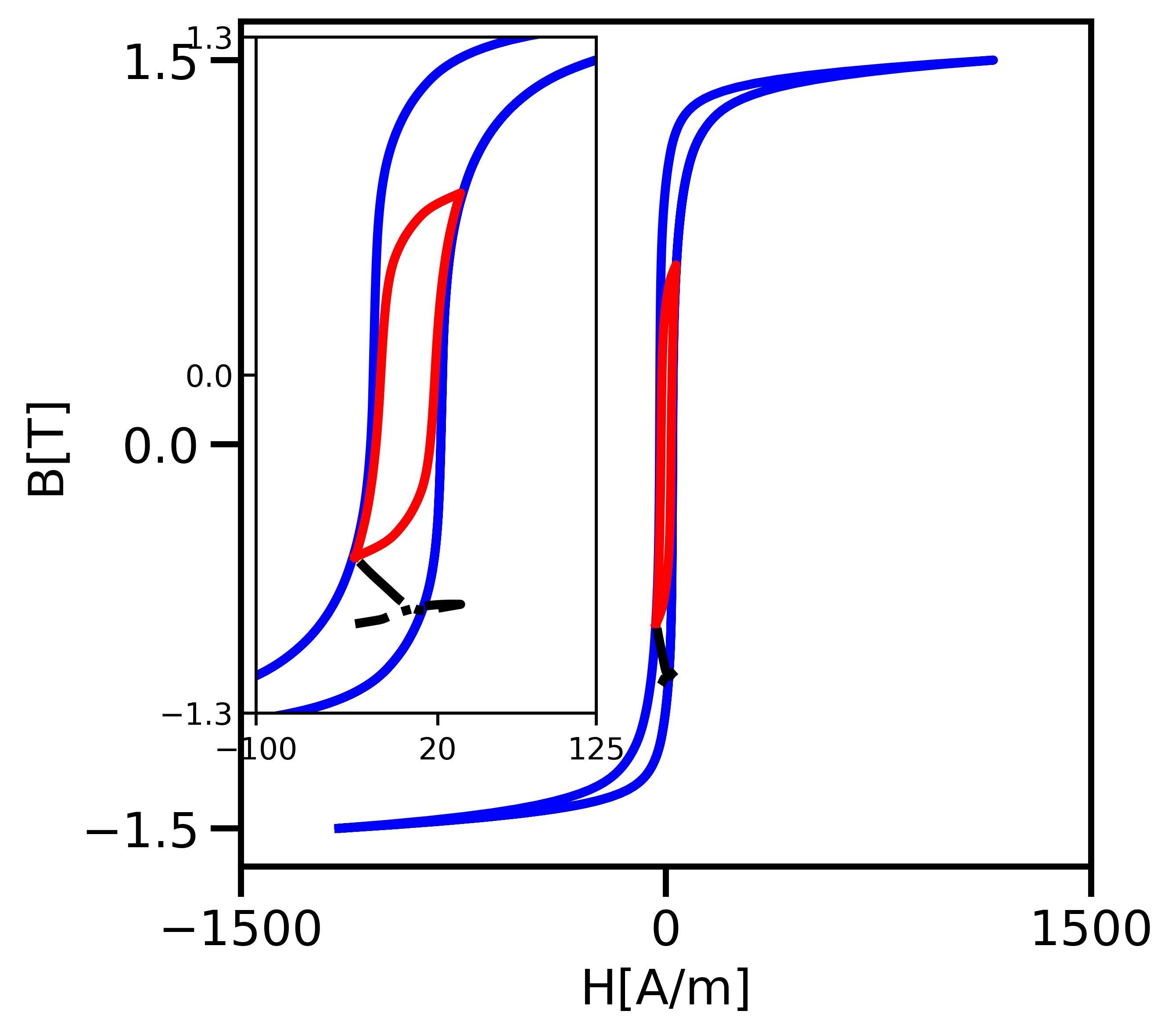}}} \hfill
      \subfigure[HystRNN]{\label{6l}{\includegraphics[height=4.5cm, width=4.5cm]{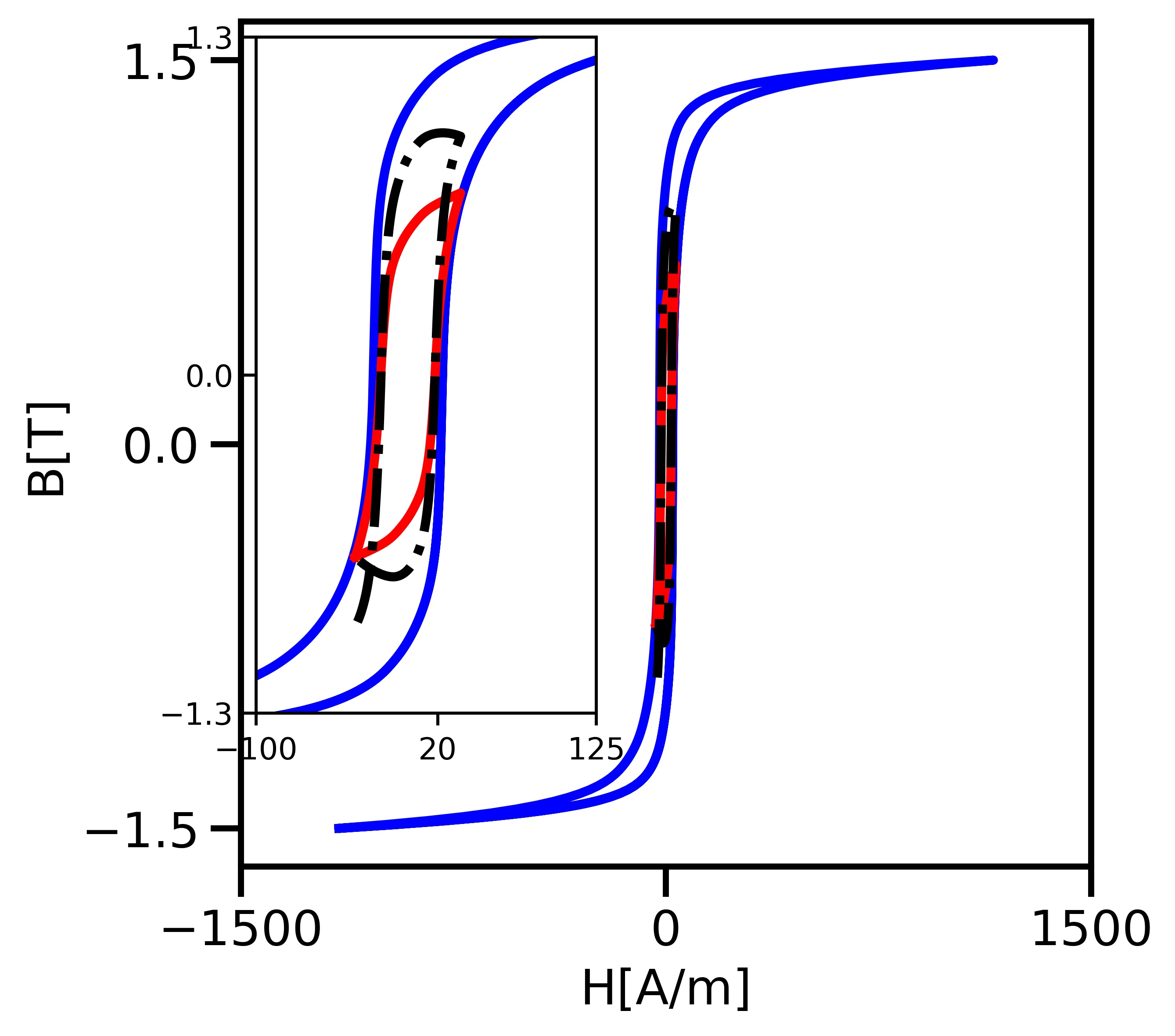}}} \hfill
    \caption{Experimental vs predicted hysteresis trajectories for experiment 4, where $\max(B)$ $=$ \SI{1.5}{\tesla}. Top two rows: predictions for $\mathcal{C_\mathrm{FORC_1}}$ and $\mathcal{C_\mathrm{FORC_2}}$ respectively. Bottom two rows: predictions for $\mathcal{C_\mathrm{minor_1}}$ and $\mathcal{C_\mathrm{minor_2}}$ respectively.} 
    \label{fig6}
\end{figure}

\subsection{Hyperparameters}
The selected hyperparameters consist of an input size of 2, a single hidden layer with a dimension of 32, and an output size of 1. The optimization process involves the utilization of the Adam optimizer, with a learning rate of $0.01$. Training is conducted for 10000 epochs, with a batch size of $1$. The hyperparameter $\Dt$ is chosen to be $0.05$. A sequence length of $595$ is chosen for all four experiments to train $\mathcal{C_\mathrm{major}}$. The determination of sequence length depends upon the data generated by the Preisach model for $\mathcal{C_\mathrm{major}}$. \emph{Uniformity} in hyperparameter settings is maintained across all experiments. Furthermore, to ensure \emph{fair comparisons} with RNN, LSTM, and GRU models, the hyperparameters are also held constant for these methods.

\subsection{Evaluation metrics}

We evaluate the proposed method using four metrics. The first is the \emph{L2-norm}, measuring the Euclidean distance between predicted and actual values. The \emph{explained variance score} indicates prediction accuracy, capturing variance proportion. \emph{Maximum error} detects significant prediction discrepancies as potential outliers. The \emph{mean absolute error} assesses average differences between predictions and actual values for overall precision. Lower L2-norm, maximum error, and mean absolute error coupled with higher explained variance signify improved performance. Metric expressions are detailed in \textbf{SM} \S \textbf{D}.

\subsection{Train and test criteria}
The trained model is tested in two distinct scenarios involving the prediction of two FORCs and two minor loops. For FORC prediction, testing sequences of lengths $199$ and $399$ are utilized, respectively. The prediction of minor loops involves a testing sequence with a length of $399$ each. As in the case of training the model, these testing sequence lengths depend on the data generated from the Preisach model for evaluating the model. The HystRNN model trained on $\mathcal{C_\mathrm{major}}$ is evaluated on $\mathcal{C_\mathrm{FORC}}$ and $\mathcal{C_\mathrm{minor}}$. This testing sequence is initiated with an input $(H_i, B_i) \in \mathcal{C_\mathrm{FORC/minor}}$, where both $H_i$, and $B_i$ are provided and $B_{i+1}$ is predicted. The output generated from this step, $B_{i+1}$, becomes the subsequent input along with $H_{i+1}$, the known magnetization for the following sequence. Such testing holds \emph{paramount importance} as \emph{practical scenarios lack prior knowledge} about the $B$ values on $\mathcal{C_\mathrm{FORC}}$ or $\mathcal{C_\mathrm{minor}}$. Thus, the sole available information for generalization stems from the predicted solution in $\mathcal{C_\mathrm{FORC}}$ or $\mathcal{C_\mathrm{minor}}$.

\subsection{Experimental Results}

Four experiments are carried out to evaluate the performance of HystRNN. The experiments differ by the \emph{maximum permitted magnetic flux density} of the electrical machine $B_\mathrm{max}$. Exact $B_\mathrm{max}$ values are indicated for the experiments in \textbf{SM} \S \textbf{G}. Performing experiments and exploring the generalization capabilities of the model for varying $B_\mathrm{max}$ values is \emph{crucial} for understanding and optimizing the performance and efficiency of different and diverse machines. For instance, machines requiring lower magnetic flux densities of $B_\mathrm{max}$ $=$ \SI{1.25}{\tesla} are typically used as high-efficiency induction motors \cite{he2016dynamic} in industrial settings for tasks like driving conveyor belts, pumps, and compressors. Meanwhile, high-performance applications, for instance, particle accelerators \cite{reinholds2012effect} and nuclear magnetic resonance \cite{mori2000magnetic}, demand higher magnetic flux densities with $B_\mathrm{max}$ $=$ \SI{1.7}{\tesla} for their operation. We perform experiments for $B_\mathrm{max}$ in this spectrum and examine its \emph{performance for generalized scenarios}, tailoring designs for diverse needs, ensuring energy efficiency for everyday devices, and pushing technological boundaries for cutting-edge systems.

For all the experiments, the data for the major loop, $\mathcal{B_\mathrm{major}}$, is collected until $\max(B)$ reaches the saturation value $B_\mathrm{max}$. HystRNN and other compared methods LSTM, GRU, and RNN are trained on $\mathcal{B_\mathrm{major}}$. Once the model is trained, they are tested for four distinct cases. The first and second test cases correspond to estimating FORC. We denote them as $\mathcal{C_\mathrm{FORC_1}}$ and $\mathcal{C_\mathrm{FORC_2}}$ respectively. Two distinct FORCs are chosen to study the effect of the distance between the origin of the FORC and $B_\mathrm{max}$. The third and fourth test cases are performed for predicting minor loops, which we denote by $\mathcal{C_\mathrm{minor_1}}$ and $\mathcal{C_\mathrm{minor_2}}$ respectively. These minor loops vary based on the maximum value of $B$ to which they are subjected. The origin of $\mathcal{C_\mathrm{FORC_1}}$, $\mathcal{C_\mathrm{FORC_2}}$ and the maximum $B$ value of the minor loop is provided in \textbf{SM} \S \textbf{G}.

Detailed performance metrics for HystRNN are outlined in Tables 1 to 4, corresponding to experiments 1 through 4, respectively. The Tables also facilitate a comprehensive comparative analysis with RNN, LSTM, and GRU. The metrics notably emphasize the \emph{superior} performance of our proposed method \mbox{HystRNN} across all numerical experimentation scenarios.

\subsubsection{Experiment 1}
 In Fig.~\ref{fig3}, the top two rows display the predictions of $\mathcal{C_\mathrm{FORC_{1,2}}}$ respectively, wherein training exclusively occurs on $\mathcal{C_\mathrm{major}}$, indicated by the blue color. Predictions are represented with black color, and ground truth is represented with red color. The colors are kept consistent for all the following experiments. The top two rows show that LSTM (Fig.~\ref{3a}, ~\ref{3d}) and GRU (Fig.~\ref{3b}, ~\ref{3e}) fail drastically to capture the shape of the FORC accurately. In contrast, HystRNN effectively captures the  \emph{structure} and \emph{symmetry} of reversal curves as shown in Fig.~\ref{3c} and ~\ref{3f}. The last two rows show the prediction for minor loop $\mathcal{C_\mathrm{minor_{1,2}}}$. For this case also predictions from LSTM (Fig.~\ref{3g}, ~\ref{3j}) and GRU (Fig.~\ref{3h}, ~\ref{3k}) are inaccurate. Neither LSTM nor GRU could form a closed loop for the predicted trajectory, posing a major challenge to compute the energy loss without a closed region, as energy loss depends on the surface area of the hysteresis loop. In contrast, our proposed method HystRNN predicts the \emph{structure} of the loop very well and efficiently models the minor loop (Fig.~\ref{3i}, ~\ref{3l}). 

\subsubsection{Experiment 2}
The top two rows of Fig.~\ref{fig4} present that LSTM (Fig.~\ref{4a}, ~\ref{4d}) and GRU (Fig.~\ref{4b}, ~\ref{4e}) fail to predict the trajectory of FORC by a huge margin. On the other hand, HysRNN shows close agreement with the ground truth for predicting $\mathcal{C_\mathrm{FORC_{1,2}}}$ (Fig.~\ref{4c}, ~\ref{4f}). Also, the prediction of HystRNN for $\mathcal{C_\mathrm{FORC_{1}}}$ is slightly better than for $\mathcal{C_\mathrm{FORC_{2}}}$, exemplifies that the model performs better when the origin of FORC is closer to $\max(B)$. A possible reason for this behavior could be the resemblance in the trajectories of $\mathcal{C_\mathrm{major}}$ and a FORC from a higher origin value. The last two rows of Fig.~\ref{fig4} present the predictions of $\mathcal{C_\mathrm{minor_{1,2}}}$ respectively. For this case, LSTM (Fig.~\ref{4g}, ~\ref{4j}) and GRU (Fig.~\ref{4h}, ~\ref{4k}) almost form a loop-like shape; however, they are very off from compared to the ground truth. HysRNN, on the other hand, captures the loop shape efficiently, as presented in Fig.~\ref{4i} and ~\ref{4l}.

\subsubsection{Experiment 3}

The predictions for the model, the ground truth, and the training data are presented in Fig.~\ref{fig5}. As presented in Fig.~\ref{5a}, ~\ref{5b}, ~\ref{5d}, and ~\ref{5e} predictions by LSTM and GRU models exhibit a lack of accuracy. In contrast, predictions of our model HystRNN for the reversal curve are notably precise, as evidenced in Figures Fig.~\ref{5c} and ~\ref{5f}. The final two rows within Fig.~\ref{fig5} present that HystRNN accurately captured the characteristics of the minor loop, as showcased in Fig.~\ref{5i} and ~\ref{5l}. Prediction by GRU manages to capture a resemblance of the loop, although not entirely, as revealed in Fig.~\ref{5h} and ~\ref{5k}. On the other hand, LSTM performs poorly, failing to capture the intricate structure of the minor loop, as depicted in Fig.~\ref{5g} and ~\ref{5j}.

\subsubsection{Experiment 4}
The predictions for the model, the ground truth, and the training data are presented in Fig.~\ref{fig6}. Predictions of the reversal curve show agreement with the nature observed in previous experiments. In this case, $\max(B)$, origin of $\mathcal{C_\mathrm{FORC_{2}}}$, and maximum $B$ value of $\mathcal{C_\mathrm{minor_{2}}}$ vary significantly, posing a challenge for both LSTM and GRU. However, HystRNN outperforms them for each case, as shown in Fig.~\ref{fig6}. The results underscore the performance of HystRNN as, for neither of the cases, the accuracy of LSTM or GRU is comparable to our proposed method. Additional visual results for all the RNN experiments are supplemented in \textbf{SM} \S \textbf{E}.

\section{Conclusions}
We introduced a novel neural oscillator, \emph{HystRNN}, aimed to \emph{advance} magnetic hysteresis modeling within \emph{extrapolated regions}. The proposed oscillator is based upon the foundation of coupled oscillator recurrent neural networks and \emph{inspired from} phenomenological hysteresis models. HystRNN was validated by predicting first-order reversal curves and minor loops after training the model \emph{solely} with major loop data. The outcomes underscore the \emph{superiority} of HystRNN in adeptly capturing intricate nonlinear dynamics, \emph{outperforming} conventional recurrent neural architectures such as RNN, LSTM, and GRU on \emph{various metrics}. This performance is attributed to its capacity to assimilate sequential information, history dependencies, and hysteretic features, ultimately achieving generalization capabilities. Access to the codes and data will be provided upon publication.

\bibliographystyle{unsrt}  
\bibliography{references}  

\section*{Supplementary Material}
\subsection*{SM \S A: Nomenclature}

The table provided below presents the abbreviations utilized within this paper.

\begin{table}[h]
\caption{Abbreviations used in this paper}
\vskip 0.15in
\begin{center}
\begin{small}
\begin{sc}
\begin{tabular}{ll}
\toprule
 Symbol & Description    \\
\midrule
\text{CoRNN} & \text{coupled-oscillatory recurrent neural network} \\
\text{GRU} & \text{Gated recurrent unit} \\
\text{LSTM} & \text{Long short-term memory} \\
\text{ODE} & \text{Ordinary differential equation} \\
\text{RNN} & \text{Recurrent neural network} \\
\bottomrule
\end{tabular}
\end{sc}
\end{small}
\end{center}
\vskip -0.1in
\end{table}

\subsection*{SM \S B: Preisach model of hysteresis and data generation for magnetic experiments}
The modeling of magnetic materials is a critical step when designing electromechanical devices, mostly to predict the behavior of the ferromagnetic core. Often, this is achieved by a so-called anhysteretic curve that describes the ideal magnetic behavior of a soft-magnetic material which is not subjected to any hysteresis effects. However, for increased insight it is desired to include this hysteresis behavior into the simulation.

One of the most widely known models to describe the hysteresis relation is the Preisach model of hysteresis. It is a phenomenological model that describes the hysteresis effect by a set of hysteresis operators, scattered on a triangular domain called the Preisach plane, which are scaled by a weight function, following
\begin{equation}
f(t) = \iint \limits_{\alpha\geq\beta} P(\alpha,\beta) \hat{\gamma}_{\alpha,\beta} u(t) d\alpha d\beta
\label{eq:Preisach:general}
\end{equation}
where:
\begin{itemize}
  \item \( u \) is the input,
  \item \( f \) is the output,
  \item \( t \) is a moment in time,
  \item \( \alpha~\&~\beta \) are two switching variables,
  \item \( P \) is the Preisach weight function,
  \item \( \hat{\gamma} \) represents the hysteresis operators.
\end{itemize}
Typically, the weight function is constructed from well-known probability density functions, for which the parameters are tuned to closely match the magnetic material behavior. Alternatively, the weight function can also be obtained from magnetic measurement data, which generally results in the best correspondence, albeit at the cost of a more complicated identification procedure. Furthermore, in its general form, the model output is not affected by the rate of change of the input, and the model describes solely the static scalar hysteresis behavior. However, various generalization exist to include, e.g. dynamical eddy currents effects, or the vector hysteresis phenomenon. Ultimately, in this work the static scalar Preisach model was applied, with a weight function fitted on a set of measurement data.

The measurement data in this work was obtained by a Brockhaus MPG 200 soft-magnetic material tester, using an Epstein frame calibrated to correspond with the IEC standard. A set of concentric hysteresis loops up to a maximum of 1.7~T was measured for NO27-1450H, obtained under quasi-DC excitation. Here, quasi-DC indicates that the rate of change of the magnetic flux density was controlled such that any eddy current fields are negligible, and the static hysteresis behavior is obtained. The Epstein sample strips used were obtained using spark eroding and cut in the rolling direction.
\begin{figure}[t]
    \centering
    \includegraphics[height=3.5cm, width=3.5cm]{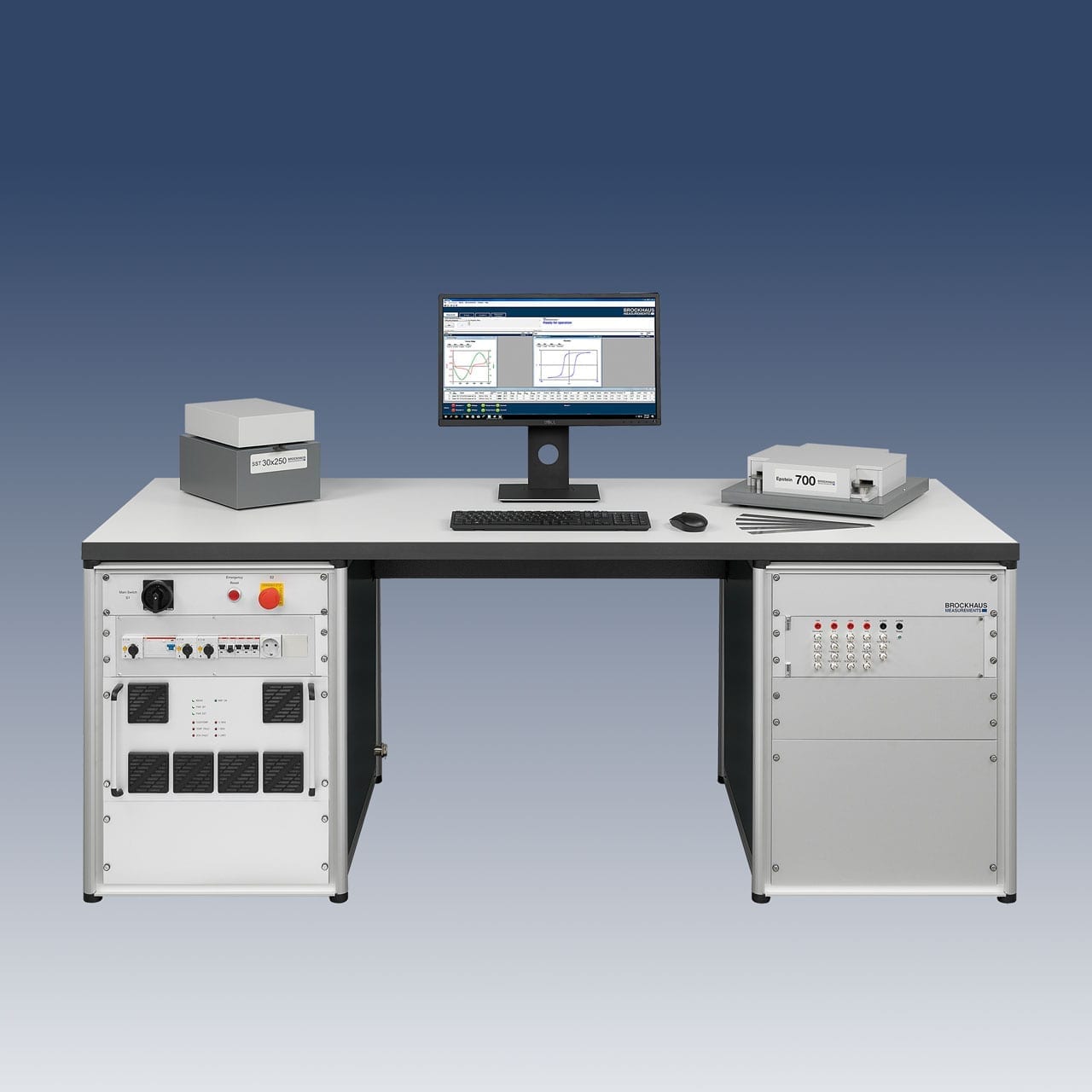}
    \caption{The MPG 200 desk by Brockhaus, with Epstein frame and sample strips shown on the right.} 
    \label{Brock}
\end{figure}

\subsection*{SM \S C: Data Normalization}
The procedure used to normalize the data for training all the models is described in the following subsections.

\subsubsection*{Min-Max Scaling}

Min-Max Scaling is a technique used to transform numerical data so that it falls within a specific range, typically between 0 and 1. The purpose of this normalization is to bring all the features to a similar scale, which can be particularly important for algorithms that are sensitive to the scale of input data.

\begin{align*}
X_{\text{min}} & \text{ represents the minimum value in the original dataset } X. \\
X_{\text{max}} & \text{ represents the maximum value in the original dataset } X. \\
X_{\text{scaled}} & \text{ is the result of scaling the original dataset.}
\end{align*}

The formula for scaling the data is:
\[
X_{\text{scaled}} = \frac{X - X_{\text{min}}}{X_{\text{max}} - X_{\text{min}}} \times 2 - 1
\]

In this formula:
\begin{itemize}
  \item $(X - X_{\text{min}})$ calculates the difference between each data point and the minimum value.
  \item $(X_{\text{max}} - X_{\text{min}})$ represents the range of the data.
  \item Dividing by $(X_{\text{max}} - X_{\text{min}})$ scales the data to the range $[0, 1]$.
  \item Multiplying by 2 and subtracting 1 scales the data to the range $[-1, 1]$.
\end{itemize}

So, $X_{\text{scaled}}$ contains the normalized data in the range of $[-1, 1]$. 

\subsubsection*{Inverse Min-Max Scaling}

After Min-Max Scaling to revert the scaled data back to its original values. The formula for this Inverse Min-max scaling is given as:

\[
X_{\text{original}} = \frac{X_{\text{scaled}} + 1}{2} \times (X_{\text{max}} - X_{\text{min}}) + X_{\text{min}}
\]

In this formula:
\begin{itemize}
  \item $\frac{X_{\text{scaled}} + 1}{2}$ scales the scaled data back to the range $[0, 1]$.
  \item $(X_{\text{max}} - X_{\text{min}})$ represents the original range of the data.
  \item Multiplying by $(X_{\text{max}} - X_{\text{min}})$ scales the data back to the original range.
  \item Adding $X_{\text{min}}$ adjusts the scaled data to its original position.
\end{itemize}

\subsection*{SM \S D: Error Metrics}
The error metrics used in this study are described in the following subsections.

\subsubsection*{$L^2$ norm}
The formula for the relative $L^2$ norm of \( \hat{B} \) with respect to ground truth \( B \) is given by:
\[
\text{Relative $L^2$ norm} = \frac{\|\hat{B} - B\|_2}{\|B\|_2}
\]

where:
\begin{itemize}
  \item \( \|\hat{B} - B\|_2 \) is the Euclidean distance between vectors \( \hat{B} \) and \( B \),
  \item \( \|B\|_2 \) is the Euclidean norm (magnitude) of vector \( B \).
\end{itemize}

\subsubsection*{Explained variance score}
The formula for the explained variance score is given by:
\[
\text{Explained Variance Score} = 1 - \frac{\sum_{i=1}^{n} (B_i - \hat{B_i})^2}{\sum_{i=1}^{n} (B_i - \bar{B})^2}
\]

where:
\begin{itemize}
  \item \( n \) is the number of data points,
  \item \( B_i \) represents the ground truth at the \( i \)-th data point,
  \item \( \hat{B_i} \) represents the predicted value at the \( i \)-th data point,
  \item \( \bar{B} \) represents the mean of the ground truth.
\end{itemize}

\subsubsection*{Max error}
The formula for the maximum absolute error is given by:
\[
\text{Max Absolute Error} = \max_{i=1}^{n} |B_i - \hat{B_i}|
\]

where:
\begin{itemize}
  \item \( n \) is the number of data points,
  \item \( B_i \) represents the ground truth at the \( i \)-th data point,
  \item \( \hat{B_i} \) represents the predicted value at the \( i \)-th data point,
  \item \( |x| \) represents the absolute value of \( x \).
\end{itemize}

\subsubsection*{Mean absolute error}
The formula for the mean absolute error is given by:
\[
\text{Mean Absolute Error} = \frac{1}{n} \sum_{i=1}^{n} |B_i - \hat{B_i}|
\]

where:
\begin{itemize}
  \item \( n \) is the number of data points,
  \item \( B_i \) represents the ground truth at the \( i \)-th data point,
  \item \( \hat{B_i} \) represents the predicted value at the \( i \)-th data point,
  \item \( |x| \) represents the absolute value of \( x \).
\end{itemize}

\subsection*{SM \S E: Additional results for RNN}
Additional results for RNN, for experiment 1 through 4 are provided in Fig. 8 to 11, respectively.

\begin{figure}
    \centering
    \subfigure[]{\includegraphics[height=3.5cm, width=3.5cm]{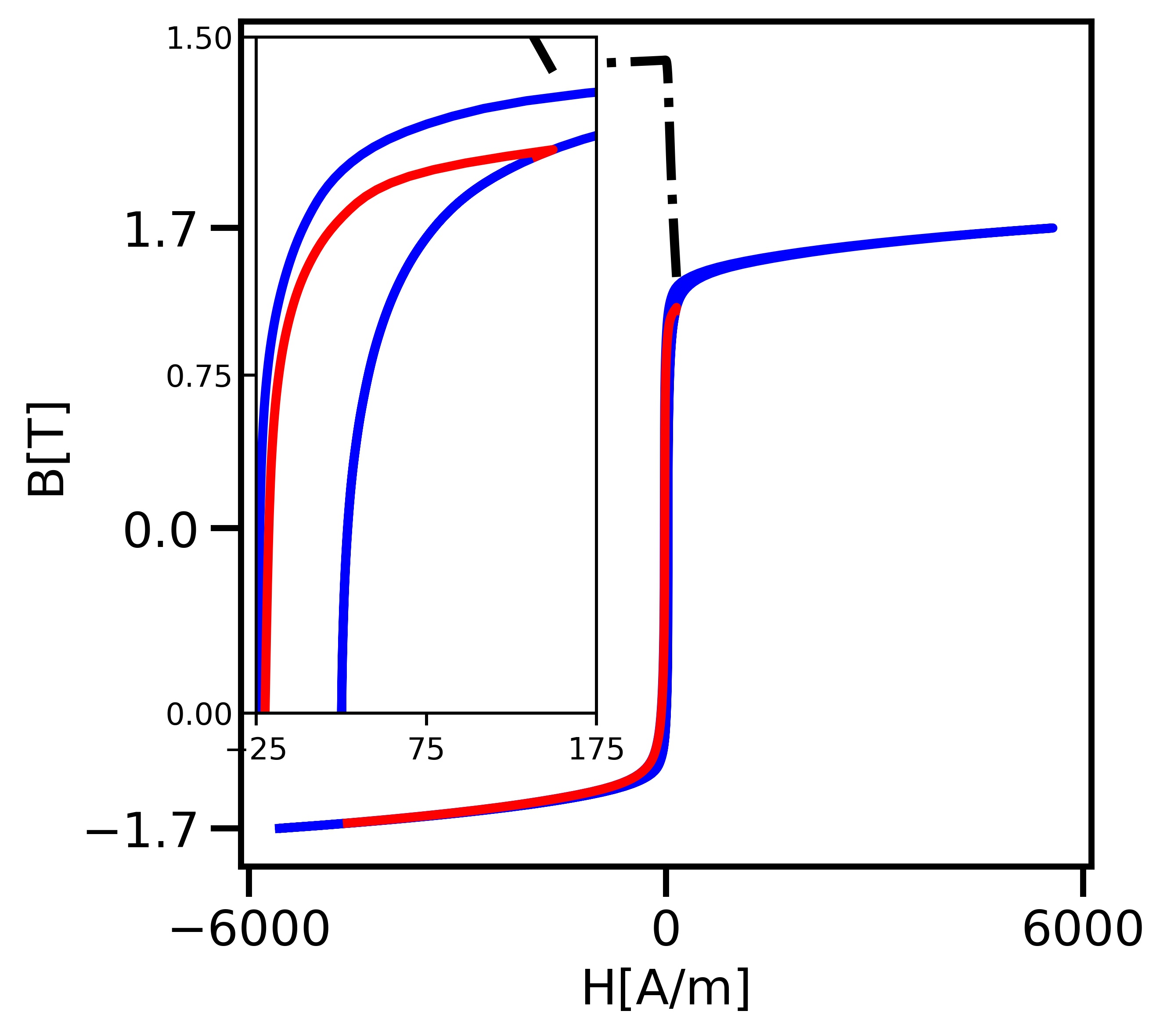}}\hfill
      \subfigure[]{\includegraphics[height=3.5cm, width=3.5cm]{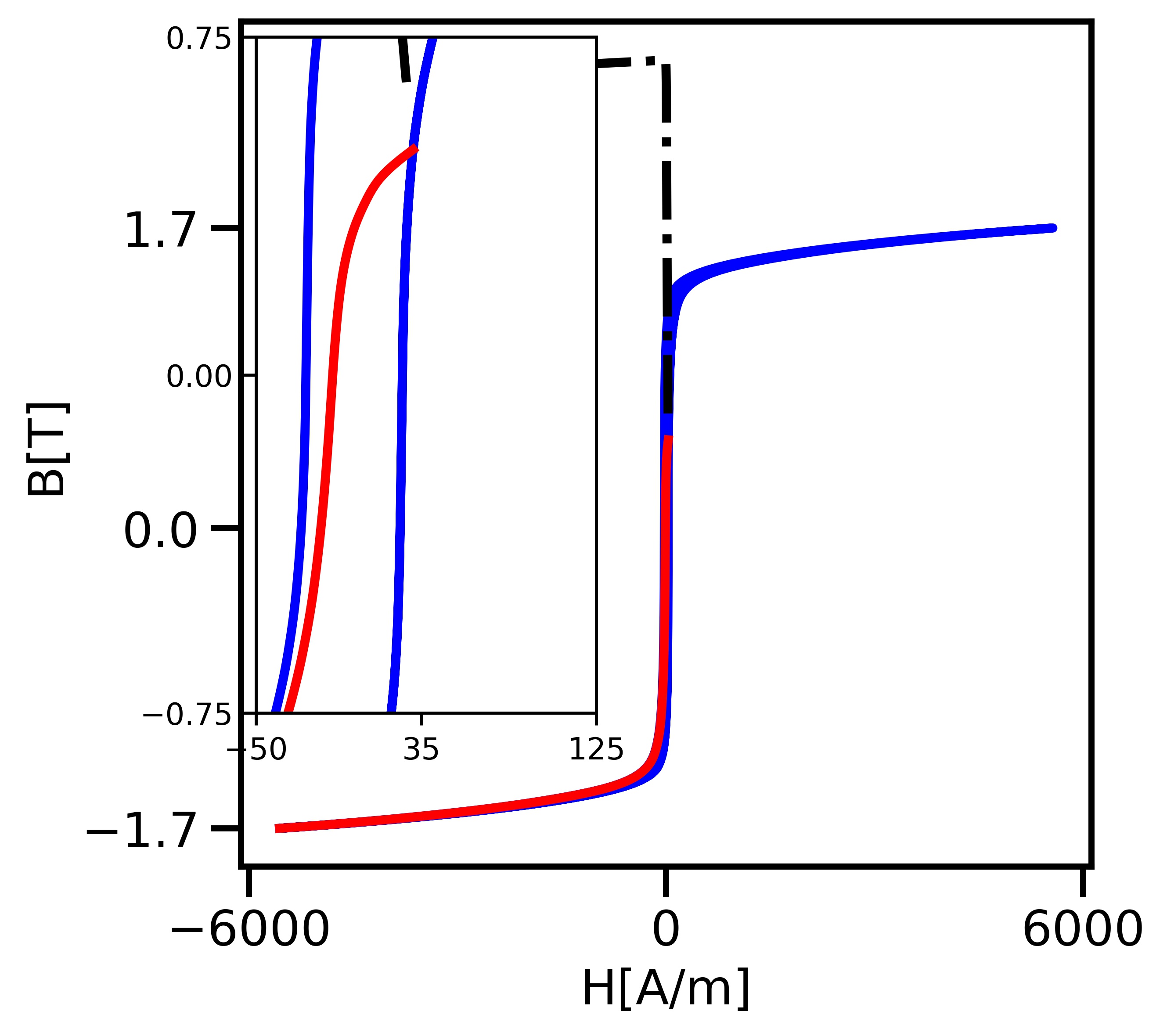}} \hfill
      \subfigure[]{\includegraphics[height=3.5cm, width=3.5cm]{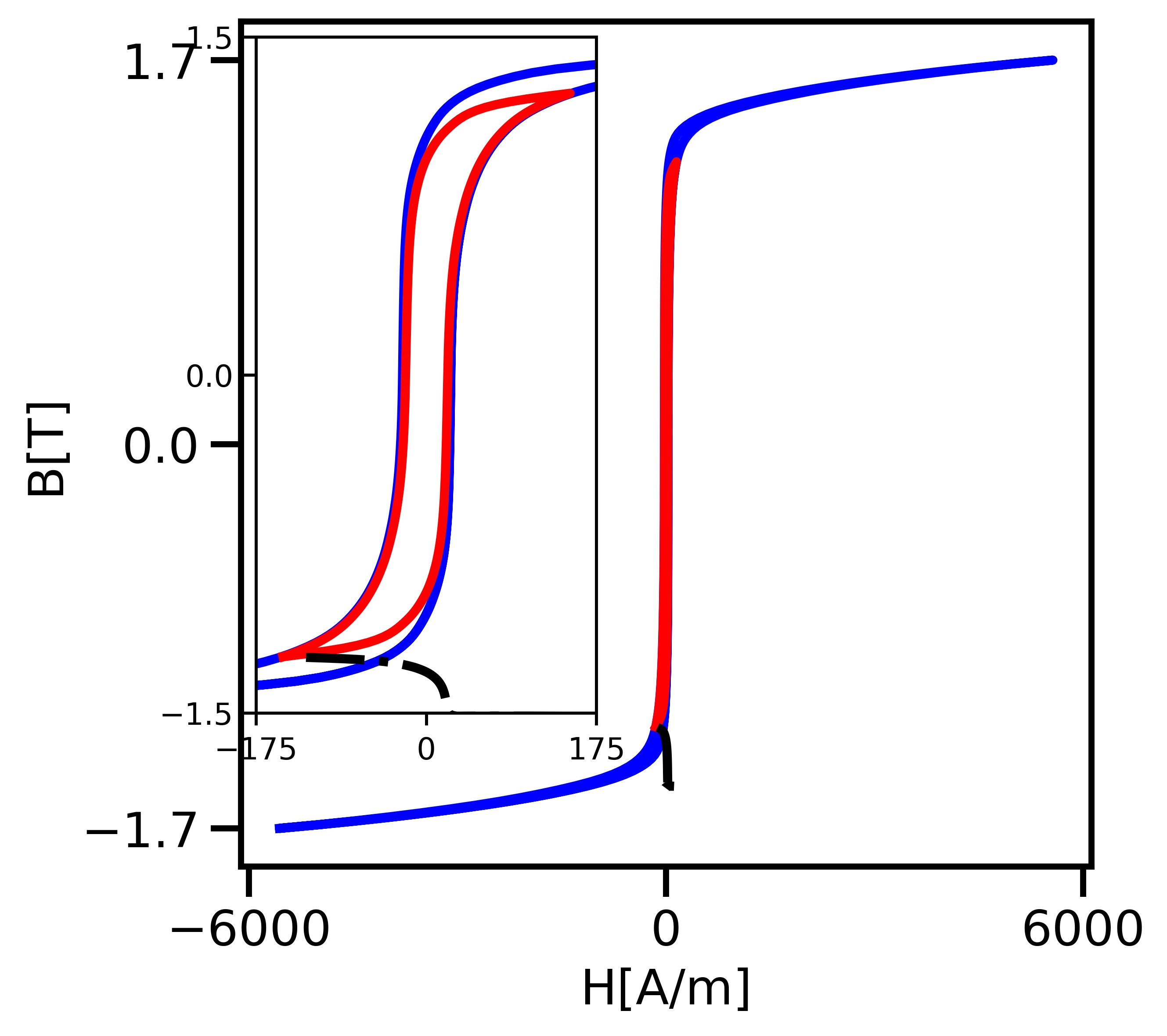}} \hfill
    \subfigure[]{\includegraphics[height=3.5cm, width=3.5cm]{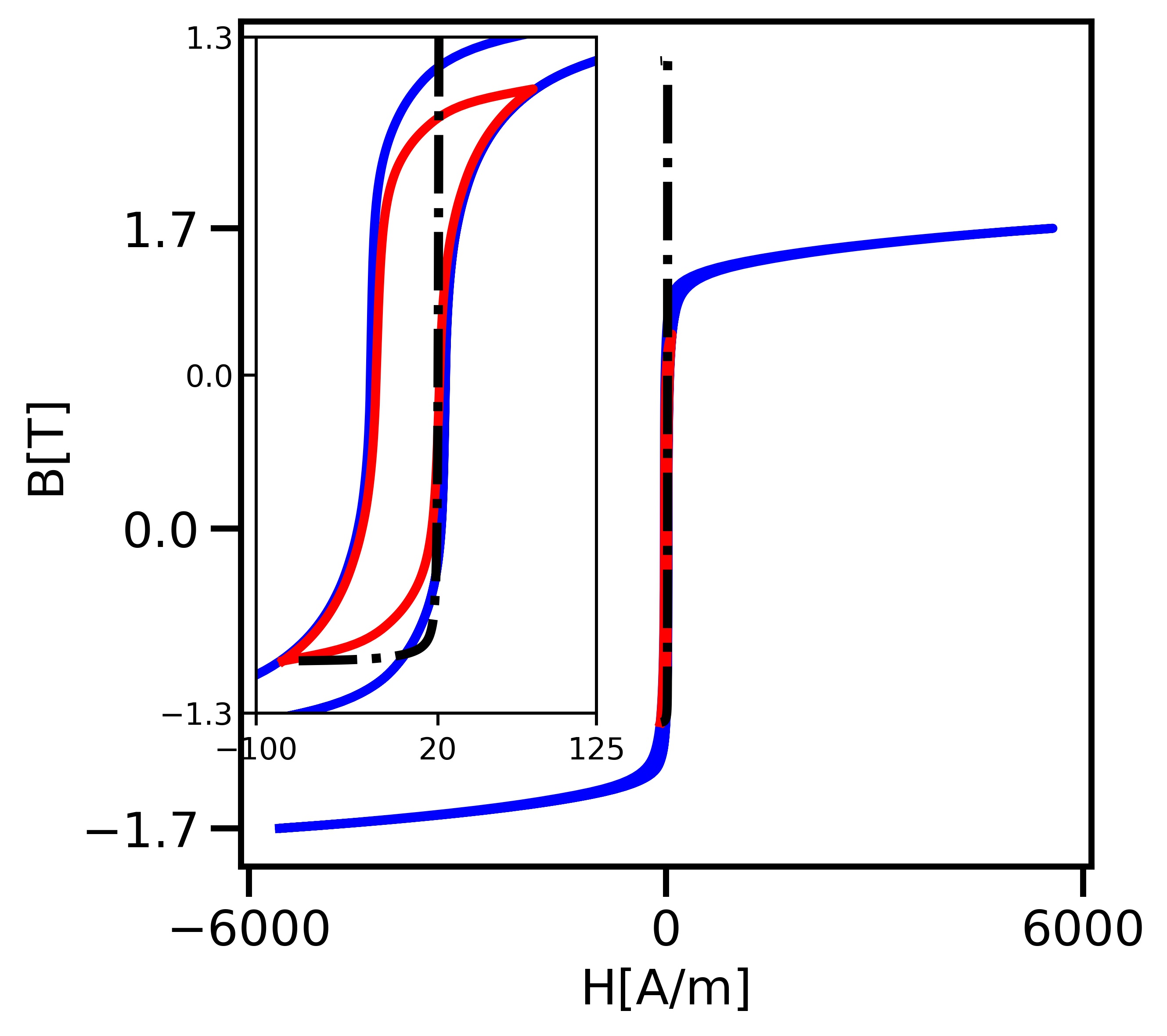}}\hfill
    \caption{Exp 1: RNN prediction for (a:) $\mathcal{C_\mathrm{FORC_{1}}}$; (b:) $\mathcal{C_\mathrm{FORC_{2}}}$; (c:) $\mathcal{C_\mathrm{minor_{1}}}$; (d:) $\mathcal{C_\mathrm{minor_{2}}}$}
\end{figure}

\begin{figure}
    \centering
    \subfigure[]{\includegraphics[height=3.5cm, width=3.5cm]{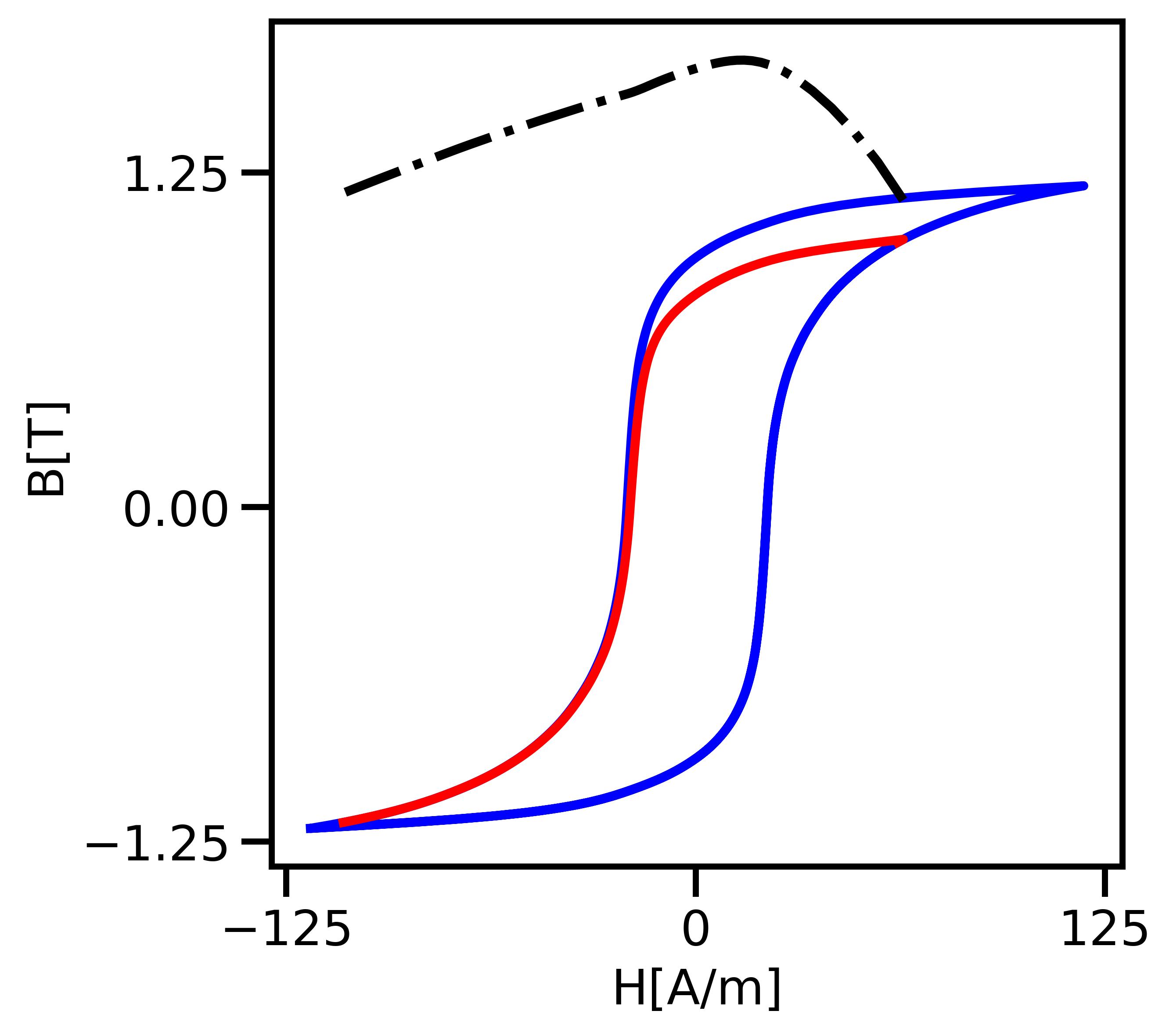}}\hfill
      \subfigure[]{\includegraphics[height=3.5cm, width=3.5cm]{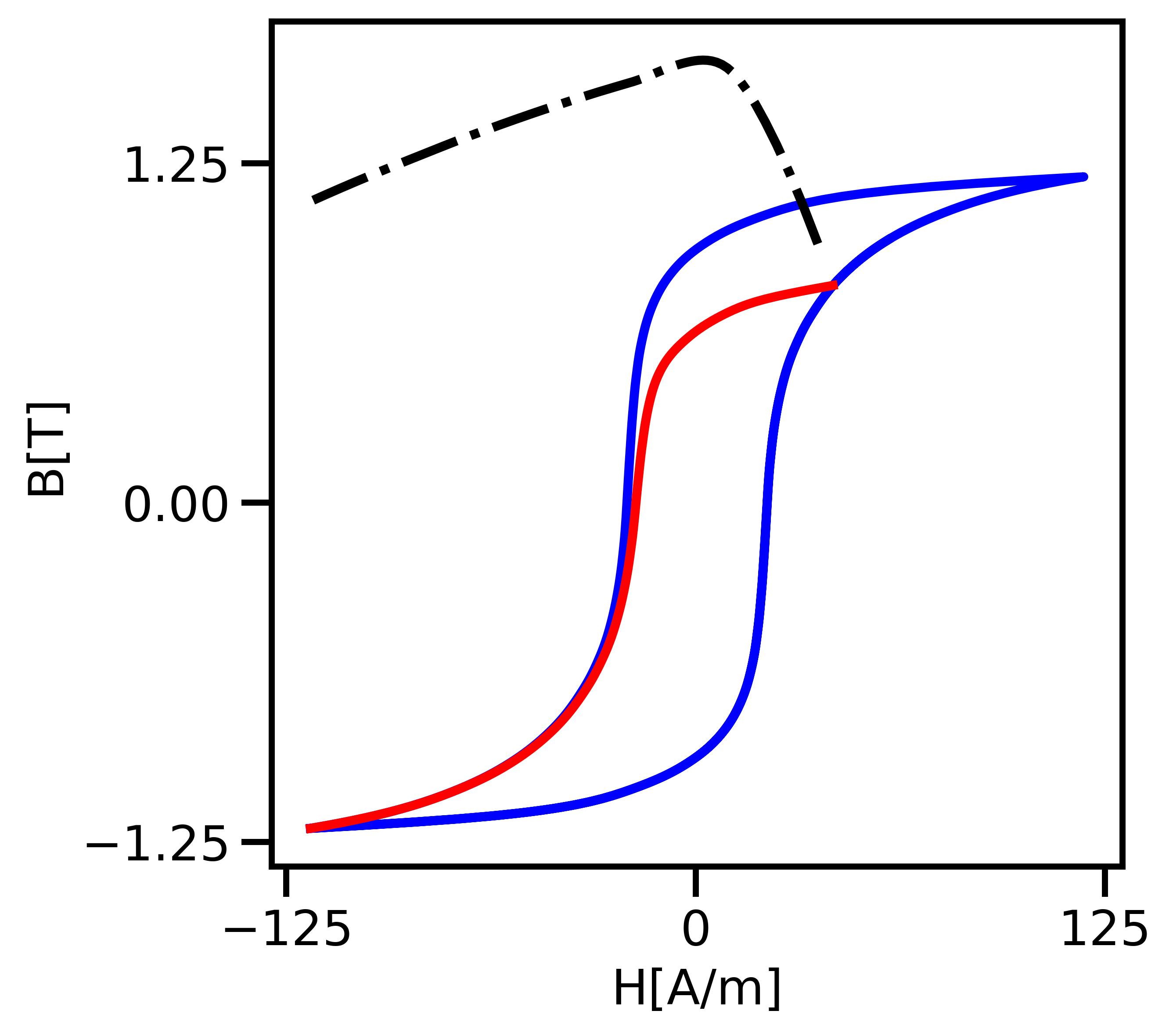}} \hfill
      \subfigure[]{\includegraphics[height=3.5cm, width=3.5cm]{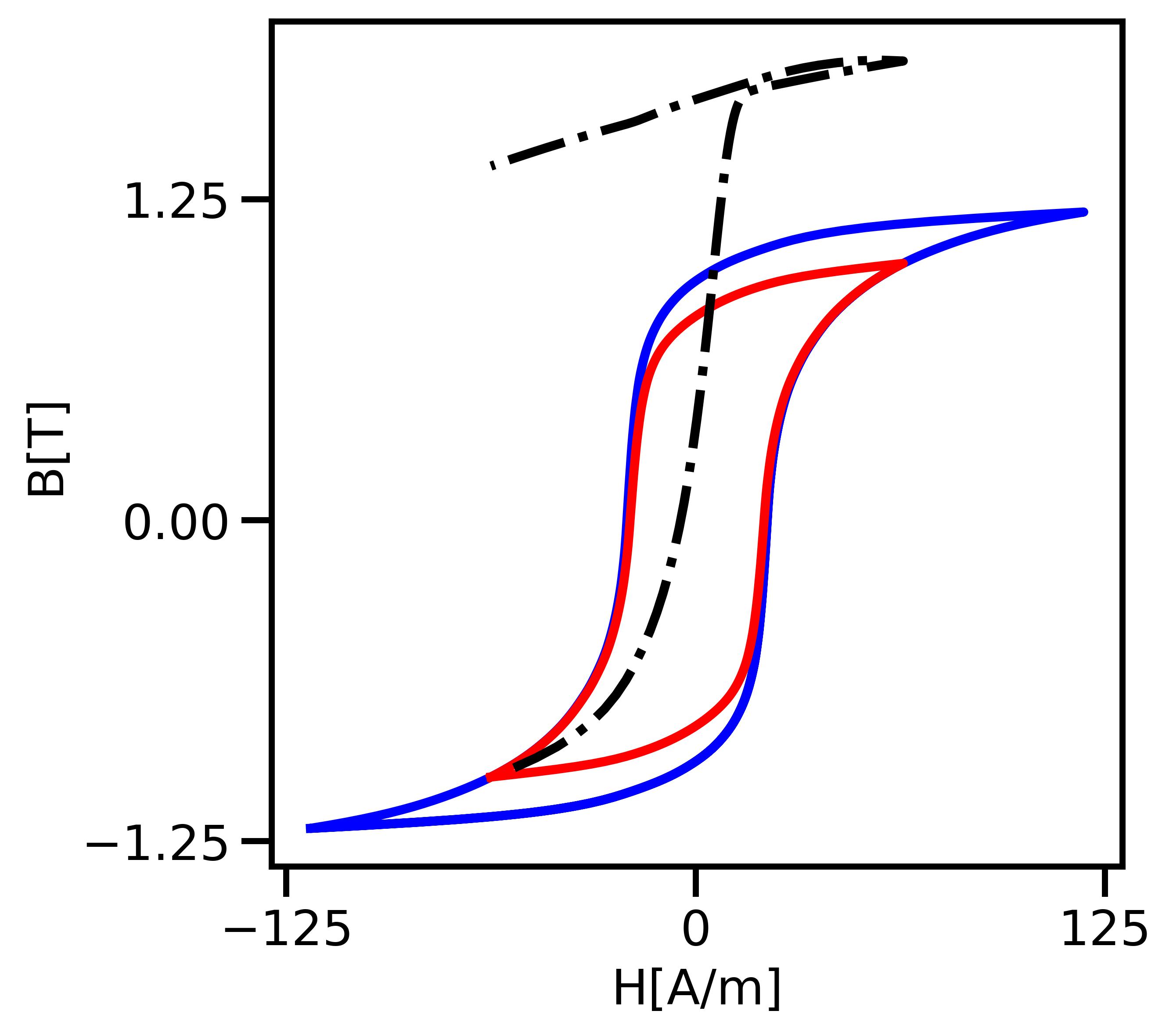}} \hfill
    \subfigure[]{\includegraphics[height=3.5cm, width=3.5cm]{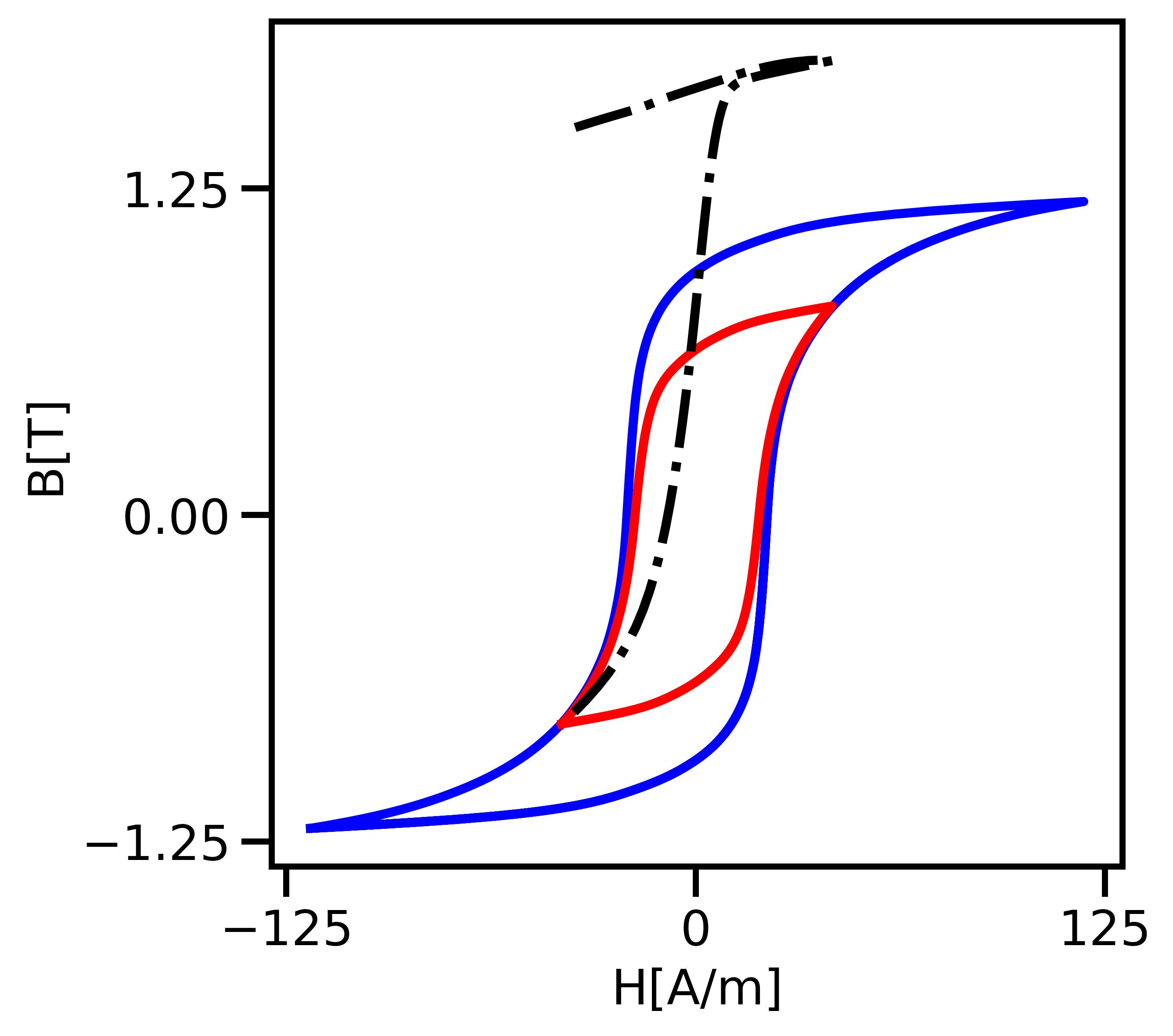}}\hfill
    \caption{Exp 2: RNN prediction for (a:) $\mathcal{C_\mathrm{FORC_{1}}}$; (b:) $\mathcal{C_\mathrm{FORC_{2}}}$; (c:) $\mathcal{C_\mathrm{minor_{1}}}$; (d:) $\mathcal{C_\mathrm{minor_{2}}}$}
\end{figure}

\begin{figure}
    \centering
    \subfigure[]{\includegraphics[height=3.5cm, width=3.5cm]{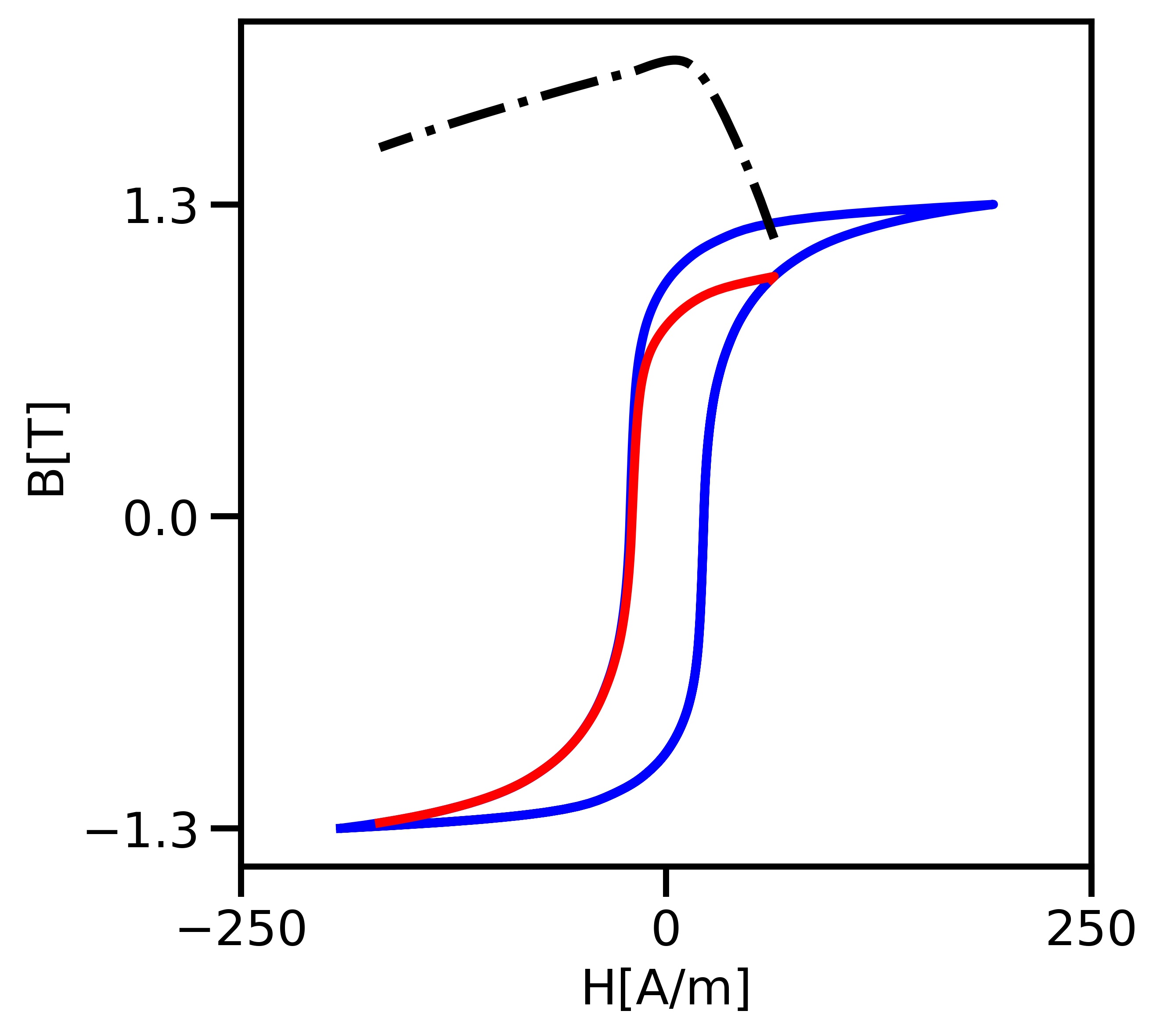}}\hfill
      \subfigure[]{\includegraphics[height=3.5cm, width=3.5cm]{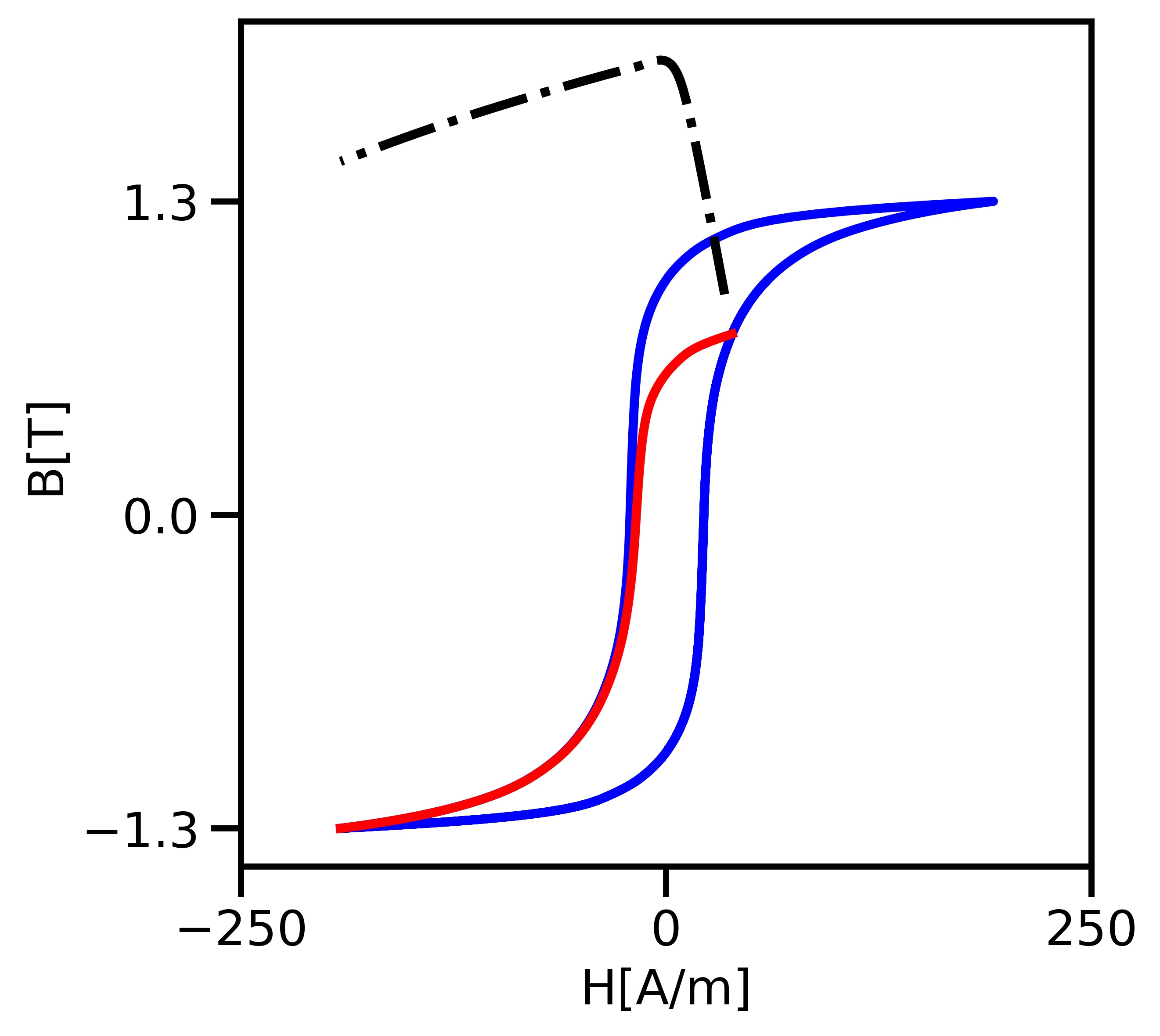}} \hfill
      \subfigure[]{\includegraphics[height=3.5cm, width=3.5cm]{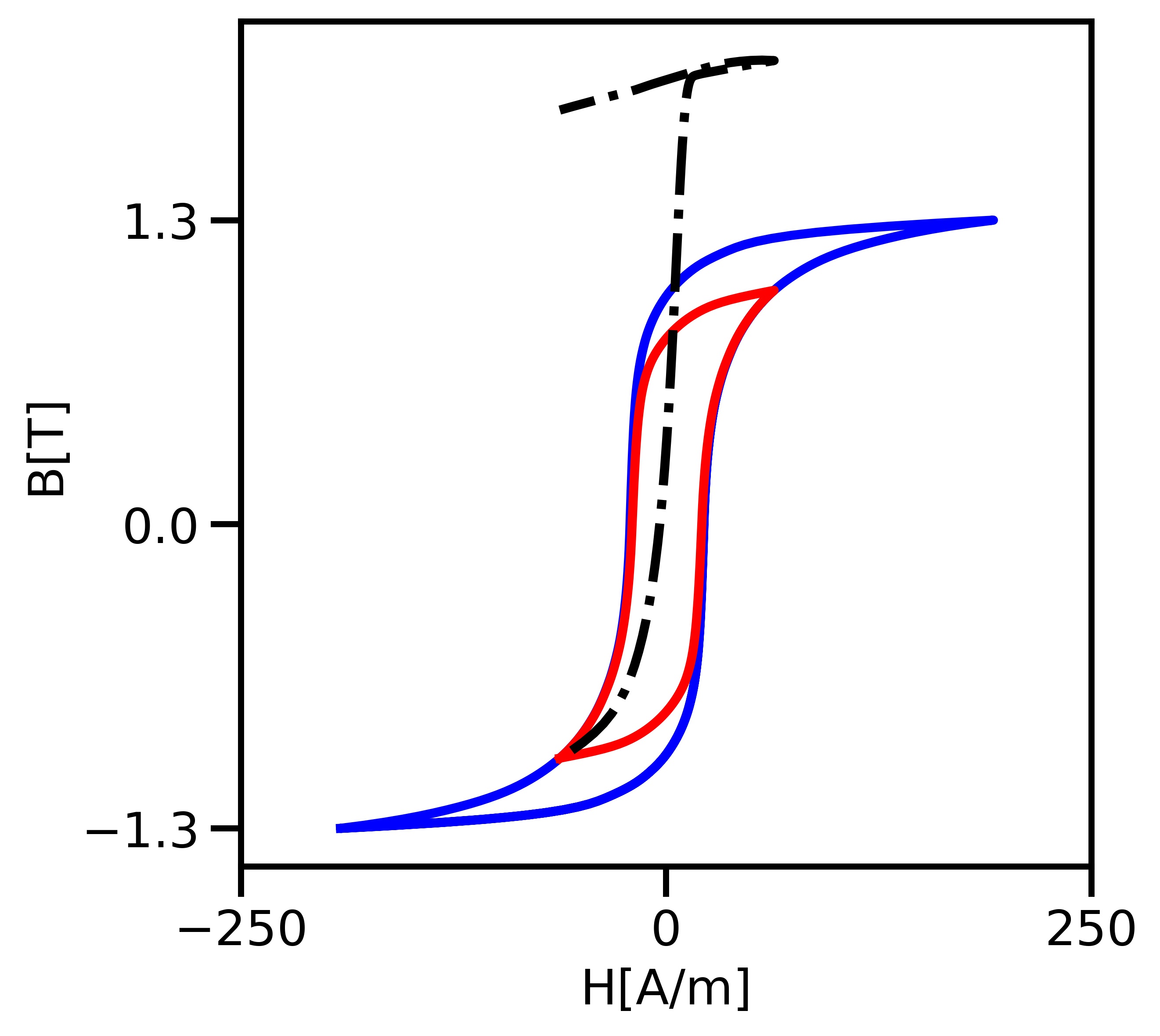}} \hfill
    \subfigure[]{\includegraphics[height=3.5cm, width=3.5cm]{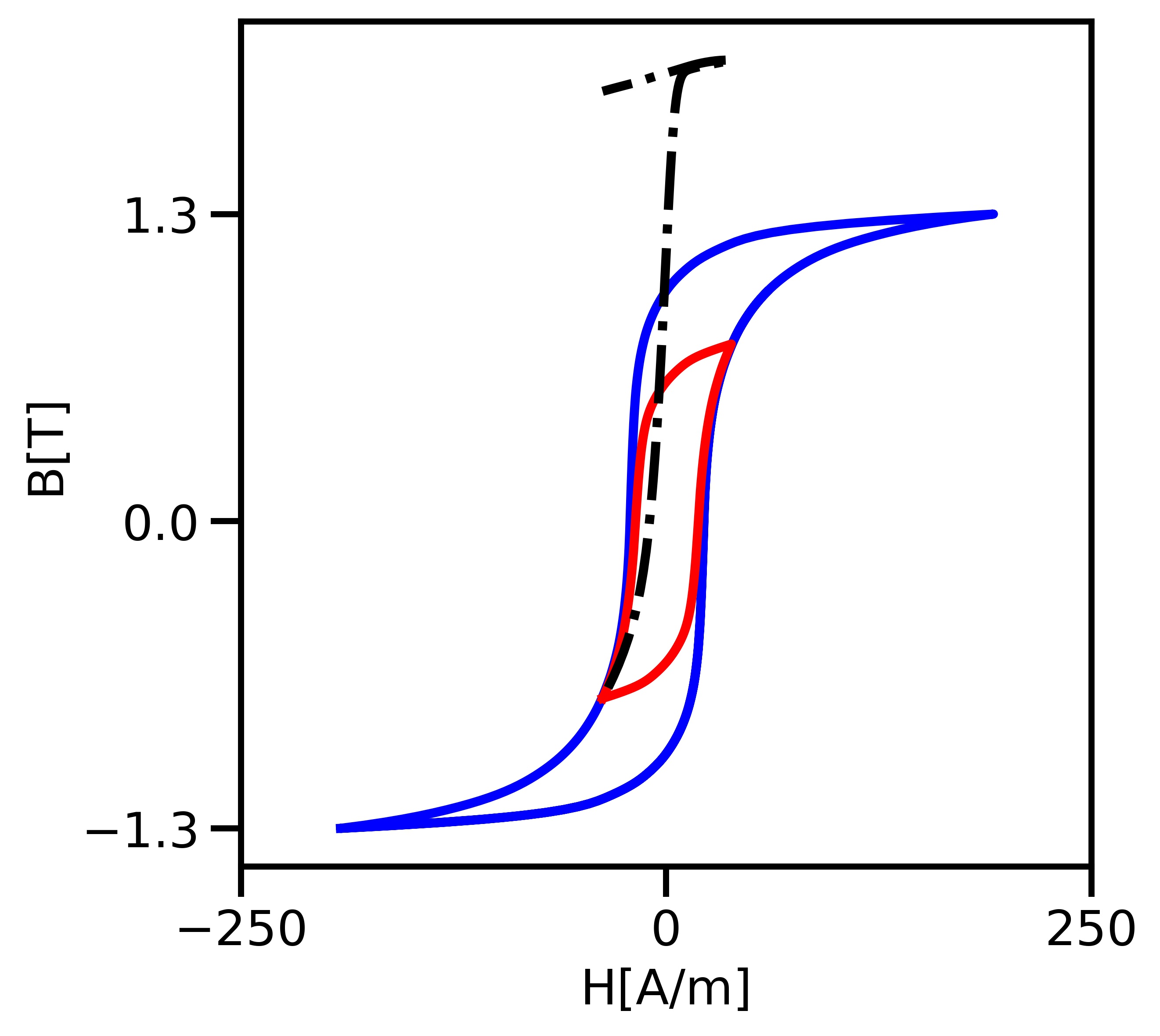}}\hfill
    \caption{Exp 3: RNN prediction for (a:) $\mathcal{C_\mathrm{FORC_{1}}}$; (b:) $\mathcal{C_\mathrm{FORC_{2}}}$; (c:) $\mathcal{C_\mathrm{minor_{1}}}$; (d:) $\mathcal{C_\mathrm{minor_{2}}}$}
\end{figure}

\begin{figure}
    \centering
    \subfigure[]{\includegraphics[height=3.5cm, width=3.5cm]{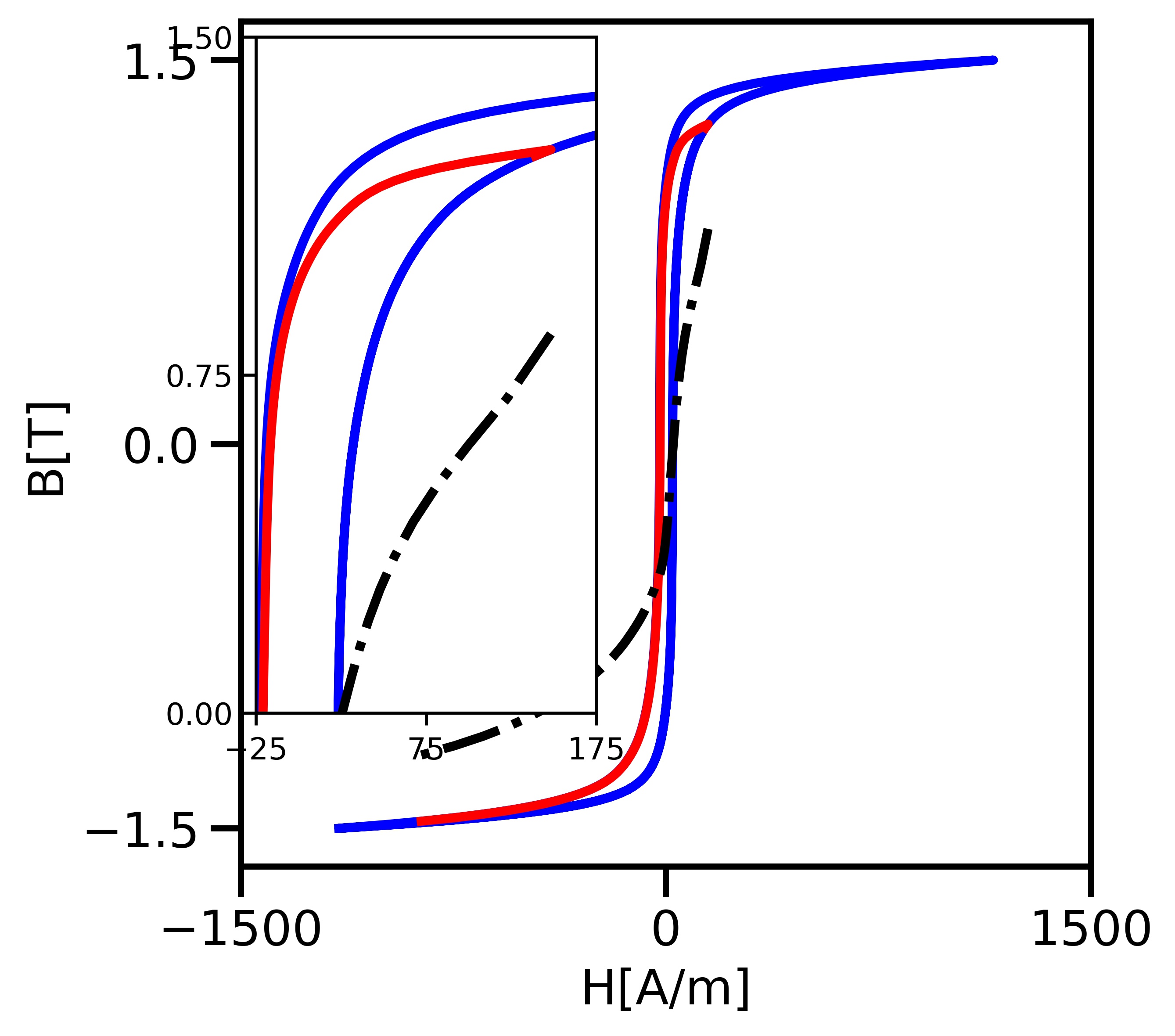}}\hfill
      \subfigure[]{\includegraphics[height=3.5cm, width=3.5cm]{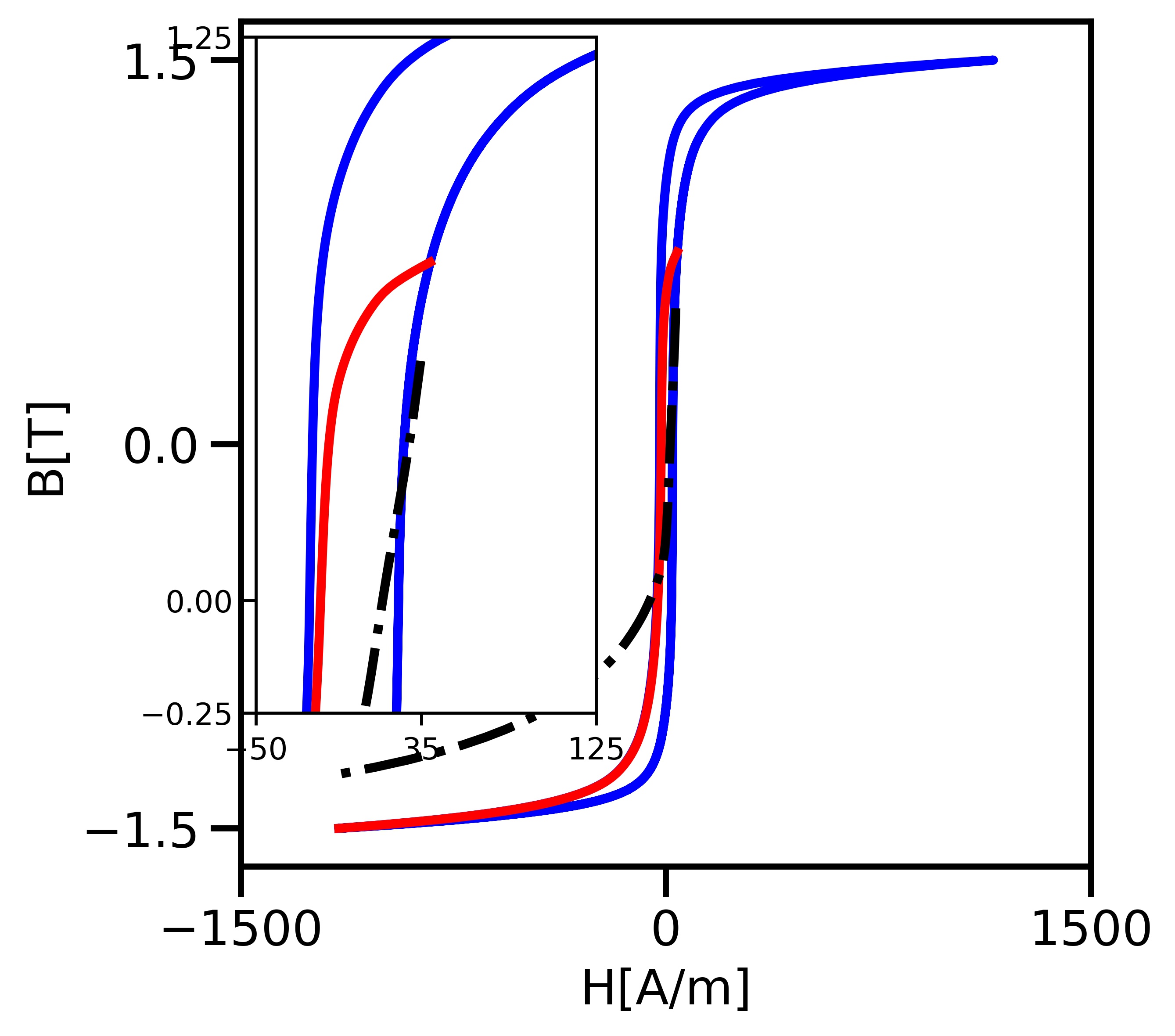}} \hfill
      \subfigure[]{\includegraphics[height=3.5cm, width=3.5cm]{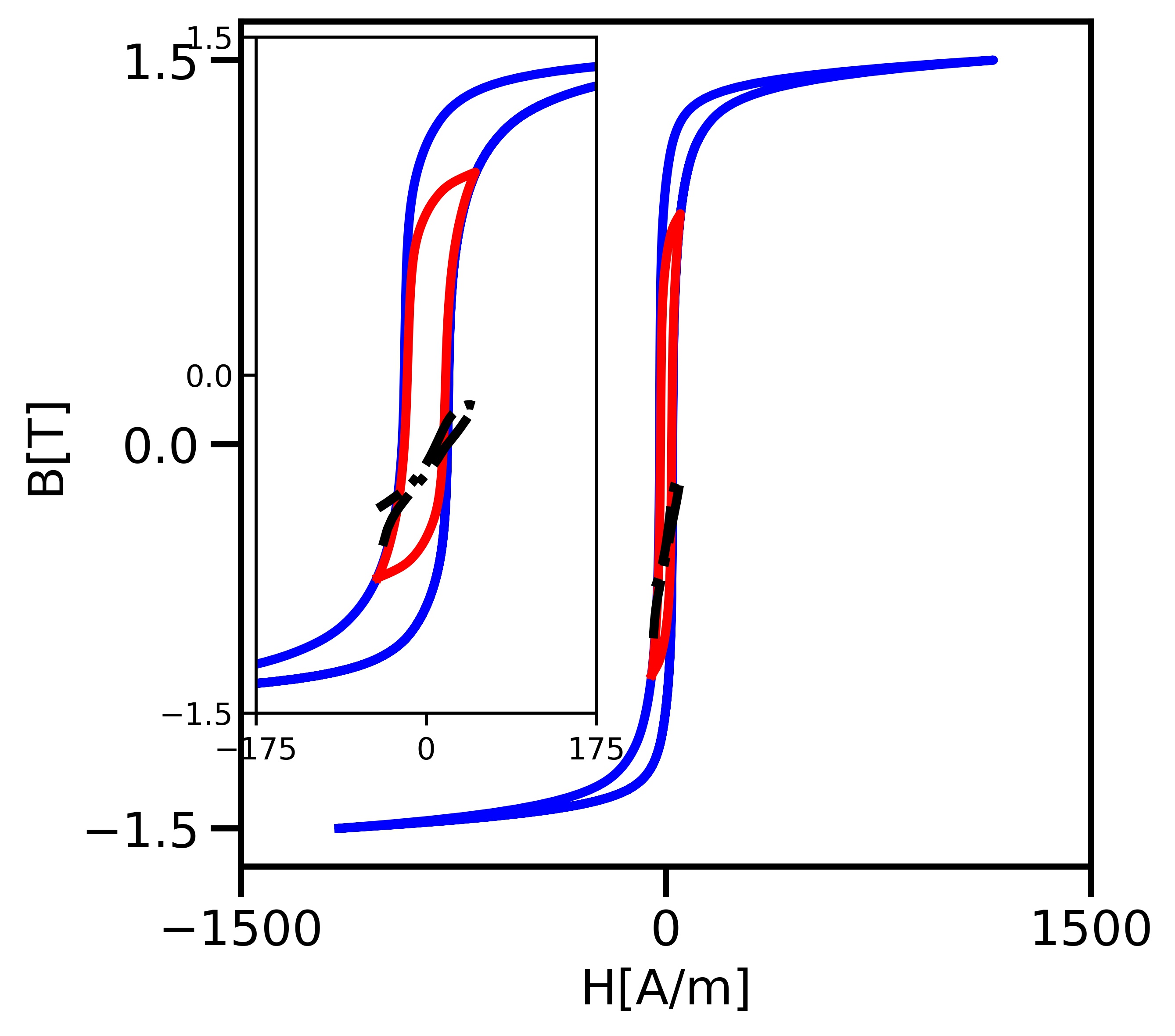}} \hfill
    \subfigure[]{\includegraphics[height=3.5cm, width=3.5cm]{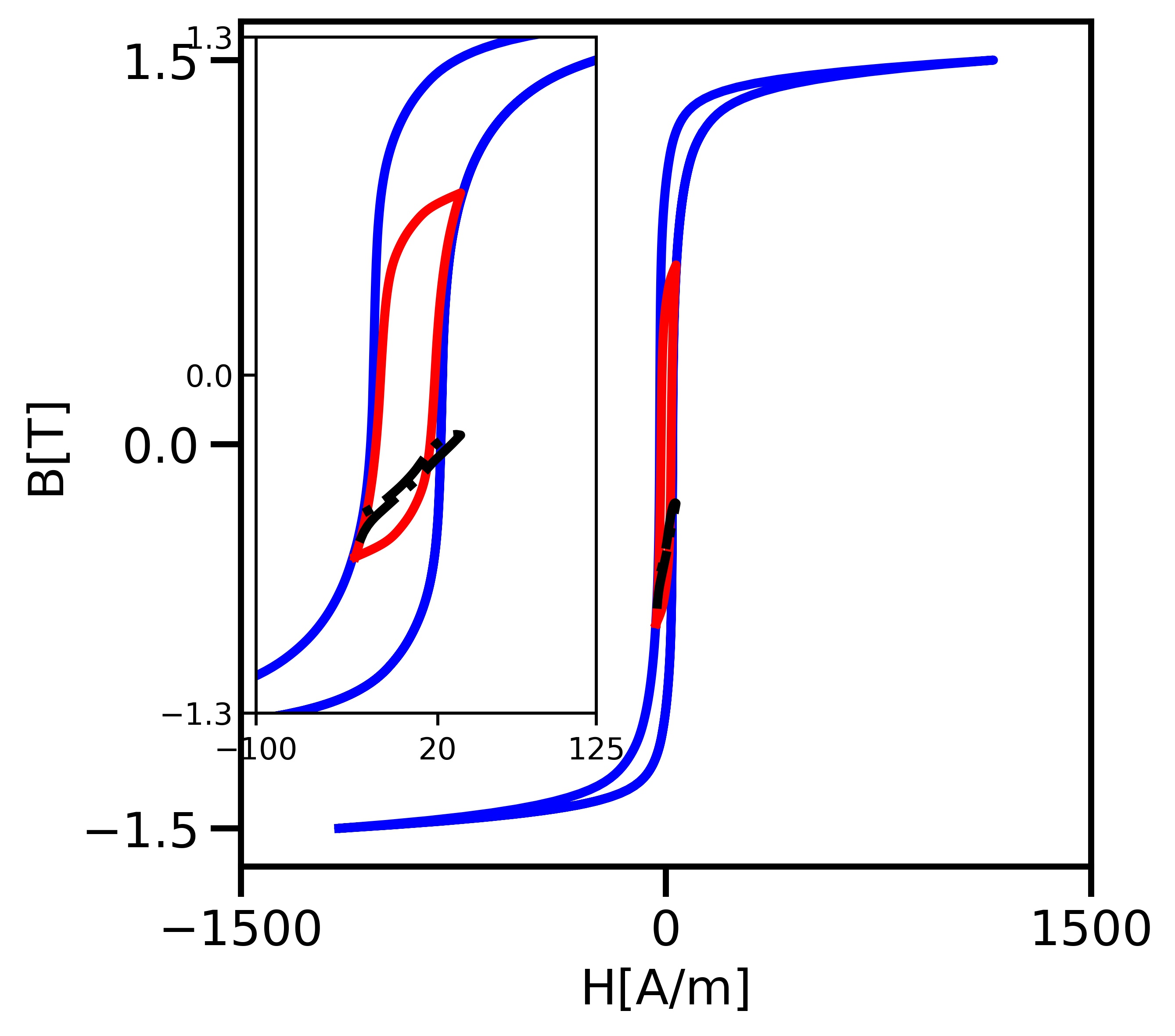}}\hfill
    \caption{Exp 4: RNN prediction for (a:) $\mathcal{C_\mathrm{FORC_{1}}}$; (b:) $\mathcal{C_\mathrm{FORC_{2}}}$; (c:) $\mathcal{C_\mathrm{minor_{1}}}$; (d:) $\mathcal{C_\mathrm{minor_{2}}}$}
\end{figure}

\subsection*{SM \S F: Phenomenological hysteresis differential models}
The Duhem hysteresis model is given by,

\begin{equation*}
    \dot{B}(t) = a  | \dot{H}(t) | \cdot [bH(t) - B(t)] + c  \dot{H}(t)
\end{equation*}
where, $a$,$b$, and $c$ are the parameters that control the shape of the hysteresis loop. The dot over a quantity denotes the time derivative of the quantity. 

The Bouc-Wen model for hysteresis is given by,

\begin{equation*}
    \dot{B}(t) = \alpha \dot{H} - \beta |\dot{H}(t)|\cdot B(t)\cdot |B(t)|^{n - 1} - \gamma \dot{H} |B(t)|^n
\end{equation*}
where, $\alpha$, $\beta$, and $\gamma$ are Bouc-Wen parameters that control the shape of the hysteresis loop.

\subsection*{SM \S G: Details regarding the conducted experiments}
\subsubsection*{Experiment 1}
For the first experiment, $\max(B)$ $=$ \SI{1.7}{\tesla}. The origin of $\mathcal{C_\mathrm{FORC_{1,2}}}$ is taken to be \SI{1.25}{\tesla} and \SI{0.5}{\tesla} respectively. The chosen maximum $B$ value of $\mathcal{C_\mathrm{minor_{1,2}}}$ is \SI{1.25}{\tesla} and \SI{1.1}{\tesla}. The predictions for the model, the ground truth, and the training data are presented in Fig.3.

\subsubsection*{Experiment 2}
For the second experiment, $\max(B)$ $=$ \SI{1.25}{\tesla}. The origin of $\mathcal{C_\mathrm{FORC_{1,2}}}$ is taken to be \SI{1.2}{\tesla} and \SI{1.0}{\tesla} respectively. The chosen maximum $B$ value of $\mathcal{C_\mathrm{minor_{1,2}}}$ is \SI{1.0}{\tesla} and \SI{0.8}{\tesla}. The predictions for the model, the ground truth, and the training data are presented in Fig.4. 

\subsubsection*{Experiment 3}
For the third experiment, $\max(B)$ $=$ \SI{1.3}{\tesla}. The origin of $\mathcal{C_\mathrm{FORC_{1,2}}}$ is taken to be \SI{1.0}{\tesla} and \SI{0.75}{\tesla} respectively. The chosen maximum $B$ value of $\mathcal{C_\mathrm{minor_{1,2}}}$ is \SI{1.0}{\tesla} and \SI{0.75}{\tesla}.

\subsubsection*{Experiment 4}
For the fourth experiment, $\max(B)$ $=$ \SI{1.5}{\tesla}. The origin of $\mathcal{C_\mathrm{FORC_{1,2}}}$ is taken to be \SI{1.25}{\tesla} and \SI{0.75}{\tesla} respectively. The chosen maximum $B$ value of $\mathcal{C_\mathrm{minor_{1,2}}}$ is \SI{0.9}{\tesla} and \SI{0.7}{\tesla}.

\end{document}